\documentclass[12pt]{article}

\usepackage{amsmath,graphicx,amssymb,multirow,setspace}
\usepackage{tabulary}
\usepackage[authoryear,round]{natbib}
\newcommand{\captionfonts}{\normalsize}

\makeatletter  
\long\def\@makecaption#1#2{%
  \vskip\abovecaptionskip
  \sbox\@tempboxa{{\captionfonts #1: #2}}%
  \ifdim \wd\@tempboxa >\hsize
    {\captionfonts #1: #2\par}
  \else
    \hbox to\hsize{\hfil\box\@tempboxa\hfil}%
  \fi
  \vskip\belowcaptionskip}
\makeatother   
\newcolumntype{K}[1]{>{\centering\arraybackslash}p{#1}}

\def\x{{\mathbf x}}

\def\p{{\mathbf p}}
\def\w{{\mathbf w}}

\def\b{{\mathbf b}}

\def\y{{\mathbf y}}

\def\f{{\mathbf f}}

\def\h{{\mathbf h}}

\def\t{{\mathbf t}}

\def\w{{\mathbf w}}

\def\v{{\mathbf v}}

\def\R{{\mathbf R}}

\def\E{{\mathbf E}}

\def\X{{\mathbf X}}

\def\Y{{\mathbf Y}}

\def\cY{{\cal Y}}

\newcommand*\diff{\mathop{}\!\mathrm{d}}

\begin{document}

\begin{center}

{\LARGE A Classification Framework for Partially Observed Dynamical Systems}

\ \\
{\bf \large Yuan Shen$^{\displaystyle 1}$, Peter Tino$^{\displaystyle 1}$, Krasimira   Tsaneva-Atanasova$^{\displaystyle 2,3}$}\\
{$^{\displaystyle 1}$School of Computer Science\\
The University of Birmingham\\
Birmingham, United Kingdom\\
Email: \{y.shen.2$|$pxt\}@cs.bham.ac.uk}\\
{$^{\displaystyle 2}$Department of Mathematics,\\
College of Engineering, Mathematics and Physical Sciences,\\
University of Exeter,\\
Exeter EX4 4QF, UK\\
K.Tsaneva-Atanasova@exeter.ac.uk}\\
{$^{\displaystyle 3}$Department of Mathematics,\\
EPSRC Centre for Predictive Modelling in Healthcare\\
University of Exeter,\\
Exeter EX4 4QF, UK\\
K.Tsaneva-Atanasova@exeter.ac.uk}\\
\end{center}

\begin{abstract}
We present a general framework for classifying partially observed dynamical systems based on the idea of learning in the model space. In contrast to the existing approaches using model point estimates to represent individual data items, we employ posterior distributions over models, thus taking into account in a principled manner the uncertainty due to both the generative (observational and/or dynamic noise) and observation (sampling in time) processes. We evaluate the framework on two testbeds - a biological pathway model and a stochastic double-well system. Crucially, we show that the classifier performance is not impaired when the model class used for inferring posterior distributions is much more simple than the observation-generating model class, provided the reduced complexity inferential model class captures the essential characteristics needed for the given classification task.
\end{abstract}

\section{Introduction}

\label{Sect:Introduction}

Classification is a basic machine learning task. Conventional classification algorithms operate on numerical vectors. Over the past decade, such algorithms have been extended for classifying data with more complex structure, e.g. time series data~\citep{TWL2005,XPK2010}. In many real-world applications, time series data can be irregularly and/or sparsely sampled. This poses a challenge for time series classification. On the other hand, the data-generating processes in such applications could be well understood and mechanistic models accounting for the data structure could have been developed in the form of dynamical systems. Using such mechanistic models in time series classification would allow for natural incorporation of the domain experts' knowledge. In this setting, time series data can be seen as partial observations of the underlying dynamical system and the machine learning task becomes classification of partially observed dynamical systems. In this work, we formulate and validate a general framework for such classification tasks.

\citet{XPK2010}  distinguish two major conventional approaches to time series classification, in particular, feature-based and distance-based approaches. Feature based approaches construct discriminative features on the time series data. These can be local patterns (i.e. short subsequences)~\citep{LZO1999}, or global ones resulting from time-frequency and wavelet analysis~\citep{CCA2002}. Distance-based methods classify time series based on a distance (e.g. Euclidean) between time series pairs. This approach is not directly applicable for the time series of variable length. To circumvent this problem, ``Dynamical Time Wrapping'' (DTW) methods have been developed. In DTW two time series are aligned according some criteria so that a distance can be calculated~\citep{SC1978}. However, such approaches are not applicable for classifying irregularly and sparsely sampled time series. More importantly, they do not utilise the available experts' knowledge about the underlying processes. Alternatively, model-based approaches have also been adopted for time series classification, e.g. Hidden Markov Model (HMM)-based approaches for biological sequence classification~\citep{EB2001}. In those approaches, a prototypical time series model is constructed for each time series class. For example, if the prototypical model is probabilistic, the class label for a new time series is given by the model with the highest likelihood  for that time series (or the highest posterior probability, if class priors are available). However, a single model may not adequately represent all time series in the given class. From this point of view, it is more desirable to represent time series by individual models. In this setting, the classifier employed  classifies individual models (that stand for individual time series) and thus operates in the model space. We refer to this approach as ''{\it Learning in the Model Space}'' (LiMS) and have adopted it for classifying partially observed dynamical systems.

In most of LiMS methods for time series classification,  given a time series, a point estimate of model parameter is used to represent that time series.  Such estimates could be used directly as feature vectors. In this case, any vector-based classifier could be employed for the task. For example, \citet{BSLLBS2011} employ dynamic causal model (DCM)~\citep{FHP03} to represent individual fMRI data from each participant. The maximum-a-posterior estimates of model parameter were then used as feature vectors for classifying DCMs.  In~\citep{CTTY2013,CTTY2015}, a reservoir computation  model was used as a generic non-parametric model to represent non-linear time series data. High dimensional dynamical reservoir was fixed and individual time series were represented by the corresponding read-out mappings from the generic dynamic reservoir. The estimated read-out parameters were then used as feature vector for time series classification. In both approaches, their respective parameter space is considered as a linear metric space and its global metric tensor can be learned in a supervised manner, so as to improve the classification performance. 

Other LiMS approaches use directly model distances (e.g. geodesic on the model manifold) instead of global metric in the parameter space. Such approaches treat the parameter space as a non-linear metric space and learn metric on the underlying manifold. Such non-linear structure could be induced by the intrinsic properties of the underlying processes, or by the constraints imposed on the models (e.g. stability of autoregressive (AR) models).  To compute geodesic distances, one can first reconstruct the underlying metric tensor field in the parameter space. \citet{FC2011} and~\citet{CS2014} propose a general framework based on pullback metric to learn discriminative metric tensors in the space of Linear Dynamical Systems (LDS) and Hidden Markov Models (HMM), respectively.  The manifold structure in the parameter space is induced by stability constraints on the LDS parameters, or by normalisation constraints on the HMM parameters. 

Yet another class of LiMS approaches is formulated in the framework of kernel machines. Although the employed kernels don't fully recover the underlying metric tensor field, they still define useful distance functions that account for the underlying non-linear structure in the parameter space.  Typically, the kernels used have been developed to operate on probability distributions/measures, for example, kernels based on (information-theoretic) divergence functions between two distributions~\citep[e.g. KL divergence,][]{MHV2003}. In particular,  \citet{CV2005,CV2007} used KL-kernels on vector auto-regressive (VAR) models to classify dynamic textures in video sequence analysis. \citet{JKH2004} proposed the probability product kernel (PPK), which can be seen as a dot product in the function space of two probability distributions. Bhattacharyya kernels, a special case of PPK, are related to the Hellinger distance between two functions.  In~\citep{JKH2004} PPK kernels were used to classify both LDS and HMM. Computation of KL and PPK kernels is analytically tractable only for simple classes of dynamical systems, such LDS and HMM. In general, their computation could be very expensive, since it can involve infinite-dimensional integral over all possible state trajectories; Binet-Cauchy kernels could be seen as a counterpart of PPK kernel for deterministic dynamical systems~\citep{VS2006,BCS2007}. In contrast to PPK, Binet-Cauchy kernels are defined as a dot product in the trajectory space. For deterministic systems, their trajectories are completely determined by their model parameters and the initial states.

Finally, we mention two kernels used in the literature for model-based time series classification that fall outside the LiMS framework since no individual models are inferred from individual time series. Fisher kernel proposed by \citet{JH1998} uses a single fixed time series model. Each time series is then represented by a tangent vector in the tangent space of that model.  AR kernel proposed by \citet{CD2011} is a marginalisation kernel applied to AR models~\citep{MS2002}. Each time series is represented by a (infinite-dimensional) "profile vector" - AR likelihood for a set of model parameters, given that time series. The kernel between two time series is the dot product of the two corresponding profile functions, weighted  by a prior distribution over the AR parameters. 

In this paper, we present a general framework for classifying partially observed dynamical systems based on LiMS. One key ingredient of this framework is that given a class of parameterised dynamical system models, we represent each partially observed dynamical system (i.e. each time series) by a posterior distribution over models. In contrast to all model-based approaches surveyed above, our  approach takes into  account the model uncertainty around each individual model. This is of particular relevance for the sparsely sampled time series as it could  give rise to a considerable amount of uncertainty around the inferred model. To classify those posterior distributions, one could employ any classifier that operates on probability distributions, for example, the classifier based on probability product kernel. We also use a distributional kernel induced by the kernel mean embedding (KME)~\citep{MFDS2012}. This embedding maps each distribution onto the Hilbert space induced by a chosen kernel~\citep{SGSS2007}. Note that the PPK kernels here are defined on two distributions over model parameter whereas the PPK kernels in~\citep{JKH2004} are defined on two prior measures over system trajectories\footnote{Each measure is specified by a model parameter vector}. Recall that the latter encodes information about the intrinsic structure in the model space. In our approach, however, this information is encoded in the posterior distributions. This means that in both cases. the classifiers do utilise  the intrinsic structure in the model space for classification. 

The rest of this paper is organised as follows. We first formulate our framework in Section~\ref{Sect:Framework}. Section~\ref{Sect:Implementation} presents an implementation of this framework by means of Kernel Logistic Regression (KLR). In~Section~\ref{Sect:RelatedWorks} we further establish connections between our classifier and two other related state-of-the-art classifiers. Section~\ref{Sect:TestBeds} introduces two classes of dynamical systems used to validate our framework and the experiments are detailed in Section~\ref{Sect:Experiments}. Finally, Section~\ref{Sect:Conclusion} summarises key research findings.

\section{Framework}

\label{Sect:Framework}

\subsection{\bf Problem Settings}

First, a classification task is formulated as follows: Suppose we have $N$ examples in the form of $N$ labelled univariate or multivariate time series, denoted by $\{( \cY^k, c^k): k = 1, ..., N \}$ where $\cY^k$ denotes the $k$-th time series and $c^k$ represents its binary label. As we do not assume that all time series are collected on a fixed, regular time grid, each time series $\cY^k$ is accompanied with a sequence of observation times $\{t^k_i\}_{i=1}^{L_k}$ at which the observations $\{\y^k_i\}_{i=1}^{L_k}$ are collected. Hence the $k$-th time series is jointly represented by $\cY^k = (\t^k, \Y^k)$  with $\t^k = \{t^k_i: i = 1, ..., L^k \}$ and $\Y^k = \{\y^k_i: i = 1, ..., L^k \}$. Note that the length of time series $L_k$ can vary across examples. However, the dimensionality $d$ of the observed time series is assumed to be fixed. The task is to predict a label for a new time series $\cY$ of length $L$. Due to variability of observation times and length of the training time series, direct application of a vector-based classifier would not be suitable. Note that if the training time series were long enough and "suitably" sampled, one could represent each time series through e.g. a vector of Fourier or wavelet coefficients. However, we do not wish to impose any such restrictions and in particular, we are interested in cases of short, sparsely and irregularly sampled time series.

We propose to represent time series by a set of individual time series models from a given model class. In particular, since the observed time series can be noisy, short and irregularly sampled, each time series will be represented as the posterior distribution over the models, given the time series itself  and model prior.
 
\subsection{\bf Model-based Representation}

In our work, a dynamical system approach is adopted to model time series. In other words, we consider a given time series as a (possibly partial) observation of some underlying dynamical  system from a parametric class of dynamical systems. In the following, we first introduce mathematical representation of the class of dynamical systems considered in this work. Next, a model accounting for partial observations is formulated. Following this, we introduce a Bayesian approach for representing partially observed dynamical systems.

A continuous-time deterministic dynamical system can be mathematically represented as a multivariate Ordinary Differential Equation (ODE):
\[
\frac{d\x_t}{dt} = \f(\x_t;\boldsymbol\psi), 
\]
where $\x_t \in X \subset \mathbb{R}^D $ denotes $D$-dimensional state vector at time $t$. The mapping $\f$ specifies the dynamics of this system by defining the functional relation between state $\x_t$ and drift $\frac{d\x}{dt}$ at time $t$. This mapping is parameterised by $\boldsymbol\psi$. Note that model parameter $\boldsymbol\psi$ includes the initial state $\x_0$, unless $\x_0$ is assumed to be known.

A stochastic dynamical system can be considered as an ODE driven by a multivariate random process parameterized by covariance matrix $\boldsymbol \Sigma$. Each component of this process is a standard univariate Brownian motion scaled by square root of the corresponding diagonal term of $\boldsymbol \Sigma$. Its covariance structure at $t$ is specified by the non-diagonal terms. It is equivalent to adding Gaussian noise to the drift. Mathematically, this system can be represented by a multivariate Stochastic Differential Equation (SDE):
\[
d\x_t = \f(\x_t;\boldsymbol\psi) \ \diff t + \boldsymbol\Sigma\  \diff\b_t
\] 
where the vector $\b_t$ collects the $D$ independent standard Brownian motions. A SDE's initial condition is specified by a probability distribution over $\x_0$, which is often assumed to be a Gaussian distribution with mean $ \boldsymbol\mu_0$ and covariance matrix  $\boldsymbol \Sigma_0$. As in ODEs, the initial condition specification is part of the model parameters $\boldsymbol\psi$.

All model parameters are collected in vector  $\boldsymbol\theta$, i.e. for ODEs $\boldsymbol\theta = \boldsymbol\psi$ and for SDEs $\boldsymbol\theta = (\boldsymbol\psi, \boldsymbol\Sigma)$. 

Observations $\{\y_1, \y_2, ... \}$, $\y_t \in Y  \subset \mathbb{R}^d$ from the underlying dynamical system's trajectory $\x_t$ are obtained through a measurement function $\h$:
\[
\y_i = \h(\x_{t_i}) + \boldsymbol\epsilon_{t_i} \quad \mbox{for} \quad i = 1, 2, ...
\]
where $\boldsymbol\epsilon_{t_i}$ denotes observation noise at time $t_i$. In general, $\h$ can be a parametric function with unknown parameters. Frequently, $\h$ represents a set of indicator functions which specify a subset of state variables that are directly observed. For clarity in formulating our general framework, we assume $\h$ to be an identity function. Observation noise $\boldsymbol\epsilon_t$ is often assumed to be i.i.d. Gaussian noise with zero mean and error covariance matrix $\R$. $\R$ can be determined form prior knowledge or learned from the data.

In the {\it learning in the model space} (LiMS) framework, the observed time series are represented through models parametrized via $\boldsymbol\theta$. Given a time series $\cY = (t_i, \y_i)_{i = 1}^{L}$, a Maximum Likelihood (ML) estimate of $\boldsymbol\theta$ can be obtained by maximizing the likelihood function
\begin{equation}
p(\Y |\boldsymbol\theta, \t; \R) = 
\prod_{i=1}^L  {\cal N}\bigg(\y_i \Big| \x_t(\boldsymbol\theta), t_i, \R\bigg) 
\label{ode_likelihood}
\end{equation}
for an ODE system and 
\begin{equation}
p(\Y |\boldsymbol\theta, \t, \R) = 
\mathbb{E}_{\x_t | \boldsymbol\theta} \Bigg[
\prod_{i=1}^L  {\cal N}\bigg(\y_i \Big| \x_t, t_i, \R\bigg) 
\Bigg]
\label{sde_marginal_likelihood}
\end{equation}
for an SDE system. However, this approach ignores uncertainty around the model estimate. In cases where only noisy and/or sparse data are available, any point estimate of the model parameter is not a sufficient representation of the partially observed dynamical system. Instead, the posterior distribution of $\boldsymbol\theta$ should be used, 
\begin{equation}
p(\boldsymbol\theta | \cY,  \R) = 
p(\boldsymbol\theta | \Y, \t,  \R) \propto
p(\Y |\boldsymbol\theta, \t, \R) \cdot 
p(\boldsymbol\theta),
\label{posterior}
\end{equation}  
where  $p(\boldsymbol\theta)$ is the prior over $\boldsymbol\theta$.  In most cases, computation of the normalizing term is analytically not tractable and the posterior has to be approximated by using a finite grid in the parameter space or by sampling/variational methods~\citep{MG2008,DFRH2013,AOSCST2008,GW2008}.

\subsection{\bf Classification Framework}

Formulation of a classifier for partially observed dynamical systems in the LiMS framework depends on the way the underlying systems are represented. We consider two different options: 

\begin{enumerate}
\item
Representation through data {\small ${\cal Y} = (\Y, \t)$}. The resulting classifier operates directly and solely on the data. A probabilistic classifier of this kind is formulated by defining the conditional probability $p(c | \cY)$ that is used to predict label $c$ in a probabilistic manner. Note that this classifier completely ignores the underlying model and thus is of disadvantage if the underlying model structure is known; 
\item
Representation via posterior distributions over models, $\pi(\boldsymbol\theta) \overset{\mbox{\tiny def}}{=} p(\boldsymbol\theta | \cY, \R)$. Thus, the counterpart of  $p(c | {\cal Y})$ is $p(c | \pi)$. The resulting classifier actually operates in the space of posterior distributions rather than in the model or data space. Such posterior distributions not only encode intrinsic information about the underlying dynamical system but also quantitatively represent the uncertainty that arises due to finite number of (possibly irregularly sampled) observations and observation noise. The posterior $\pi(\boldsymbol\theta)$ is shaped by the metric structure in the model space. 
\end{enumerate}

To define our classifier, we first consider the classifier that operates on data. Recall that data ${\cal Y}$ is assumed to be sampled from a hidden trajectory $\x_t$ generated by model instance $\f_{\boldsymbol\theta}$ (a model from model class $\f$  with an unknown  model parameter $\boldsymbol\theta$). To take this additional knowledge into account, we express $p(c | {\cal Y})$ as
\begin{equation}
p(c | {\cal Y}) = \int \diff \x_t \int \diff  \boldsymbol\theta  \ 
p(c | {\cal Y}, \x_t, \boldsymbol\theta) \ 
p(\x_t, \boldsymbol\theta | {\cal Y}, \R ).
\label{marginalisation}
\end{equation}
where the hidden trajectory $\x_t$ and unknown model parameter $\boldsymbol\theta$ are both mar\-gi\-nalised out. The density $p(\x_t | \boldsymbol\theta, {\cal Y}; \R )$ is defined with respect to the standard Brownian motion and $ \int \diff \x_t $ represents path integral over trajectories. The above formulation implies a classifier $p(c | {\cal Y}, \x_t, \boldsymbol\theta)$ which  utilises the model instance $\boldsymbol\theta$, the trajectory $\x_t$ generated by $\f_{\boldsymbol\theta}$, and noisy observations $\cY$ assumed to be sampled from $\x_t$. Given  $\f_{\boldsymbol\theta}$, $\x_t$ is either specified deterministically (in the case of ODEs), or is driven by a standard Brownian motion (in the case of SDEs). Assuming that no additional relevant information for the classification task could be extracted from observation noise or observation times (the noise and observation times processes are not conditional on the class label), all the relevant information in $({\cal Y}, \x_t, \boldsymbol\theta)$ for the class label prediction can be collapsed into the model $\boldsymbol\theta$. Consequently, we replace $p(c | {\cal Y}, \x_t, \boldsymbol\theta)$ with $p(c | \boldsymbol\theta)$.

Eq. (\ref{marginalisation}) now reads:
\begin{eqnarray}
p(c | {\cal Y}) & = & \int \diff \x_t \int \diff  \boldsymbol\theta  \ p(c | \boldsymbol\theta) \ \p(\x_t, \boldsymbol\theta | {\cal Y}, \R ) 
\nonumber \\
                  & = & \int \diff  \boldsymbol\theta   \ p(c | \boldsymbol\theta) \ \int \diff \x_t \ p(\x_t, \boldsymbol\theta | {\cal Y}, \R ) 
\nonumber \\
                  & = & \int \diff  \boldsymbol\theta   \ p(c | \boldsymbol\theta) \ \pi(\boldsymbol\theta ) \nonumber \\
                  & = & \mathbb{E}_{\pi(\boldsymbol\theta)}  \Big[  p(c | \boldsymbol\theta) \Big] \nonumber \\
                  & = &  q(c | \pi).
                  \label{our_classifier}
\end{eqnarray}
Note that the classifier $q(c | \pi)$ operates on posterior distributions $\pi$, but is formulated based on classifier $p(c | \boldsymbol\theta)$  operating in the model space.

In the following, we define the theoretical risk for $q(c | \pi)$. Generally, theoretical risk for a classifier is defined through a joint distribution over the input/label spaces and a loss function quantifying the cost of miss-classification. In our case, the joint distribution of $(\pi,c)$ is written as $p(c) \cdot {\cal P}( \pi | c)$, where ${\cal P}$ denotes a distribution over distributions (random measure). The loss function we employ is the negative log-likelihood, $-\log  p(c | \pi)$. The theoretical risk of $q(c | \pi)$ can be written as
\begin{equation}
{\cal R}({q(c | \pi)})  = \mathbb{E}_{p(c)} \Bigg[   \mathbb{E}_{{\cal P}( \pi | c)}  \Big[   
- \log q(c | \pi) 
\Big]     \Bigg].
\label{eq:Risk} 
\end{equation}

It is difficult to formulate ${\cal P}$ as a parametric generative model. For the classifier $q(c | \pi)$, however, based on (\ref{our_classifier}), we have ${\cal R}(q(c | \pi)) = {\cal R}(p(c | {\cal Y}))$. The theoretical risk for  $p(c | {\cal Y})$ is given by
\[
{\cal R}(p(c | {\cal Y})) = 
\mathbb{E}_{p(c)} \Big[ 
\mathbb{E}_{p(\boldsymbol\theta, \x_t, \Y, \R, \t | c)} \big[ 
- \log p(c | {\cal Y})  
\big] 
\Big] 
\]
where
\[
p(\boldsymbol\theta, \x_t, \Y , \R, \t|c) = p(\boldsymbol\theta | c) \cdot p(\t ) \cdot p(\R) \cdot p(\X_t | \boldsymbol\theta) \cdot p(\Y | \x_t, \t, \R).
\]
A parametric formulation of the above theoretical risk is obtained by adopting {\it (i)} a parametric noise model for $p(\Y | \x_t, \t, \R)$; {\it (ii)} a parametric dynamical noise model $p(\x_t | \boldsymbol\theta)$; {\it (iii)} a prior for the covariance of the observational noise $p(\R)$; {\it (iv)} a point process for $p(\t)$ in the observation window\footnote{We write a density for $\t$ since it is defined with respect to the standard Poisson process.} and {\it (v)} an appropriate model for $p(\boldsymbol\theta | c)$.

\section{Implementation}

\label{Sect:Implementation}

\subsection{Computing the posterior distributions}

For partially observed non-linear dynamical systems the computation of posterior distributions is analytically not tractable. Therefore, the expectation over $\boldsymbol\theta$ w.r.t. $\pi(\boldsymbol\theta)$ in Eq.\ref{our_classifier} can only be computed via approximation. There exist two principled approximation strategies that have the required convergence properties: Approximation by sampling and  Finite-grid approximation.

In the first approach (Approximation by sampling), the posterior distribution is approximated by
\[
\pi(\boldsymbol\theta) \approx \frac{1}{N_{\boldsymbol\theta}} \sum_{n = 1}^{N_{\boldsymbol\theta}} \delta(\boldsymbol\theta - \boldsymbol\theta_n)
\]
where $\boldsymbol\theta_1$, ...,$\boldsymbol\theta_{N_{\boldsymbol\theta}}$ are $N_{\boldsymbol\theta}$ parameter vectors which are independently sampled from $\pi(\boldsymbol\theta)$. Accordingly, the classifier defined in Eq.~\ref{our_classifier} is approximated by
\begin{equation}
q(c | \pi) \approx 
\frac{1}{N_{\boldsymbol\theta}} \cdot \sum_{n = 1}^{N_{\boldsymbol\theta}}  
  p(c|\boldsymbol\theta_n).
%\label{appprox-eq-def-clas}
\end{equation}
As the posterior distribution is only known up to normalising constant, MCMC algorithms are the most efficient sampling method. 

In the second approach (finite-grid approximation), one could first compute the unnormalised posterior density (that is, the product of normalised prior and likelihood densities) over a finite grid approximating the parameter space and then normalise those values into a multinomial distribution approximating the posterior density. We denote this grid and the multinomial posterior probabilities on the grid by 
\[
{\cal G}_{\boldsymbol\theta} = \{ \boldsymbol\theta^{\cal G}_1, ..., \boldsymbol\theta^{\cal G}_{N^{\cal G}_{\boldsymbol\theta}} \}
\]
and 
\begin{equation}
\Big\{ \pi^n = \frac{p(\cY |\boldsymbol\theta^{\cal G}_n) \cdot p(\theta^{\cal G}_n)}{\sum_{k = 1}^{N^{\cal G}_{\boldsymbol\theta}} p(\cY |\boldsymbol\theta^{\cal G}_k)  \cdot p(\theta^{\cal G}_k)}:
n = 1, ..., N^{\cal G}_{\boldsymbol\theta}\Big\},
\end{equation}
respectively. The resulting approximate classifier is given by
\begin{equation}
q(c | \pi) \approx 
\sum_{n = 1}^{N^{\cal G}_{\boldsymbol\theta}} \pi^n \cdot
 p(c|\boldsymbol\theta^{\cal G}_n).
%\label{appprox-eq-def-clas}
\end{equation}
For SDE, however, the marginal likelihood for each parameter vector on the grid is analytically not tractable and thus the likelihood is not normalised. To solve this problem at low computational cost, we employ the Variational Gaussian Process Approximation method for computing the approximate marginal likelihood~\citep{AOSCST2008}. 

\subsection{Kernel Logistic Regression for Binary Classification}

\label{sec:klr}

In the following, we first briefly introduce Kernel Logistic Regression (KLR) as a (binary) classifier for vectors (e.g. model parameter $\boldsymbol\theta$). We then present an extension of KLR for distributions so that the classifier can be directly applied to posteriors $\pi(\boldsymbol\theta)$.

A binary KLR classifier operating on $\boldsymbol\theta$s is defined via
\begin{equation}
p( c = 1 | \boldsymbol\theta) 
 = \zeta \Big( \w^{\intercal}\boldsymbol\Phi(\boldsymbol\theta)  \Big),
\end{equation}
where $\zeta(\cdot)$ denotes a sigmoid function\footnote{For real number a, $\zeta(a)$ is defined as $\frac{1}{1 + \exp(-a)}$.}, $\w$ is $m$-dimensional classifier parameter, and $\boldsymbol\Phi$ represents a (non-linear) mapping of $D_{\boldsymbol\theta}$-dimensional model parameter vector $\boldsymbol\theta$ to $m$-dimensional feature  space:
\[
\boldsymbol\Phi: \boldsymbol\theta \longmapsto  \Big[{K}\big(\boldsymbol\theta, \boldsymbol\theta^{\cal F}_1\big), ..., 
{K}\big(\boldsymbol\theta, \boldsymbol\theta^{\cal F}_m \big)\Big]^{\intercal},
\]
where ${K}$ represents a kernel function operating on the the space of model parameter vectors  and ${\cal F}_{\boldsymbol\theta} = \{\boldsymbol\theta^{\cal F}_1, ...,  \boldsymbol\theta^{\cal F}_m\}$ denotes the set of model parameters for constructing this feature map. In this work, we adopt a Gaussian kernel, 
\[
{K}\Big(\boldsymbol\theta_1, \boldsymbol\theta_2\Big) = \exp \biggl( 
-\frac{ \| \boldsymbol\theta_1 - \boldsymbol\theta_2 \|^2}{\rho}
\biggr),
\]
where $\| \cdot \|$ is the Euclidean norm and $\rho>0$ is a scale parameter.

For learning the classifier parameter $\w$,  a training set of $N$ labelled model parameters $S = \{({\boldsymbol\theta}_1, c_1), ..., ({\boldsymbol\theta}_N, c_N)\}$, $c_i \in \{ 0,1\}$, would be used to obtain the maximum likelihood estimate (MLE) of $\w$:
\[
\hat \w^{\mbox{\tiny \bf MLE}} = \underset{\w}{\operatorname{arg\,max}} \prod_{k = 1}^N z_k^{c_k}  (1-z_k)^{1-c_k} 
\quad \mbox{with} \quad z_k = \zeta \big( \w^{\intercal}\boldsymbol\Phi({\boldsymbol\theta}_k)  \big).
\]
This is equivalent to minimizing the Cross Entropy Error 
\[
E_{\boldsymbol\theta}(\w|S) = - \sum_{k = 1}^N \log p(c_k|{\boldsymbol\theta}_k;\w).
\]
For a gradient-based minimisation of $\E$ w.r.t. $\w$, the gradient is computed as 
\[
\nabla_{\w} \E = \sum_{k: c_k = 1} (z_k - 1) \cdot \boldsymbol\Phi({\boldsymbol\theta}_k)
+ \sum_{k: c_k = 0} z_k \cdot \boldsymbol\Phi({\boldsymbol\theta}_k).
\]

For a classifier that operates on the posterior distributions and a training set given as 
$
V = \bigg\{\big(\pi_1(\boldsymbol\theta), c_1\big), ..., \big(\pi_N(\boldsymbol\theta), c_N\big)\bigg\},
$
the classifier parameter is obtained by minimizing the Cross Entropy Error
\[
E_\pi(\w|{V}) = - \sum_{k = 1}^N \log 
\biggl(
\int    \diff \boldsymbol\theta \ \pi_k(\boldsymbol\theta)    \cdot p(c_k|\boldsymbol\theta;\w)
\biggr).
\]
The approximate cross-entropy error is computed by
\[
\hat E_\pi(\w|V) = - \sum_{k = 1}^N \log 
\biggl(
\sum_{n = 1}^{N^{\cal G}_{\boldsymbol\theta}} \pi^n_k  \cdot p(c_k|\boldsymbol\theta^{\cal G}_n;\w)
\biggr),
\]
where $\pi^n_k$ denotes the normalised posterior weight on the $n$-th grid point for the $k$-th posterior. The corresponding gradient is given by
\[
\nabla_{\w} \hat E_\pi  = \sum_{n = 1}^{N^{\cal G}_{\boldsymbol\theta}}  \Bigg[
Z_n^1   \cdot
\Big( (z_n - 1) \cdot \boldsymbol\Phi(\boldsymbol\theta^{\cal G}_n) \Big)
 +
 Z_n^0 \cdot 
\Big( z_n \cdot \boldsymbol\Phi(\boldsymbol\theta^{\cal G}_n) \Big) \Bigg]
\]
where $z_n = \zeta \big( \w^{\intercal}\boldsymbol\Phi(\boldsymbol\theta^{\cal G}_n) \big)$,
\[
Z_n^1 = \sum\limits_{k: c_k = 1}  \frac{\pi^n_k \cdot z_n}{\sum_{l = 1}^{N^{\cal G}_{\boldsymbol\theta}} \pi^l_k \cdot z_l} \ \ \ \ \ \hbox{and} \ \ \ \ \
Z_n^0 = \sum\limits_{k: c_k = 0}   \frac{\pi^n_k \cdot (1 - z_n)}{\sum_{l = 1}^{N^{\cal G}_{\boldsymbol\theta}} \pi^l_k \cdot (1 - z_l)}.
\]
Note that $\boldsymbol\Phi(\boldsymbol\theta^{\cal G}_n)$ is a $m$-dimensional vector whose $j$-th component is given by $K\Big(\boldsymbol\theta^{\cal G}_n, \boldsymbol\theta^{\cal F}_j\Big)$. The two grids on the parameter space, 

\section{Connection to Related Works}

\label{Sect:RelatedWorks}

In literature, most distributional classifiers combine an existing kernel-based classfier, such as SVM, with a kernel that is defined on the space of distributions. An example of such a kernel is the so-called probability Product Kernel~\citep{JKH2004},
\[
K_{\mbox{\tiny PPK}}(\pi_1, \pi_2) = \int_{{\varTheta}} \diff{\boldsymbol\theta} \ \pi_1^\alpha(\boldsymbol\theta) \cdot \pi_2^\alpha(\boldsymbol\theta), 
\]
where $\pi_1$ and $\pi_2$ are two distributions over a metric space ${\varTheta}$ and $\alpha>0$ is a tempering parameter. In recent literature, another kernel on distributions has been introduced based on Hilbert Space Embedding~\citep{SGSS2007,MFDS2012}. Given a universal kernel $k: {\varTheta} \times {\varTheta} \longrightarrow \mathbb{R}$, there exists an injective mapping from distribution space $Q$ to feature space,
\begin{equation}
\mu_{Q}: {Q} \rightarrow {\cal H}, \quad \pi   \longmapsto \int_{{\varTheta}} k(\boldsymbol\theta, \cdot) \pi(\boldsymbol\theta) \diff{\boldsymbol\theta}.
\end{equation}
This mapping is called kernel mean embedding (KME). As the embedding is bijective, no information encoded in the probability distribution is lost through the mapping. The mapping in turn defines a kernel on probability distributions, $K: {\cal Q} \times {\cal Q} \longrightarrow \mathbb{R}$:
\begin{equation}
K_{\mbox{\tiny KME}}(\pi_1, \pi_2) = \langle \mu_{\pi_1}, \mu_{\pi_2} \rangle_{\cal H} 
= \int_{{\boldsymbol\theta \in \varTheta}} \diff{\boldsymbol\theta}  
\int_{{\boldsymbol\eta \in \varTheta}} \diff{\boldsymbol\eta}  \ \ \ \pi_1(\boldsymbol\theta) \cdot \pi_2(\boldsymbol\eta) \cdot k(\boldsymbol\theta, \boldsymbol\eta).
\end{equation}

We compare these two distributional classifiers (one based on $K_{\mbox{\tiny PPK}}$, the other one based on $K_{\mbox{\tiny KME}}$) with our classifier in terms of their predictive class distributions, given a test input (distribution) $\pi$:
\begin{itemize}
\item Probabilistic classifier based on Probability Product Kernel (PPK):
\begin{eqnarray}
p(c = 1 |\pi) 
& = &
\zeta \Bigg( 
\int_{{\varTheta}} \pi^\alpha(\boldsymbol\theta) \cdot  
\bigg[ 
\underbrace{\sum_{i = 1}^N v_i \cdot \pi_i^\alpha (\boldsymbol\theta)}_{ \Upsilon_{\mbox{\tiny PPK}}(\boldsymbol\theta; \v)}
\bigg]  \diff{\boldsymbol\theta}
\Bigg) 
\label{eq:f_ppk}
\\ 
& = & 
\zeta \Bigg(   \mathbb{E}_{\pi}  \bigg[ \Upsilon_{\mbox{\tiny PPK}}(\boldsymbol\theta; \v) \bigg]   \Bigg),
\label{eq:c_ppk}
%\\
%& = &
%\sigma(\nu(\pi; \w))
\end{eqnarray}
where $\Upsilon_{\mbox{\tiny PPK}}(\boldsymbol\theta; \v)$ denotes the function to be learnt (by adjusting the free parameter $\v$) for classifying distributions;
\item Probabilistic classifier based on Kernel Mean Embedding (KME):
{\small 
\begin{eqnarray}
p(c = 1 | \pi) 
& = &
\zeta \Bigg( 
\int_{\boldsymbol\theta \in {\varTheta}} \pi(\boldsymbol\theta) \cdot 
\bigg[ 
\int_{\boldsymbol\eta \in {\varTheta}} 
\big[ 
\sum_{i = 1}^L v_i \cdot \pi_i(\boldsymbol\eta)
\big]  
\cdot k(\boldsymbol\theta, \boldsymbol\eta) \diff{\boldsymbol\eta}
\bigg]
 \diff{\boldsymbol\theta}
\Bigg)
\nonumber
\\
& = &
\zeta \Bigg( 
\int_{\boldsymbol\theta \in {\varTheta}} \pi(\boldsymbol\theta) \cdot 
\bigg[   
\sum_{i = 1}^L v_i \cdot 
\int_{\boldsymbol\eta \in {\varTheta}}
\pi_i(\boldsymbol\eta)
\cdot k(\boldsymbol\theta, \boldsymbol\eta) \diff{\boldsymbol\eta}
\bigg]
 \diff{\boldsymbol\theta}
\Bigg)
\\
& = &
\zeta \Bigg( 
\int_{\boldsymbol\theta \in {\varTheta}} \pi(\boldsymbol\theta) \cdot 
\underbrace{
\bigg[   
\sum_{i = 1}^L v_i \cdot 
\tilde \pi_i(\boldsymbol\theta)
\bigg]}_{\Upsilon_{\mbox{\tiny KME}}(\boldsymbol\theta; \v)} \diff{\boldsymbol\theta}
\Bigg)
\label{eq:f_mke}
\\
& =&
\zeta \Bigg(   \mathbb{E}_{\pi}  \bigg[ \Upsilon_{\mbox{\tiny KME}}(\boldsymbol\theta; \v) \bigg]   \Bigg)
\label{eq:c_mke}
\end{eqnarray}
}
where $\tilde \pi_i$ are kernel-smoothed posteriors $\pi_i$ and $\Upsilon_{\mbox{\tiny KME}}(\boldsymbol\theta; \v)$ is the function to be learnt;
\item Probabilistic classifier proposed in this work (Eq.~\ref{our_classifier} ):
\begin{eqnarray}
p(c = 1 | \pi)  
& = & 
\int_{{\varTheta}} \pi(\boldsymbol\theta) \cdot  
\zeta \Bigg(
\underbrace{ \sum_{i = 1}^m w_i \cdot k(\boldsymbol\theta, \boldsymbol\theta^{\cal F}_i) }_{\Upsilon_{\mbox{\tiny LiMS}}(\boldsymbol\theta; \w)} 
\Bigg)
 \diff{\boldsymbol\theta} 
\label{eq:f_kr}
\\
& = &
\mathbb{E}_{\pi} \bigg[ \zeta( \Upsilon_{\mbox{\tiny LiMS}}(\boldsymbol\theta; \w)) \bigg] 
\label{eq:c_kr}
\end{eqnarray}
where $\Upsilon_{\mbox{\tiny LiMS}}(\boldsymbol\theta; \w)$ is learnt by adjusting the free parameter $\w$.
\end{itemize}

To see a deeper connection between the three classifiers above, consider first the usual setting of kernel logistic regression,
\begin{equation}
p(c = 1 | \boldsymbol\theta) =    \zeta( \Upsilon(\boldsymbol\theta)).
\label{eq:kr}
\end{equation}
This can be interpreted as follows: The model imposes a smooth field (natural parameter of Bernoulli distribution) $\Upsilon(\boldsymbol\theta)$ over the inputs $\boldsymbol\theta$.

The field assigns to each input a real number that expresses the `strength' with which that particular input wants to belong to class +1. Pushing the field through the link function $\zeta$ creates a new field $ \zeta( \Upsilon(\boldsymbol\theta))$ over the inputs, assigning to each $\boldsymbol\theta$ the probability with which it belongs to class +1. 

In case our inputs are not individual models $\boldsymbol\theta$, but  (posterior) distributions $\pi$ over the models, the classifier (\ref{eq:kr}) can be generalized in two ways:
\begin{enumerate}
\item
Use the posterior distribution $\pi$ to average over individual natural parameters $\Upsilon(\boldsymbol\theta)$ to create the overall mean natural parameter $\mathbb{E}_{\pi}  [ \Upsilon(\boldsymbol\theta) ]$. This can then be passed through the link function $\zeta$ to calculate the class +1  probability for $\pi$, $\zeta(\mathbb{E}_{\pi}  [ \Upsilon(\boldsymbol\theta) ])$. This scenario can be described as forming an (infinite) ensemble to form the overall opinion about the strength of $\pi$ belonging to class +1 and only then turning it into the class probability. This option is taken by the classifiers based on Probability Product Kernel and Kernel Mean Embedding, (\ref{eq:c_ppk}) and (\ref{eq:c_mke}), respectively 
\item
Use the posterior distribution $\pi$ to average over individual class probabilities $\zeta(\Upsilon(\boldsymbol\theta))$ to form the overall class probability $\mathbb{E}_{\pi}  [\zeta( \Upsilon(\boldsymbol\theta))]$ of $\pi$. This corresponds to creating an ensemble of probabilistic classifiers $\zeta( \Upsilon(\boldsymbol\theta))$ acting on individual models $\boldsymbol\theta$, as done by the proposed classifier (see (\ref{eq:c_kr})).
\end{enumerate}

One can view the latter approach $\mathbb{E}_{\pi}  [\zeta (\Upsilon(\boldsymbol\theta)) ]$ as a regularization of the former one $\zeta(\mathbb{E}_{\pi}  [ \Upsilon(\boldsymbol\theta) ])$. Loosely speaking, when collecting ensemble votes to form an opinion about the probability of class +1 given $\pi$, $\mathbb{E}_{\pi}  [\zeta( \Upsilon(\boldsymbol\theta))]$ ignores the (potentially huge) differences between individual natural parameters $ \Upsilon(\boldsymbol\theta))$ giving negligible differences in the probabilities $ \zeta(\Upsilon(\boldsymbol\theta)))$ because of the saturation regions at both extremes of the link function $\zeta$. This effectively collapses input regions of models $\boldsymbol\theta$ with high positive field values into a single high class probability region. Analogously, regions of  models $\boldsymbol\theta$ with low negative values will be identified into  a low class probability region.

Another point of view is to compare the models for the field $\Upsilon(\boldsymbol\theta)$ utilized in the three classifiers. In all cases the fields are modelled as linear combinations of basis functions. Because kernels of the classifiers based on Probability Product Kernel and Kernel Mean Embedding operate on full distributions, the basis functions for modelling the field $\Upsilon(\boldsymbol\theta)$ are the (possibly tempered) training posterior distributions $\pi_i^\alpha$ or their kernel-smoothed versions $\tilde \pi_i$, respectively (see (\ref{eq:f_ppk}) and (\ref{eq:f_mke}). In contrast, the proposed classifier (\ref{eq:c_kr}) models the field $\Upsilon(\boldsymbol\theta)$ in a less constrained framework of kernel regression as a linear combination of kernel basis functions $k(\cdot, \boldsymbol\theta^{\cal F}_i)$ (see (\ref{eq:f_kr})). In particular, no assumption is made that the field should lie in the span of the training distributions $\pi_i^\alpha$ or their smoothed versions $\tilde \pi_i$.

\section{Testbeds}

\label{Sect:TestBeds}

In this work, we validate our general framework using two example dynamical systems: Gonadotropin-Releasing Hormone Signalling model (GnRH)~\citep{TAMCAM2012} and stochastic double-well systems (SDW)~\citep{PT2001}. GnRH is an example of ordinary differential equation (ODE) systems and SDW is an example of stochastic differential equation (SDE). GnRH is also an example of biological pathway/compartment model.

\subsection{GnRH signalling model}

\label{sec:GnRH}

\begin{figure}[ht]
\centering
\begin{minipage}[b]{0.45\linewidth}
\includegraphics[width=6cm,height=6cm]{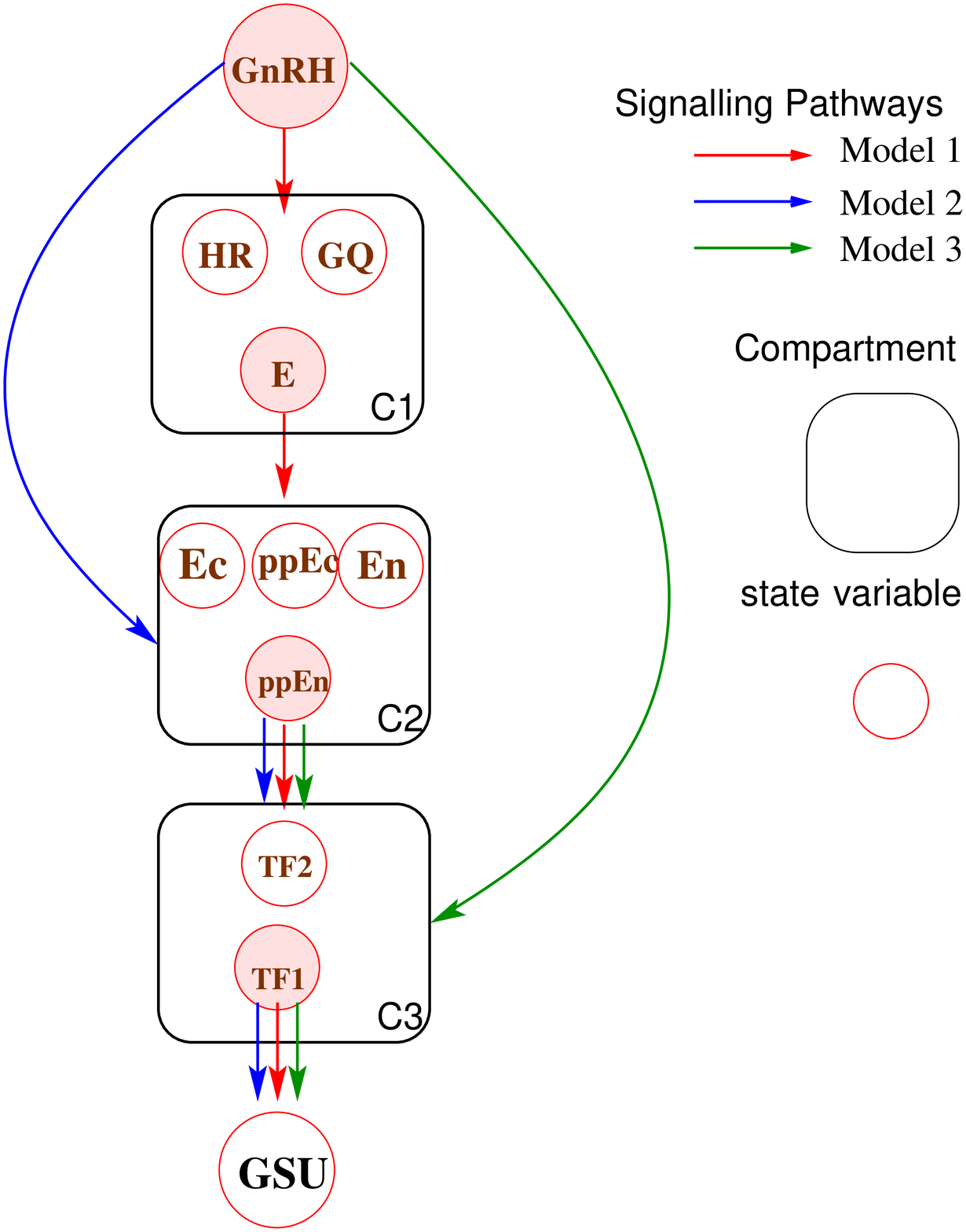}
\end{minipage}
\quad
\begin{minipage}[b]{0.45\linewidth}
\includegraphics[width=5cm,height=6cm]{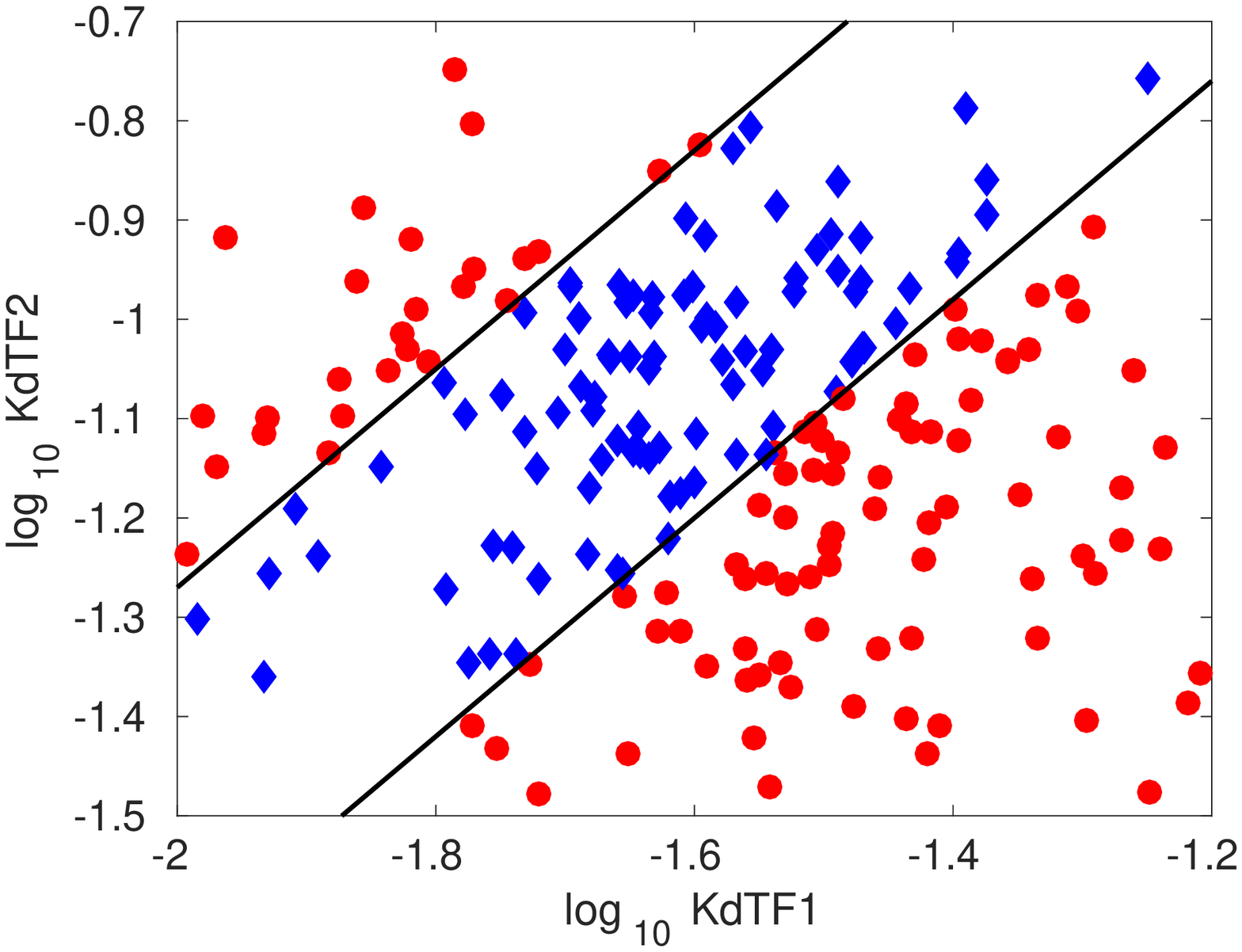}
\end{minipage}
\caption{
Left panel: schematical representation of three nested GnRH signalling models ($M1$, $M2$, and $M3$). The signalling pathway of these models is highlighted by the flow of red, blue and green arrows, respectively. Each model comprises of GnRH signal as the driving input, GSU as the measurable output, and one to three compartments along its signalling pathway. Right panel: Two classes of GnRH signalling models: (1) Class of normal subjects with bell-shaped frequency-response relationship (Blue Diamonds) and (2) Class of abnormal subjects with simple frequency dependency of response (Red Disks). These two classes are separated by two straight lines in the log-log parameter space.
}
\label{GnRH_model}
\end{figure}

Mathematically, GnRH signalling model is an ODE system with 11 state variables. These state-variables include concentrations of gonadotropin releasing  hormones ([GnRH]) and  gonadotropin hormones ([GSU]) as the driving input and measurable output, respectively, of this model. The remaining state variables can be grouped into three compartments along the signalling pathway: 
(1) C1 for GnRH binding process;  
(2) C2 for extracellular signal regulated kinase (ERK) activation; and 
(3) C3 for transcription factor (TF) activation. 
We refer to this model as $M_1$ and consider it as the full model in a hierarchy of three nested GnRH signalling models. This hierarchy is schematically illustrated in Figure~\ref{GnRH_model}. We highlight the signalling pathway in {\it M1} by red arrows. By removing the compartment C2 from the pathway, we obtain a two-compartment model denoted by {\it M2}. When we further
remove C3 from the pathway, {\it M2} is reduced to {\it M3} in which GnRH signals directly modulate stimulation of transcriptional activation. The pathways of {\it M2} and {\it M3} are highlighted in  Figure~\ref{GnRH_model} by blue and green arrows, respectively.

GnRH signal is the chemical signal which stimulates the reproductive endocrine system. This signal is modeled by 
\begin{equation}
\frac{\diff \mbox{[GnRH]}}{\diff t} = - \mbox{[GnRH]} + p_{\mbox{\tiny \bf GnRH}} \cdot \Big\{ H\big(t \ \mbox{\bf \footnotesize mod} \ f^{-1} \big) - 
H\big(
(t \ \mbox{\bf \footnotesize mod} \ f^{-1})-t_p\big)
\Big\},
\label{GnRH_input}
\end{equation}
where $p_{\mbox{\tiny \bf GnRH}}$ is the GnRH pulse magnitude, $f$ is the pulse frequency and $t_p$ is the pulse duration. In this work, we set $p_{\mbox{\tiny \bf GnRH}}$ to be a constant (i.e. $p_{\mbox{\tiny \bf GnRH}}$ = 0.1) and treat both $f$ and $t_p$ as model parameters.

The amount of $\mbox{TF}_1$ and $\mbox{TF}_{2}$, denoted by $[\mbox{TF}_1]$ and $[\mbox{TF}_{2}]$, are two state variables in C3 which modulate the  dynamics of GSU expression as follows:
{\footnotesize
\begin{equation}
\frac{\diff {[\mbox{GSU}]}}{\diff t} =
{K_{\mbox{complex}}} \cdot
\left(
\frac{
\frac{{ [\mbox{TF}_1]}}{{K_{d_{\mbox{\tiny TF}_1}} }}
\cdot
\frac{{ [\mbox{TF}_{2}]}}{{ K_{d_{\mbox{\tiny TF}_2}}}}
\cdot
{ [DNA_{\mbox{TOT}}]}^2
}
{\left( 
1
+
\frac{{ [\mbox{TF}_1]}}{{ K_{d_{\mbox{\tiny TF}_1}}}}
+
\frac{{ [\mbox{TF}_{2}]}}{{ K_{d_{\mbox{\tiny TF}_2}}}}
\right)^2}
\right) -  d_{[\mbox{GSU}]} \cdot {[\mbox{GSU}]}
\label{GnRH_output}
\end{equation}
} 
where $K_{d_{\mbox{\tiny TF}_1}}$ and $K_{d_{\mbox{\tiny TF}_2}}$ are the dissociation constants of $[\mbox{TF}_1]$ and $[\mbox{TF}_2]$, respectively. They are both considered as model parameters. The remaining model parameters are set values reported in the literature~\citep{TAMCAM2012}. In summary, the GnRH signalling model has one observable and four free model parameters. The observable is GSU and the model parameters are: GnRH pulse frequency $f$, GnRH pulse duration $t_p$, the dissociation constant of $[\mbox{TF}_1]$, $K_{d_{\mbox{\tiny TF}_1}}$, and the dissociation constant of $[\mbox{TF}_2]$, $K_{d_{\mbox{\tiny TF}_2}}$.
  
It is widely accepted that the reproductive system is controlled via GnRH pulse frequency. This frequency varies under different physiological conditions, affecting the transcription of GSU and secretion of reproductive hormones that are crucial for the physiology of the reproductive system. GnRH frequency decoding mechanisms vary under normal and pathological conditions, but two main possibilities exist: {\bf(1)} Increasing pulse frequency simply increases output (GSU) until a maximal response is maintained with continuous stimulation (see {\it Figure 6 Panel a} in~\citep{TAMCAM2012}); and  {\bf(2)} Pulsatile stimuli may elicit maximal responses at sub-maximal frequencies, generating bell-shaped frequency-response relationship (see {\it Figure 6 Panel b} in~\citep{TAMCAM2012}). In this work, we utilise these two mechanisms to define two classes of subjects: "abnormal" (mechanism (1)) and "normal" (mechanism (2)) subjects). As these two classes differ in how they respond to a change in  pulse frequency, it is not sufficient to represent individual subjects by a single GnRH mode. Instead, every subject needs to be represented by an ensemble of GnRH models with different frequencies that adequately cover the entire permissible range. In this work, we define such an ensemble with six different pulse frequencies: $f_1 = \frac{1}{8}$, $f_2 = \frac{1}{4}$, $f_3 = \frac{1}{2}$, $f_4 = 1$, $f_5 = 2$, and $f_6 = 4$. For a given model setting {\footnotesize $\big( K_{d_{\mbox{\tiny TF}_1}}, K_{d_{\mbox{\tiny TF}_2}}, t_p \big)$} we thus have an ensemble of 6 models {\footnotesize $\big( K_{d_{\mbox{\tiny TF}_1}}, K_{d_{\mbox{\tiny TF}_2}}, t_p, f_i \big)$}, $i=1,2,...,6$. Further, the measurable output of this ensemble model is $([\mbox{GSU}]_1, ..., [\mbox{GSU}]_6)^{\intercal}$ where the flow $[\mbox{GSU}]_i$ is the output of the $i$th ensemble member.

It has been shown that the frequency-response behaviour of GnRH models is determined by $K_{d_{\mbox{\tiny TF}_1}}$ and $K_{d_{\mbox{\tiny TF}_2}}$, but not by  $t_p$. The right panel of Figure~\ref{GnRH_model} shows that in the space of ($\log K_{d_{\mbox{\tiny TF}_1}}$, $\log K_{d_{\mbox{\tiny TF}_2}}$), there exist three linearly separated domains in which only one of two frequency-response behaviours (linear or bell-shaped) is observed. The domain in the middle represents the normal subjects, whereas both remaining domains represent the abnormal subjects. 

\subsection{Stochastic Double-well Systems}

\label{sec:SDW}

\begin{figure}[ht]
\centering
\begin{minipage}[b]{0.45\linewidth}
\includegraphics[width=6.5cm,height=6.5cm]{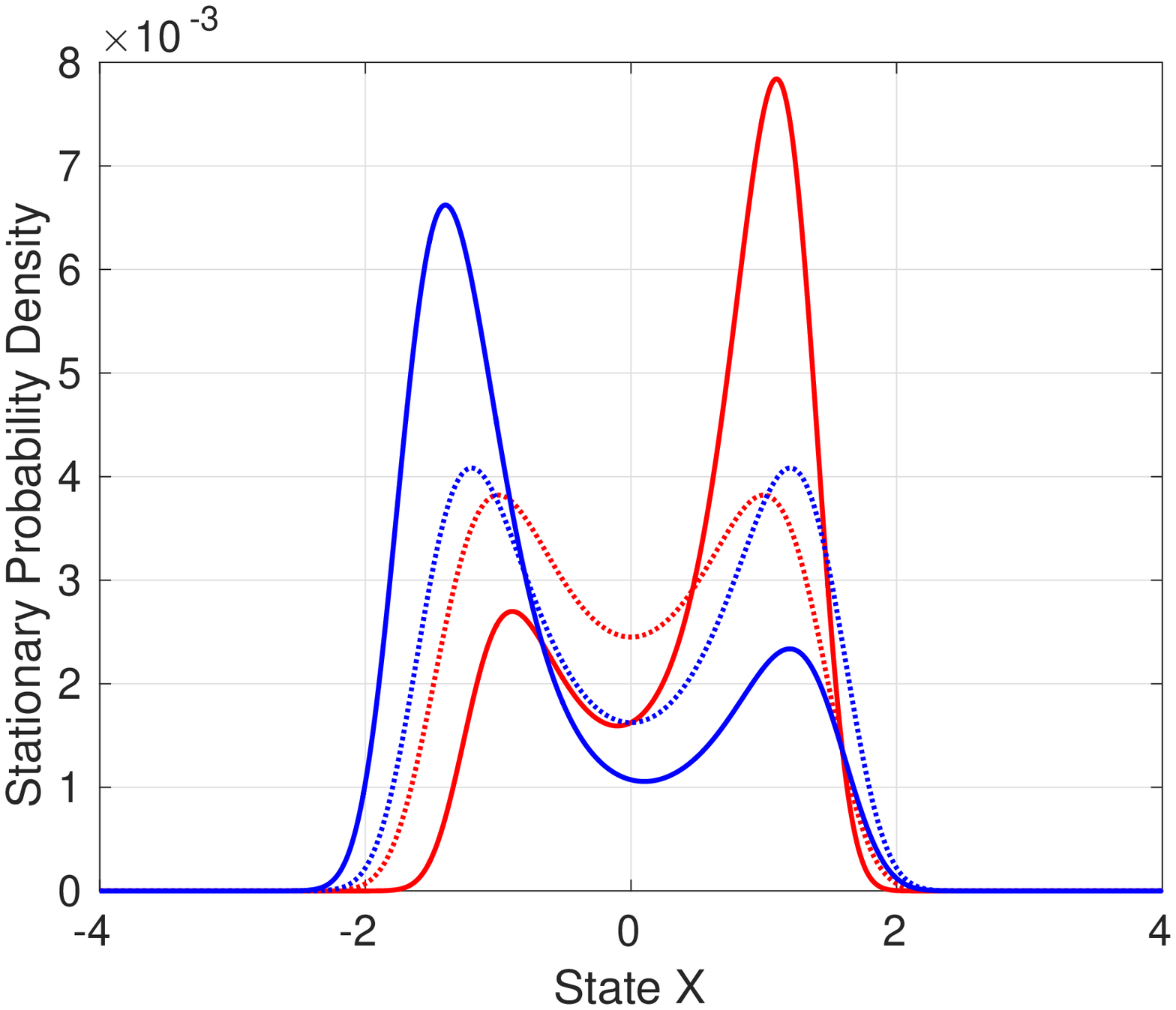}
\end{minipage}
\quad
\begin{minipage}[b]{0.45\linewidth}
\includegraphics[width=6.5cm,height=6.5cm]{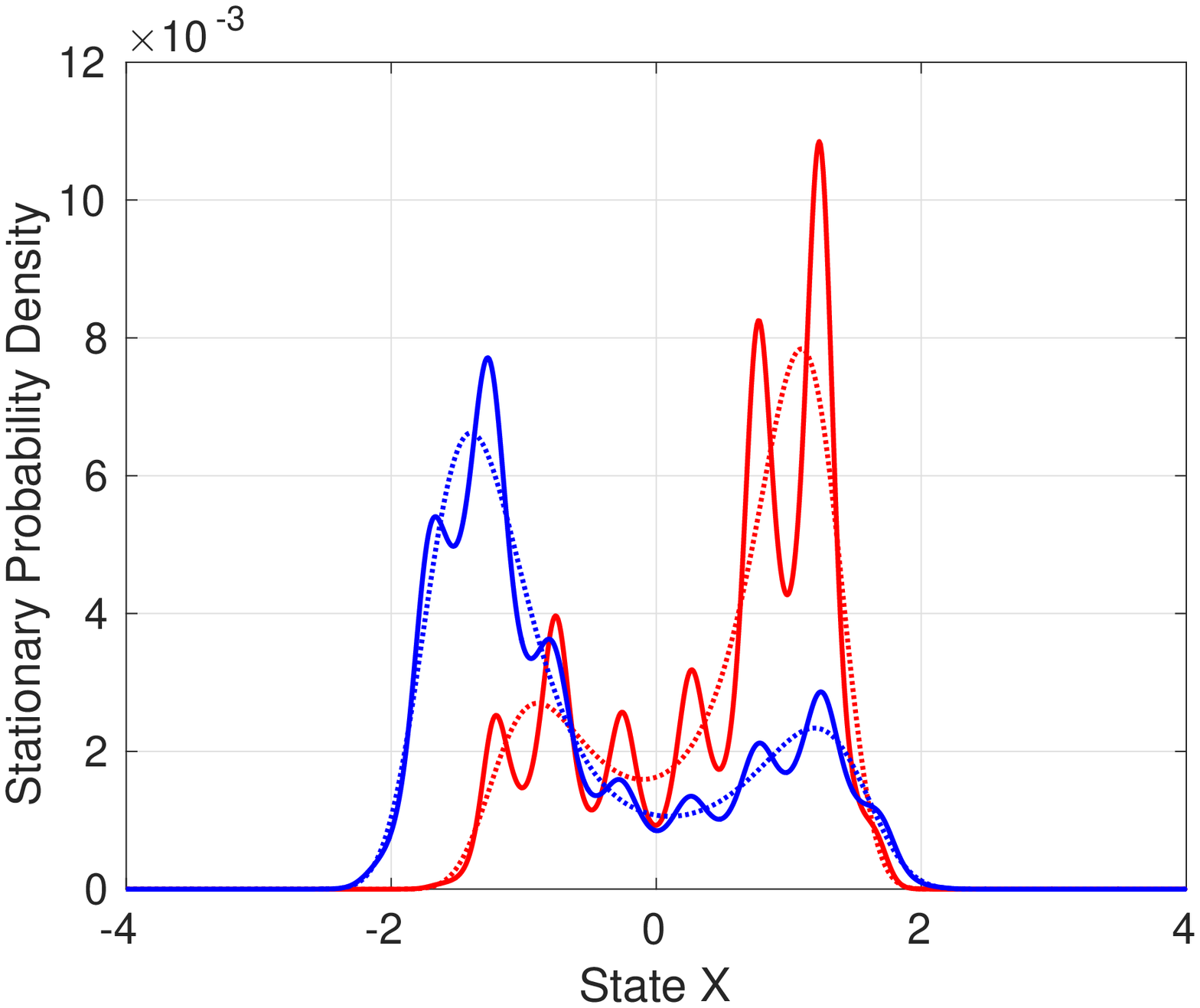}
\end{minipage}
\caption{Left Panel: Equilibrium probability distribution of states $x$ for four example Stochastic Double-well Systems with 
($d$, $\kappa$, $a$) = (1.0, 1.0, 0.1) (red solid curve),
($d$, $\kappa$, $a$) = (1.3, 1.5, -0.1) (blue solid curve),
($d$, $\kappa$, $a$) = (1.0, 1.5, 0) (red dotted curve), and
($d$, $\kappa$, $a$) = (1.2, 1.5, 0) (blue dotted curve). 
Right Panel: The same as in 
in Right Panel but for Stochastic Multi-well Systems.
}
\label{fig:pgdw}
\end{figure}

Stochastic Double-Well (SDW) system is mathematically defined as
\begin{equation}
\diff x_t = \underbrace{4(x_t - a)(d^2 - x_t^2)}_{f(x_t)} + \kappa^2 \cdot \diff b_t,
\label{gdw}
\end{equation}
where $b_t$ represents the univariate standard Brownian motion and $\boldsymbol\theta= (d, \kappa, a)$ collects the three model parameters, namely the well location parameter $d$, well asymmetry parameter $a$ and standard deviation $\kappa$ of the dynamical noise. Eq.~\ref{gdw} shows that the drift term $f(x_t)$ is not explicitly time-dependent. Therefore, the underlying dynamics is governed by the potential $u(x)$ with $f(x) = - \nabla_x u(x)$. Moreover, the equilibrium probability distribution of its state $x$ is given by $p^{\mbox{\tiny \bf eq}}(x) \propto \exp(-\frac{u(x)}{\kappa^2})$~\citep{Honerkamp}. The potential corresponding to Eq.~\ref{gdw} is given by 
\[
u(x) = x^4 - \frac{4}{3}ax^3 - 2 d^2 x^2 + 4 a d^2 x.
\]
 The equilibrium probability distribution of two example SDWs is shown in the left panel of Fig.~\ref{fig:pgdw}. We can see that there exist two meta-stable states located at $x$ = $d$ and $x$ = $-d$. The larger is the dynamical noise variance, $\kappa^2$, the more frequent are the transitions from one meta-stable state to the other. Figure~\ref{fig:pgdw} shows that the peak probability for $\kappa$ = 1.0 (red solid curve) is larger than that for $\kappa$ = 1.5 (blue solid curve). For positive well asymmetry parameter $a$,  the transition from $x$ = $-d$ to $x$ = $d$ is more likely than the transition in the opposite direction. As a result, the equilibrium probability at $x$ = $d$ is higher than that at $x$ = $-d$ (see red solid curve in Figure~\ref{fig:pgdw}). Analogously, the equilibrium probability at $x$ = $-d$ is higher than that at $x$ = $d$ for negative  well asymmetry parameter (see blue solid curve in Figure~\ref{fig:pgdw}). The dynamics of double-well systems is dominated by switching between the two wells. We also study more complex multi-well systems where the potential has more than two wells. An example of such a multi-well system dominated by an overall two-well structure  (wells in positive range of $x$ are generally deeper than those in the negative range (or vice-versa)) is given below (see also right panel of Figure~\ref{fig:pgdw}):
\begin{equation}
\diff x_t = - \nabla_{x} \tilde u(x) +  \kappa \cdot \diff w
\quad
\mbox{with}
\quad
\tilde u(x) = u(x) + \frac{1}{2} \cdot \cos(4\pi x),
\label{pgdw}
\end{equation}
where $\tilde u$ denotes the perturbed potential. 

\begin{figure}[t]
\centering
\begin{minipage}[b]{0.45\linewidth}
\includegraphics[width=7cm,height=10.5cm]{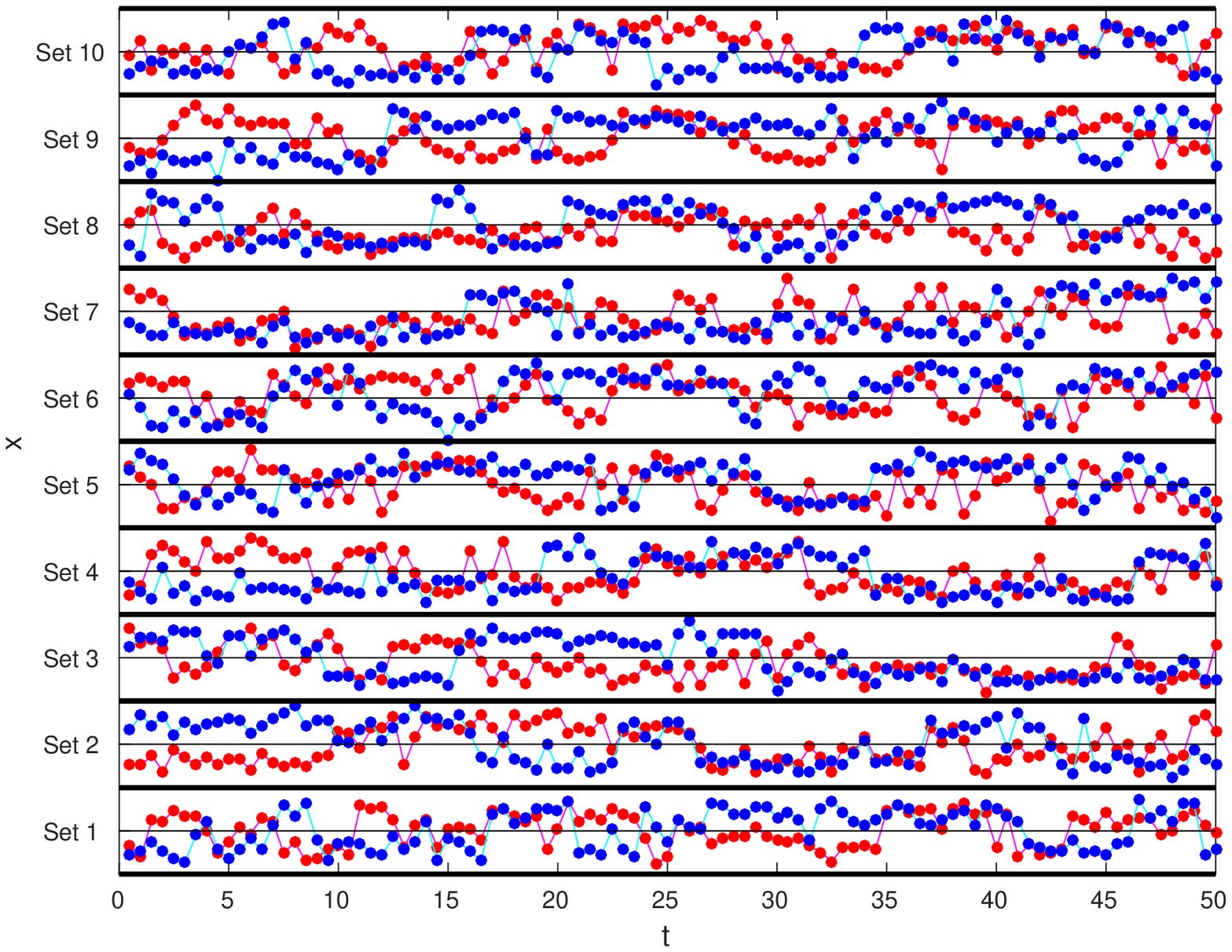}
\end{minipage}
\quad
\begin{minipage}[b]{0.45\linewidth}
\includegraphics[width=7cm,height=10.5cm]{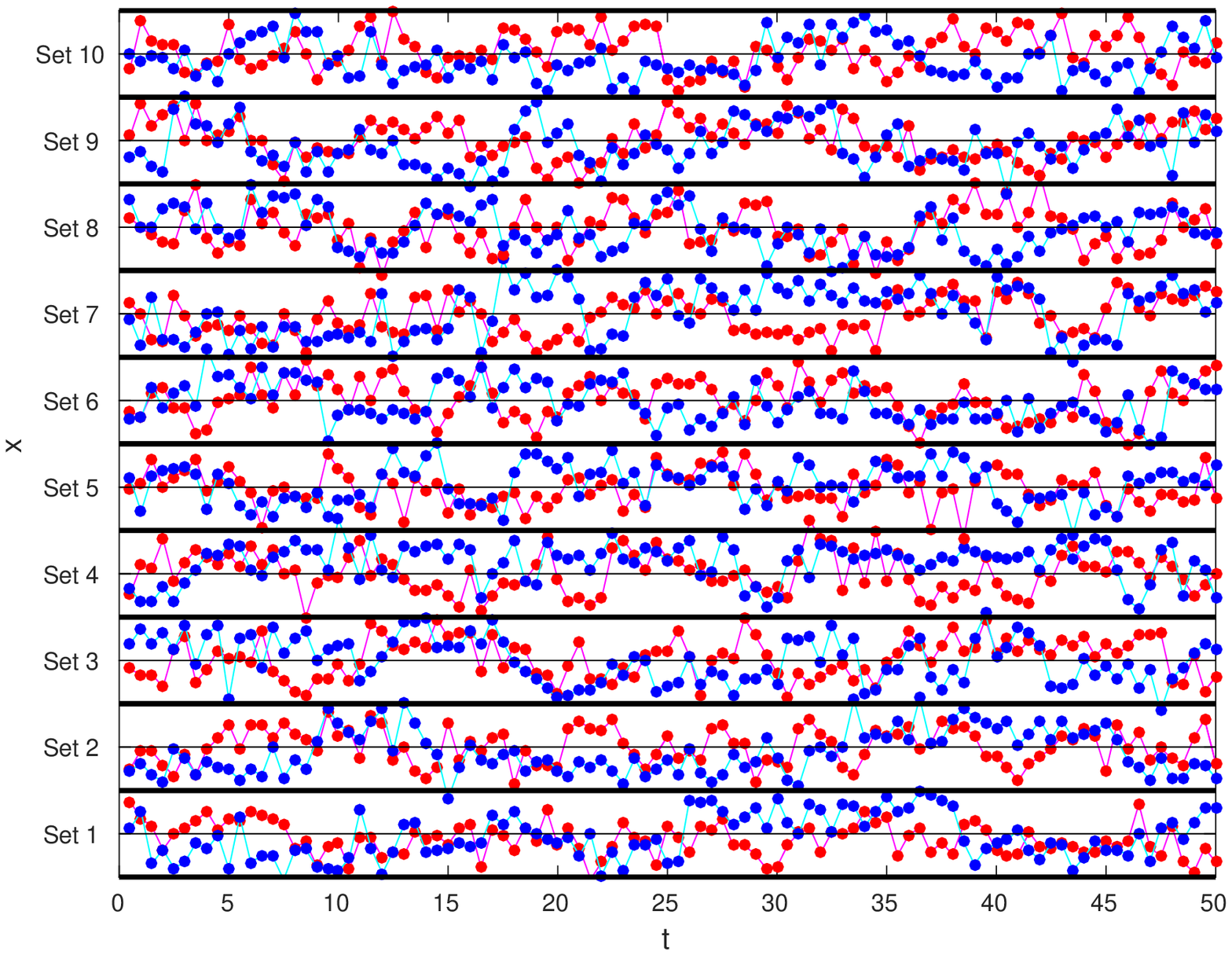}
\end{minipage}
\caption{Left Panel: The observed trajectories of ten example stochastic double-well systems from each of two clusters in Figure~\ref{fig:pgdw} (red vs. blue solid curves). The range of each sub-panel's vertical axis is scaled to [$-$2.5, $+$2.5]. The inter-sample interval ($ISI$) is 0.5, and the variance $\sigma^2$ of Gaussian distributed observation noise is 0.04. Right Panel: The same as in Left Panel but for $\sigma^2$ = 0.36.}
\label{pgdw_cluster_trajectory_new}
\end{figure}

In this work, we formed two classes of SDWs through two class-conditional Gaussian distributions in the parameter space as follows: ($\bar d_1$ + $\epsilon_d$, $\bar \kappa_1$ + $\epsilon_{\kappa}$, $\bar a_1$) for {\it Class 1} and ($\bar d_0$ + $\epsilon_d$,  $\bar \kappa_0$ + $\epsilon_{\kappa}$, $\bar a_0$) for {\it Class 0}, where ($\bar d_1$, $\bar \kappa_1$, $\bar a_1$) and ($\bar d_0$,  $\bar \kappa_0$, $\bar a_0$) denote the class-conditional prototypical model parameter; $\epsilon_d$ and $\epsilon_{\kappa}$ are Gaussian-distributed zero-mean random variables with standard deviations 0.1/3 and 0.05/3, respectively. An example of such two classes of SDWs is defined by ($\bar d_1$, $\bar \kappa_1$, $\bar a_1$) = (1.3, 1.5, $-$0.1) and ($\bar d_0$, $\bar \kappa_0$, $\bar a_0$) = (1.0, 1.0, 0.1) corresponding to the blue and red solid curves in the left panel of Figure~\ref{fig:pgdw}, respectively. It is more likely for the trajectories from \textit{Class 0} to stay above, rather than below, the horizontal line with $\x_t = 0$. The opposite holds for \textit{Class 1}. This is because the asymmetry parameters of these two classes take their values with opposite signs.  As a result, the classification task can be well accomplished by a classifier based on simple features directly extracted from the signal - in this case the overall trajectory mean. Figure~\ref{pgdw_cluster_trajectory_new} illustrates a contrasting task in which two classes of SDWs are defined by ($\bar d_1$, $\bar \kappa_1$, $\bar a_1$) = (1.2, 1.5, 0)  and ($\bar d_0$, $\bar \kappa_0$, $\bar a_0$) = (1.0, 1.5, 0) (see the blue and red dotted curves, respectively, in the left panel of Figure~\ref{fig:pgdw}). As the mean asymmetry parameter is set to zero for both classes, the overall trajectory mean fluctuates around zero across the trajectories in each of these two classes. We thus hypothesise that in such cases, the proposed classification LiMS framework will be superior to classification based on direct signal based features.

\section{Experiments}

\label{Sect:Experiments}

\subsection{General Issues}

\label{sec:issues}
In the experiments we evaluate performance of the three classifiers, namely the proposed classifier (LiMS) and two well-established distributional classifiers based on {Probability Product Kernel} (PPK) and {Kernel Mean Embedding} (KME), on two classes of dynamical systems, one representant of ODE (GnRH, section \ref{sec:GnRH}), the other of SDE (SDW, section \ref{sec:SDW}). For a fair comparison all three classifiers were implemented in the framework of Kernel Logistic Regression. Our study addressed two important issues for classifying partially observed dynamical systems (PODS): 
\begin{enumerate}
\item
{\it The influence of model uncertainty on classification in the model space.}\\
Model uncertainty arises when the underlying system is not completely observed. It is represented through posterior distribution over the underlying dynamical systems inferred from the partial observations. It is natural to expect that the posterior over possible models, given the observations, is a better (model space) representation of the observed time series than a single model, e.g. MAP point estimate. It is also natural to expect that the classification performance will increase with reducing model uncertainty. We compare the LiMS, PPK, and KME classifiers in terms of capability to deal with increased levels of model uncertainty quantified through entropy of the posterior distributions. We also use the level of observation noise $\sigma$, or the number of observations $n$ as surrogate uncertainty measures.
\item
{\it Performance degradation when the  model class used to represent the observed time series through posterior distributions over it is a reduced sub-model class of the true model class  
generating the training and test data.} \\
There can be several reasons for the inferential model to be different from the underlying data generating model. For example, in real-world applications, it is inevitable that there is a gap between the real-world and the mathematical model developed to account for it. Alternatively, while the given mathematical model can be considered adequate, it is too complex  and computationally expensive to simulate. To circumvent this problem, a reduced model could be used to represent time series, as long as it captures characteristics relevant for the given classification task. We compare the classification performance between different inferential models ranging from the full, multiple-compartment pathway ODE model to the trivial single compartment model. Analogous experiments were performed in the SDE case - SDW models representing data generated by stochastic multi-well systems.
\end{enumerate}

\subsection{Practical Issues}

In this section we discuss a number of practical issues related to testing the LiMS, KME and PPK classifiers: 
\begin{itemize}
\item {\it Does the input of a distributional classifier need to be normalised?}\\
For the task of classifying PODS, the actual input is the posterior distribution over parameter vectors. In our setting, it includes a set of posterior probabilities defined on a grid of parameter vectors. For PPK classifiers, only those probabilities are used and thus there is no need for normalisation. For the other two classifiers, however, we use parameter vectors (on the grid) together with the corresponding posterior probabilities. Moreover, the parameter vectors are involved in the classification via a spherical kernel function that is defined on the product of two parameter grids. Therefore, we normalise the parameter grid to vary in each dimension from 0 to 1. Of course, the original parameter values associated with grid points will be preserved.
\item {\it How to initialise the classifier's parameters for gradient-based training?}\\
We implement all three classifiers in the KLR framework. Hence, the PPK-based classifier parameter effectively weights the training examples, whereas in the case of LiMS and KME, the parameter puts weights on the model grid. In this work, all elements of the parameter vectors are initialised by drawing from Gaussian distribution with zero mean and unit variance. The parameters are then optimized through gradient descent as explained in Section~\ref{sec:klr}. This procedure is repeated $N^{\text{init}}$ times, resulting in $N^{\text{init}}$ classifiers combined in flat ensemble outputting the average of the $N^{\text{init}}$ predictive class probabilities (given a test input). We set $N^{\text{init}}=15$.
\item For the binary classification tasks in this work, we first generate the training and hold-out test sets with balanced class distribution, each containing 200 observation time series. Both classes from the training set are randomly sub-sampled (without replacement) to 45 time series (out of 100), yielding a training batch of 90 time series. This is repeated $N^{\text{rand}}=10$ times. We then report the mean ($\pm$std. deviation) classification performance on the test set across the $N^{\text{rand}}$ runs.
\end{itemize}

\subsection{GnRH Signalling Model}

\label{sec:GnRH_simple}

To conduct experiments with the classification task defined in Section~\ref{sec:GnRH}, we generate two independent sets of GnRH models for training and testing (200 labelled models each). To that end we randomly sample 400 parameter vectors $\boldsymbol\theta_{\mbox{\tiny \bf GnRH}} = \big(\log K_{d_{\mbox{\tiny TF}_1}}, \log K_{d_{\mbox{\tiny TF}_2}}, t_p \big)$  of the GnRH model\footnote{We consider log values of  $K_{d_{\mbox{\tiny TF}_1}}$ and $K_{d_{\mbox{\tiny TF}_2}}$ since their permissible range extends over several magnitudes.}. Each of the three model parameters are sampled from the corresponding Gaussian distribution truncated to the permissible range. For each parameter, the mean and standard deviation of the untruncated Gaussian are set to the mid-point and radius, respectively of the permissible range (see Table~\ref{GnRH_model_par}). The parameter vectors are then labelled as {\it Class 0} (normal conditions) or  {\it Class 1} (abnormal conditions) as described in Section~\ref{sec:GnRH} (see the right panel of Figure~\ref{GnRH_model}). 
 
\begin{table}
\centering
\begin{tabular}{| c |  l | l | l | l |}
\hline
Parameter                        & Mean & Variance & Lower bound & Upper bound \\ \hline
$\log K_{d_{\mbox{\tiny TF}_1}}$ & -1.6 & 0.2      & -2.0        & 0.2         \\ \hline
$\log K_{d_{\mbox{\tiny TF}_2}}$ & -1.1 & 0.2      & -1.5        & 0.2          \\ \hline
$t_p$                            & 7.5  & 0.8333   &  5          & 10           \\ \hline
\end{tabular} 
\caption{The truncated Gaussian distributions of three GnRH model parameters (i.e. $\log K_{d_{\mbox{\tiny TF}_1}}$, $\log K_{d_{\mbox{\tiny TF}_2}}$ and $t_p$) used for generating the training and testing set of GnRH models. \label{GnRH_model_par}}
\end{table}

As the task is to classify PODS, we generate a variety of observation time series with different observation settings (number of observations, observation times and observational noise level). To sample observations from the GnRH model we first simulate GnRH (8-hour window) and record the observable trajectory [GSU] at six different pulse frequencies. This results in a six-dimensional [GSU] trajectory with a time resolution of 1 minute. Throughout the experiments, the initial values of state variables in GnRH model are fixed but the trajectory over the first half an hour is discarded. This ensures that the transient behaviour has been ignored and only the attractor part of individual trajectories is used for sampling observations and thus the initialisation of the GnRH model has little influence on inferring the underlying model from observations~\citep{TAMCAM2012}. Given a simulated [GSU] trajectory,  we generate 15 observation sets using different pairs of observation noise level $\sigma$ and the inter-sample interval ($ISI$). The observation sets are organised in three groups (5 sets in each group):
\begin{description}
\item[{\it Group 1:}] In each of the 5 observation sets, observations were sampled regularly every $ISI=75$ minutes over 7.5 hours, yielding 6 observation times. The level $\sigma$ of observation noise in the 5 observation sets was set to 0.1, 0.03, 0.01, 0.005 and 0.001. Hence, the observation sets in this group correspond to the partially observed GnRH model with five different levels of model uncertainty controlled by $\sigma$. 
\item[{\it Group 2:}] Unlike in {\it Group 1}, the 5 observation sets in this group are generated by fixing the observation noise to $\sigma=0.03$ and varying the number of regularly spaced observation times within the 7.5 hour window. In particular, the 5 observation sets contained 5, 6, 10, 15 and 30 observation times with $ISI$ = 90, 75, 45, 30 and 15, respectively. In this case, the model uncertainty is controlled by the sparsity of observations. 
\item[{\it Group 3:}] 
The $\sigma$- and $ISI$-values are as same as in {\it Group 2}, but the observation times are placed randomly with uniform distribution over the 7.5 hour window.
\end{description}

In order to apply a distributional classifier for the classification task, each observation set is represented by the corresponding posterior distribution over the GnRH models. Recall that in this work we approximate posteriors on finite grid: $\log_{10} \big(K_{d_{\mbox{\tiny TF}_1}}\big) = -2.3 + 1.4 \cdot \frac{i}{41}$, $\log_{10} \big(K_{d_{\mbox{\tiny TF}_2}}\big) = -2.1 + 1.4 \cdot \frac{i}{41}$, $i=0, 1, 2, ..., 40, 41$ and $t_p \in \{5, 6, 7, 8, 9, 10\}$. The finite-grid encodes our prior knowledge about the biologically permissible parameter ranges. As the classes are discriminated by $K_{d_{\mbox{\tiny TF}_1}}$ and $K_{d_{\mbox{\tiny TF}_2}}$, the inferred posteriors are marginalised over $t_p$.

\begin{figure}[!ht]
\centering
\begin{minipage}[b]{0.45\linewidth}
\includegraphics[width=5.5cm,height=6.25cm]{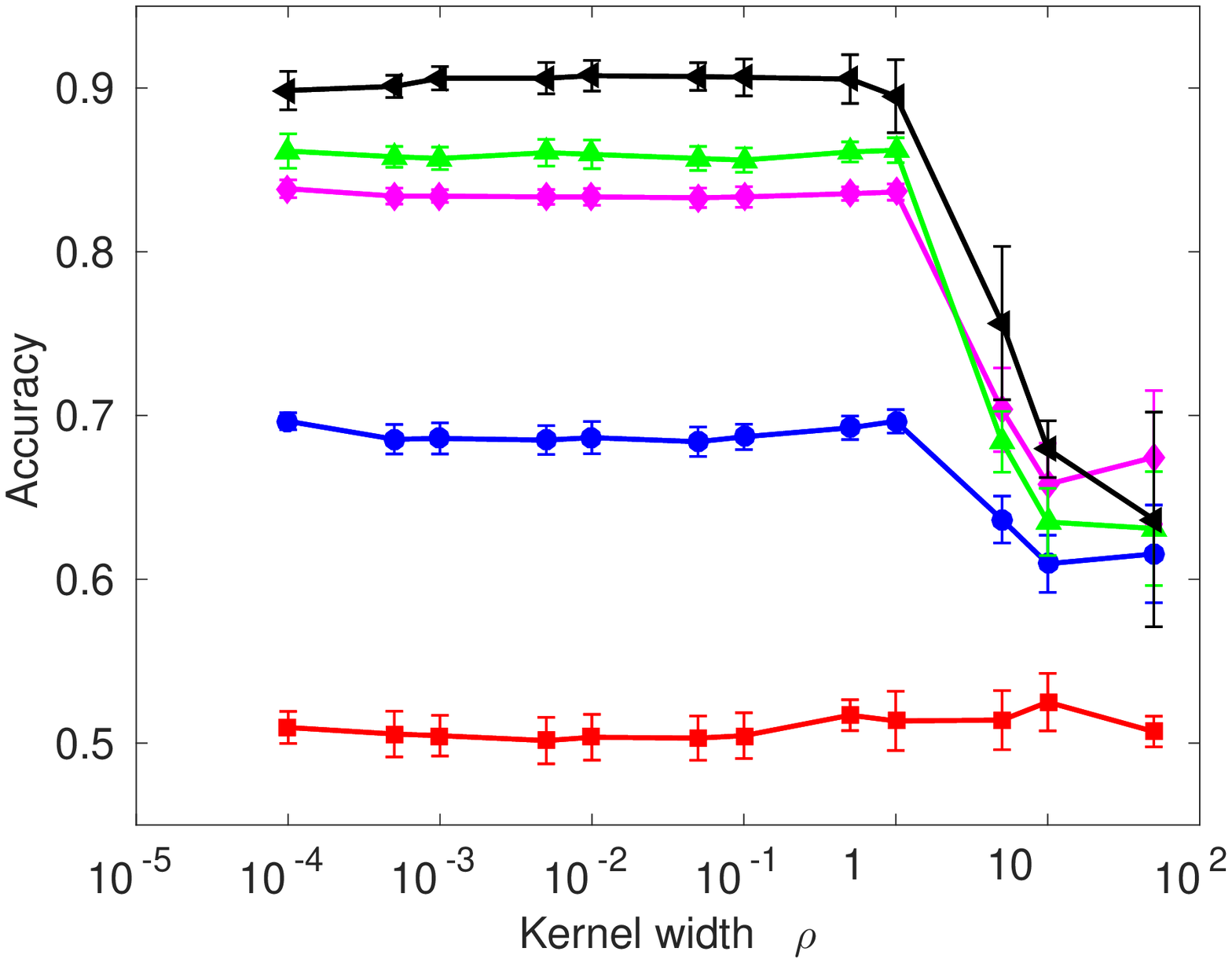}
\end{minipage}
\begin{minipage}[b]{0.45\linewidth}
\includegraphics[width=5.5cm,height=6.25cm]{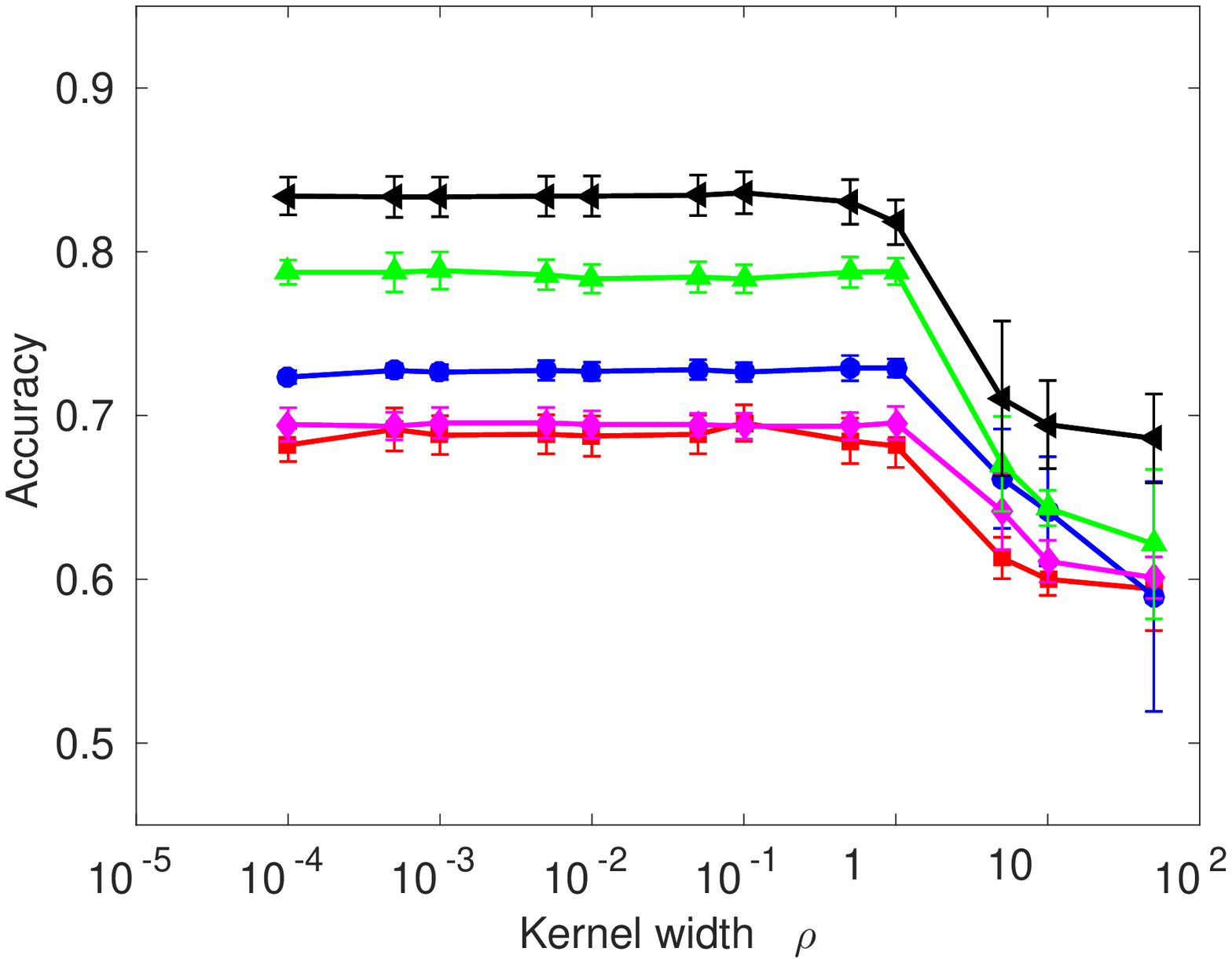}
\end{minipage}
\caption{Classification performance as function of  log kernel width (i.e. $\log_{10} \rho$) using Learning in Model Space (LiMS) method to classify Partially Observed GnRH Models.
Left panel:  The inter-sample intervals ($ISI$) is fixed 75 while the noise standard deviation varies across the set 
$\{0.1, 0.03, 0.01, 0.005, 0.001\}$ (red, blue, magenta, green, and black, respectively).
All observations were sampled on a fixed regular grid over a 7.5h time window.
Right panel: The standard deviation of observation noises is fixed to 0.3 while the $ISI$ value varies across the set 
$\{15, 30, 45, 75, 90 \}$ (black, green, magenta, blue, and red, respectively).
The observation times are random and the $ISI$-values given are the expected value.
Values of the kernel width hyper-parameter for the LiMS classifier were taken from
$\{0.0001, 0.0005, 0.001, 0.005, 0.01, 0.05, 0.1, 0.5, 1, 5, 10, 50\}.$
\label{LiMS-GnRH}
}
\end{figure}

\begin{figure}[!ht]
\centering
\begin{minipage}[b]{0.45\linewidth}
\includegraphics[width=5.5cm,height=6.25cm]{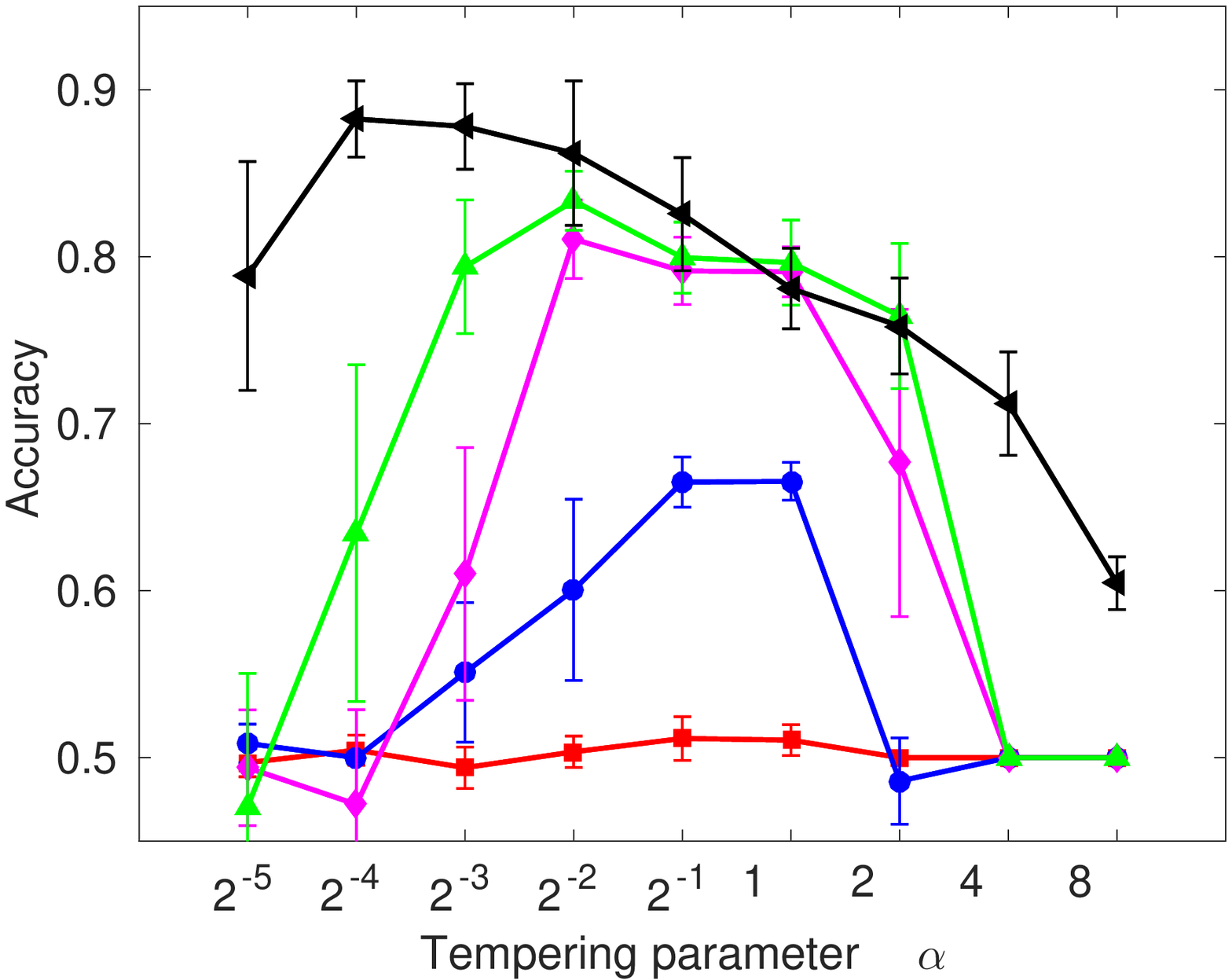}
\end{minipage}
\begin{minipage}[b]{0.45\linewidth}
\includegraphics[width=5.5cm,height=6.25cm]{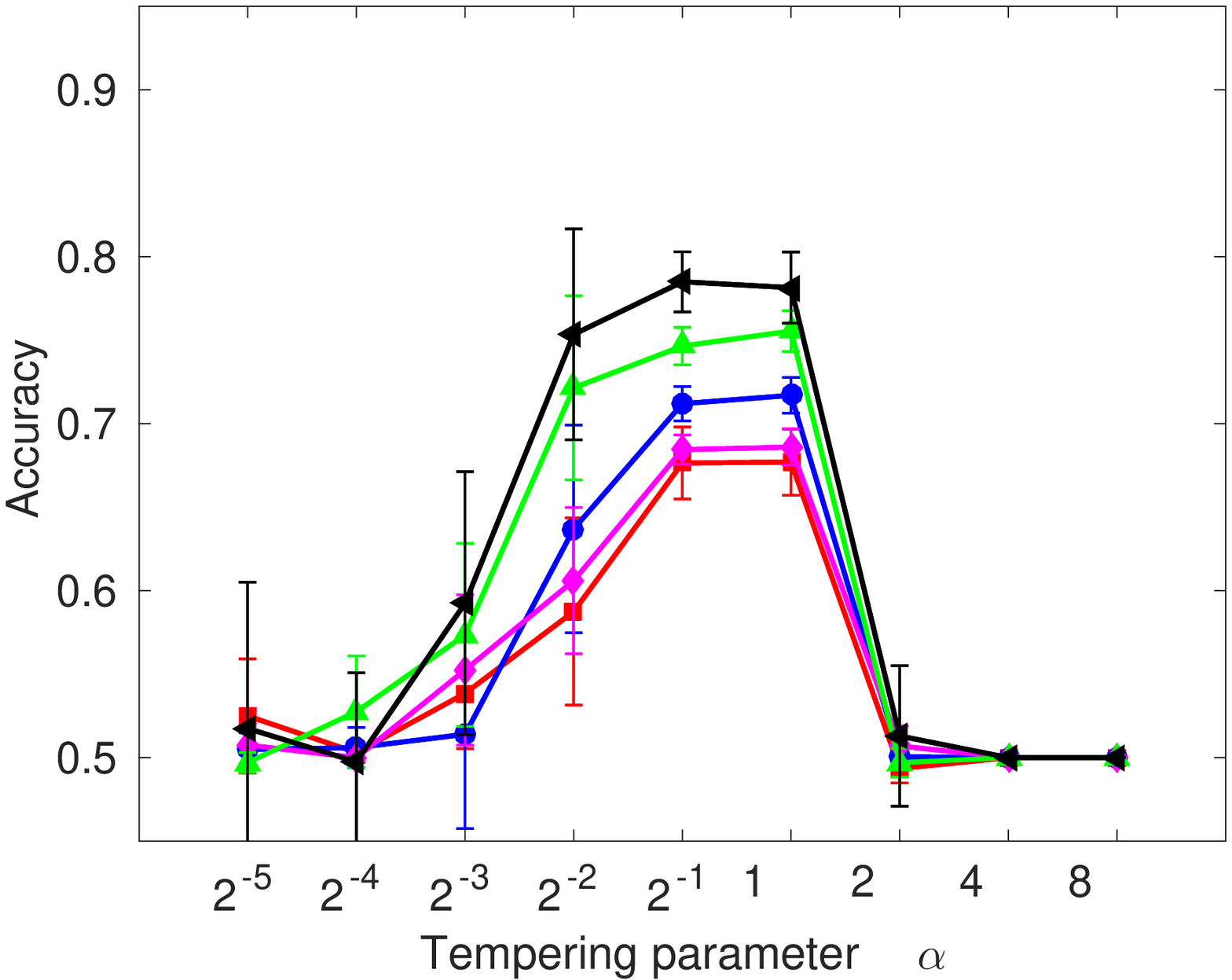}
\end{minipage}
\caption{As same as in Figure~\ref{LiMS-GnRH} but for classification performance as function of  log tempering parameter (i.e. $\log_{2} \alpha$) using Probability Product Kernel (PPK) method. The tempering PPK hyperparameter took values from $\{1/32, 1/16, 1/8, 1/4, 1/2, 1, 2, 4, 8\}$.
}
\label{PPK-GnRH}
\end{figure}

\begin{figure}[!ht]
\centering
\begin{minipage}[b]{0.45\linewidth}
\includegraphics[width=5.5cm,height=6.25cm]{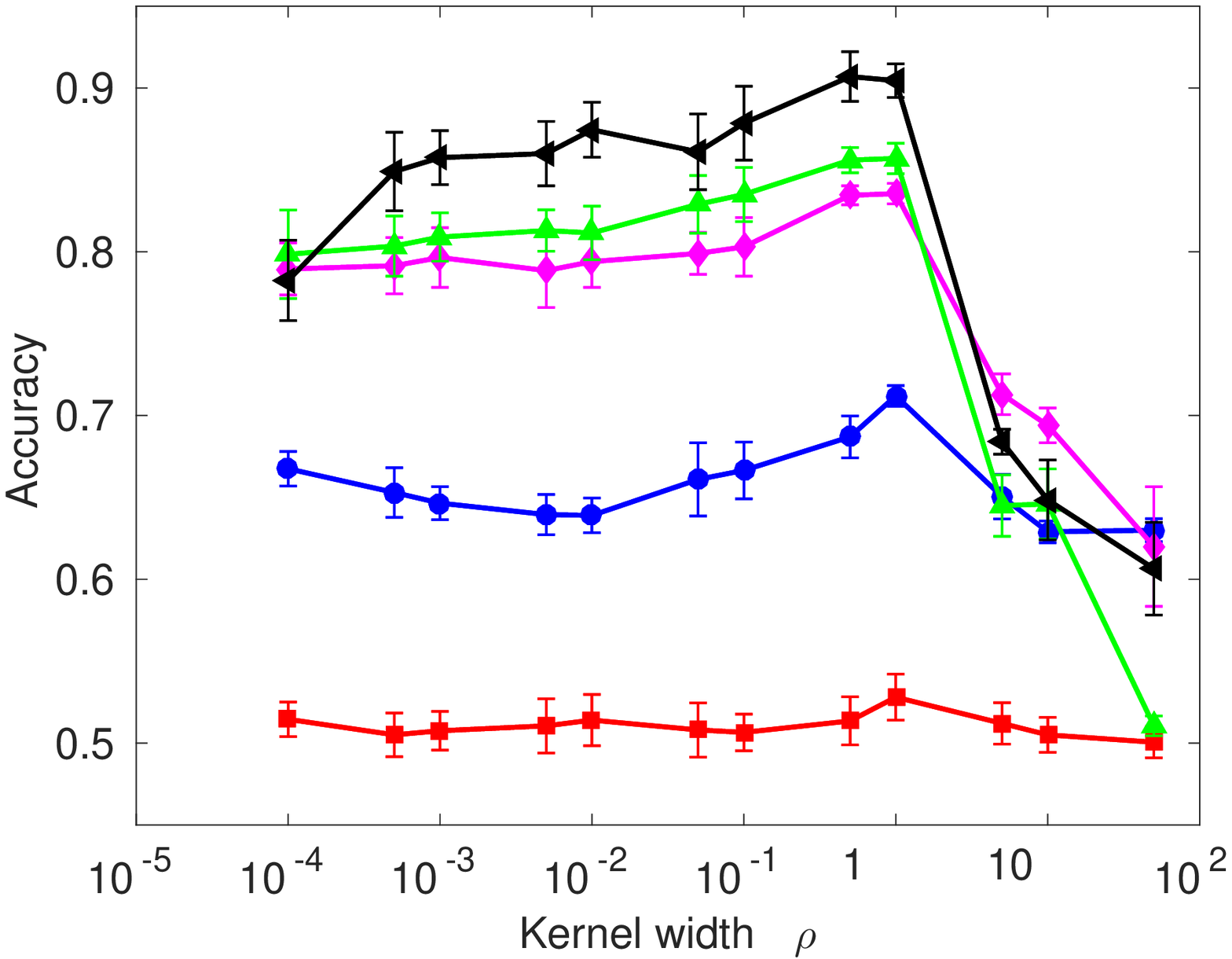}
\end{minipage}
\begin{minipage}[b]{0.45\linewidth}
\includegraphics[width=5.5cm,height=6.25cm]{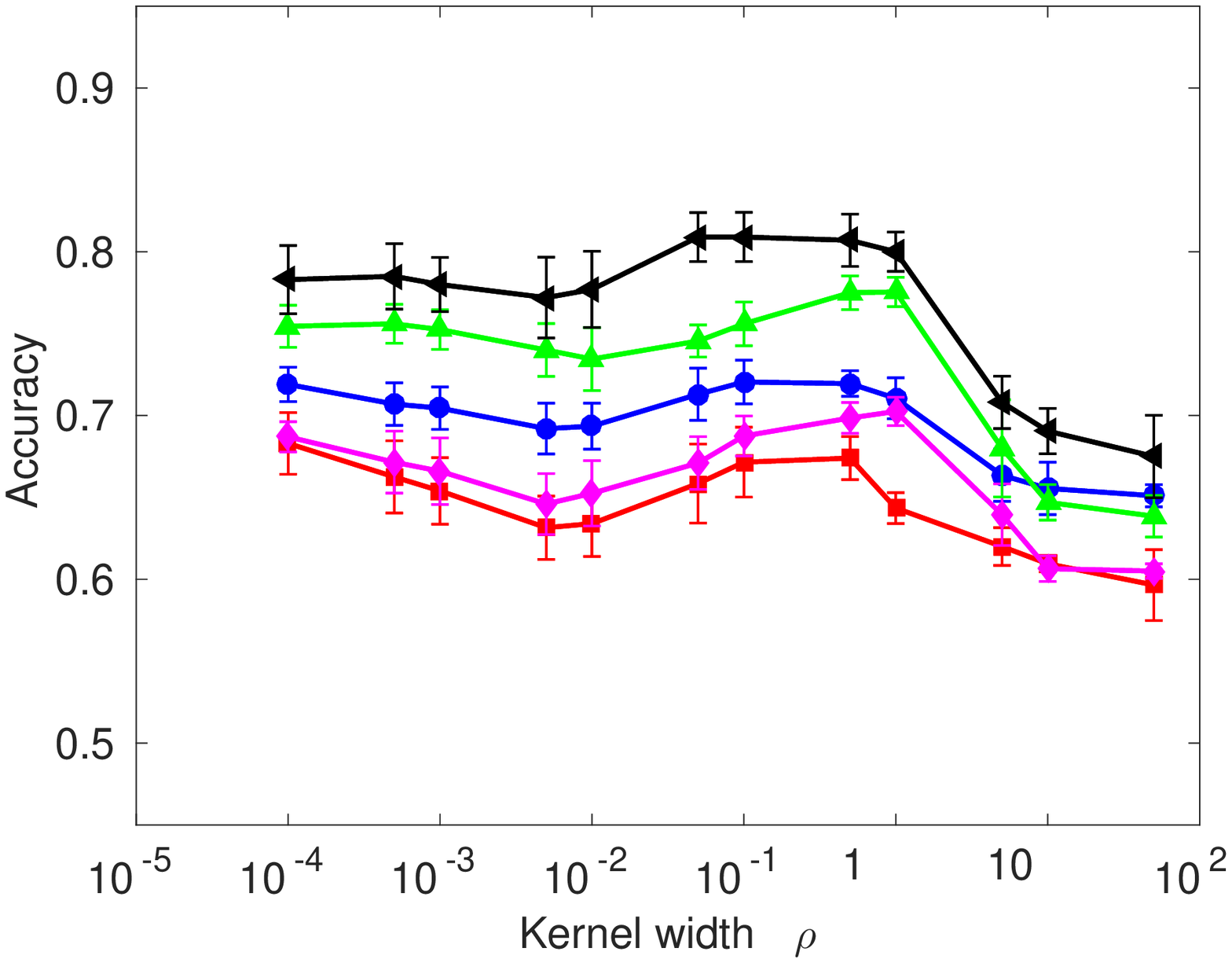}
\end{minipage}
\caption{As same as in Figure~\ref{LiMS-GnRH} but for  Kernel Mean Embedding (KME) Classifier}
\label{KME-GnRH}
\end{figure}

\begin{figure}[t]
\centering
\includegraphics[width=10cm]{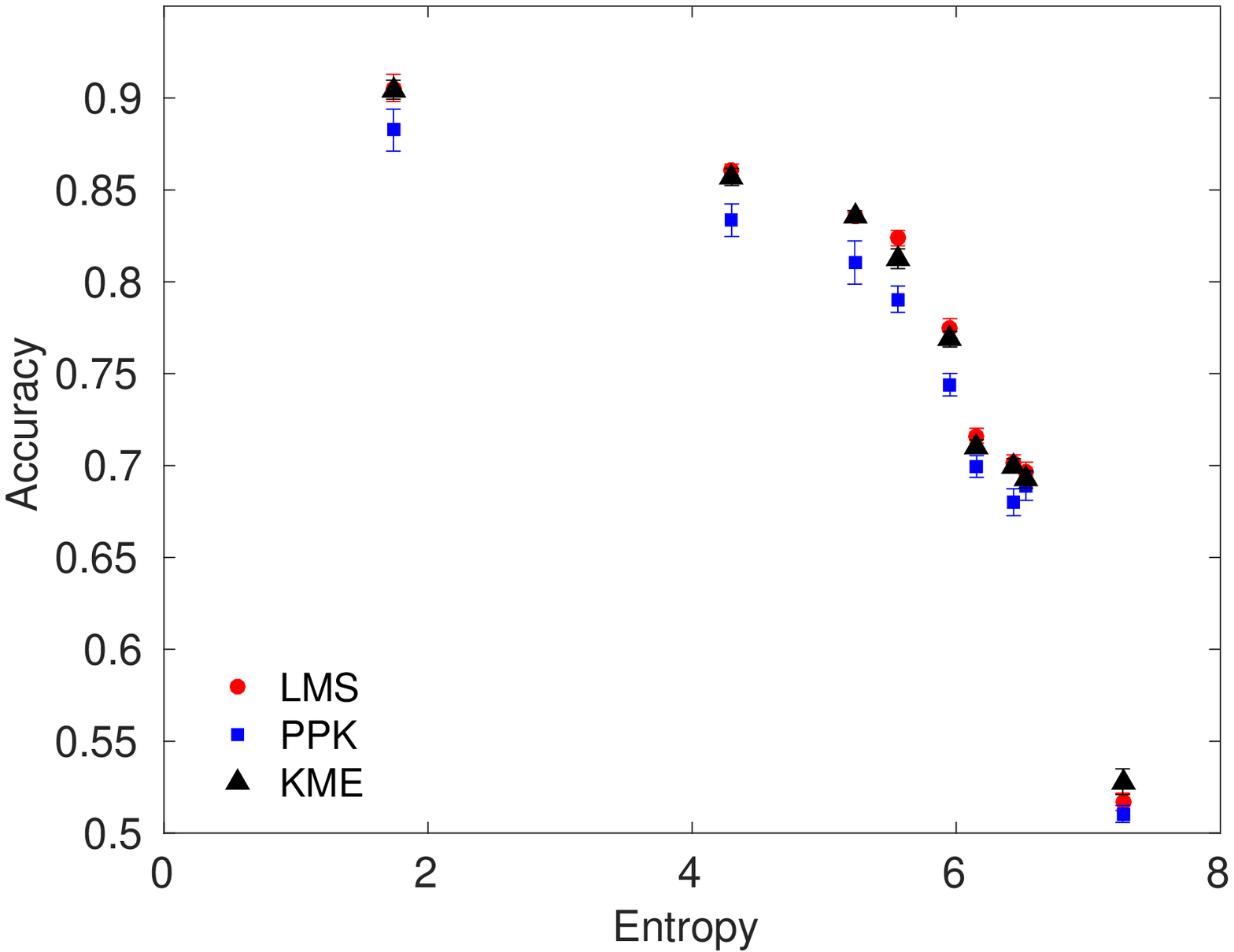}
\caption{Relationship between Classification Performance and Model Uncertainty measured by average Posterior Entropy. Each data point corresponds to one of 15 data sets in {\it Group 1}, {\it Group 2}, and {\it Group 3}. The $x$- and $y$ coordinate of each $\diamond$, $\square$, and $\circ$ point display the average posterior entropy and the accuracy for LiMS, PPK, KME classifiers, respectively.
\label{fig_GnRH_Results_Uncertainty}}
\end{figure}

\begin{flushleft}
{\bf Experiment 1}
\end{flushleft} 
\vspace{0.1cm}
\noindent
In this experiment, we investigated the interplay between the classification performance of the three classifiers (LiMS, PPK, KME) and the level of model uncertainty. In particular, we first used {\it Group 1} data  to study the relation between the accuracy and the level of observation noise (left panel in Figures~\ref{LiMS-GnRH}--\ref{KME-GnRH}). We then used {\it Group 2} and  {\it Group 3} data to evaluate the relation between the accuracy and frequency of observations. The results are presented in the right panel of Figures~\ref{LiMS-GnRH}--\ref{KME-GnRH}.  Only performance curves for {\it Group 3} are shown as the results for {\it Group 2} are very similar to those for {\it Group 3}. Finally, we used all groups to assess the interplay between the accuracy and model uncertainty quantified by the average posterior entropy (Figure~\ref{fig_GnRH_Results_Uncertainty}). 

Figure~\ref{LiMS-GnRH} shows results for the LiMS classifier. In each panel, we plot the testing accuracy against the log kernel width (i.e. $\log_{10} \rho$) for different $\sigma$ values (left panel), or different  $ISI$ values (right panel). Figure~\ref{LiMS-GnRH}  shows in general that the performance increases with decreasing kernel width until a saturation level is reached (approximately at $\rho$ = 1\footnote{Recall that parameter vectors were normalised to lie within the unit cube. Kernel widths substantially larger than 1 introduce a strong model bias that leads to performance degradation.}). The only exception is the case where the model uncertainty is so large that the classifier performs as bad as random guess (see the curve corresponding to $ISI$ = 75 and $\sigma$ = 0.1 in the left panel). On the other hand, it is interesting to observe that the performance is quite robust with respect to kernel width variations below the critical scale of 1. Figure~\ref{LiMS-GnRH} also shows that the performance increases (almost) monotonically with decreasing $ISI$ or $\sigma$, which confirms the hypothesised relationship between classification performance and model uncertainty. 

The results for PPK and KME classifier are displayed in Figure~\ref{PPK-GnRH} and Fig~\ref{KME-GnRH}, respectively. Together with Figure~\ref{LiMS-GnRH}, they show that the character of the interplay between classification performance and model uncertainty is very similar for all three classifiers. Kernel parameters of LiMS and KME classifiers can be related to each other and from this point of view, the LiMS classifier appears to be more robust to variations in the kernel parameter, which is a desirable property. However, the role of tempering kernel parameter in PPK classfier is very different and hence no direct comparison of performance stability with varying kernel parameter can be made with LiMS and KME classifiers.   

Kernels in KME and PPK classifiers effectively smooth and temper, respectively,  the input posterior distributions. As in the case of LiMS classifier, for KME the optimal kernel width is around 1 (parameter vectors are normalised to lie within a unit cube). For PPK, it seems that in most cases, high classification performance is obtained when the posterior distributions are not (or just slightly) tempered. The only exception is \textit{Group-1} PPK curve corresponding to $\sigma$ = 0.0001 and $ISI$=150, where the tempering flattens the model posteriors. 

Figure~\ref{fig_GnRH_Results_Uncertainty} shows the classification performance as a function of model uncertainty for the three PODS classifiers (LiMS, KME, and PPK). For each of 15 data sets, the level of model uncertainty is computed by averaging entropies of   model  posterior distributions inferred from the individual observed time series. The performance is quantified through the accuracy at the kernel parameter determined individually for each classifier and each data set on the validation data\footnote{For LiMS classifier, we chose $\rho$ = 0.5 as the overall ``optimal" kernel width. For KME classifier, it's ``optimal" kernel width is chosen as  $\rho$ = 1.0 for {\it Group 1 \& 2} data and as $\rho$ = 0.5 for {\it Group 3} data. In the case of PPK classifier, we chose $\alpha$ = 0.5 for {\it Group 2} data and $\alpha$ = 1.0 for {\it Group 3} data. For {\it Group 1} data, however, the PPK's ``optimal" tempering parameter decreases with $\sigma$, that is,  $\alpha$ = 0.5 for $\sigma$ = 0.1 \&  $\sigma$ =  0.03, $\alpha$ = 0.25 for $\sigma$ = 0.01 \&  $\sigma$ =  0.005, and $\alpha$ = 0.0625 for $\sigma$ = 0.001.}. Recall that for each of the 15 observation sets, we have 10 performance measures obtained on 10 resampled training/hold-out sets. We combine all performance and uncertainty measures corresponding to the same $ISI$ and $\sigma$ (regardless of whether the observation times are random or not) into a single set. This results in 9 sets of (uncertainty, performance) values. For each set, the corresponding average posterior entropy, observation noise level $\sigma$, and inter-sample interval $ISI$ are given in the first two columns of Table~\ref{signrank}. In Figure~\ref{fig_GnRH_Results_Uncertainty} the means and standard deviations of the performance measures are plotted against the corresponding average posterior entropy. The plot shows a clear drop-off in classification performance at high model uncertainty (of about 5 nats). In this respect, there is no significant difference among the three classifiers.  However, for low and moderate model uncertainty levels both LiMS and KME outperform PPK. LiMS and KME also have comparable classification performance. In Table~\ref{signrank}, the $p$-values from sign-rank statistical tests are given for the following one-sided hypothesis: ({\bf H1}) LiMS outperforms KME; ({\bf H2}) LiMS outperforms PPK; and ({\bf H3}) KME outperforms PPK. The $p$-values here mean the probability for the corresponding (one-sided) hypothesis being true just by chance. 

\begin{table}[!th]
\centering
\begin{tabular}{| l |  l | l | l | l |}
\hline
Entropy & ($\sigma$, $ISI$) & {\bf H1}   & {\bf H2}  & {\bf H3}  \\ \hline
1.7 & (0.001, 75) & 0.39 & {\bf 0.01} & {\bf 0.01} \\ \hline
4.3 & (0.005, 75) & 0.22 & {\bf 0.00} & {\bf 0.02} \\ \hline
5.2 & (0.01, 75)  &  0.49 & {\bf 0.02} & {\bf 0.01} \\ \hline
5.6 & (0.03, 15)  &  {\bf 0.01} & {\bf 0.00} & {\bf 0.00}\\ \hline
6.0 & (0.03, 30)  &  0.17 & {\bf 0.00} & {\bf 0.00} \\ \hline
6.2 & (0.03, 45)  &  {0.09} & {\bf 0.01} & {\bf 0.02}\\ \hline
6.4 & (0.03, 75)   &  0.36 & {\bf 0.00} &{\bf  0.00}\\ \hline
6.5 & (0.03, 90)   &  0.12 & {0.07} & 0.29\\ \hline
7.3 & (0.1, 75)   &  0.95 & 0.15 & {\bf 0.01}\\ \hline
\end{tabular} 
\caption{
Sign-rank tests for comparing the performance of LiMS, KME, and PPK classifiers at different levels of model uncertainty with the following one-sided hypothesis:
({\bf H1}) LiMS outperforms KME;
({\bf H2}) LiMS outperforms PPK; and
({\bf H3}) KME outperforms PPK.
The $p$-values from these tests are given in Column 4--6 and
all $p$-values smaller than 0.05 are highlighted in bold font. 
The level of model uncertainty is measured by (average) posterior entropy (Column 1). 
The corresponding observation noise level $\sigma$ and the inter-sample interval $ISI$ values are given in Column 2 and 3, respectively.
\label{signrank}}
\end{table}

\begin{flushleft}
{\bf Experiment 2}
\end{flushleft} 
\vspace{0.1cm}
\noindent
In this experiment we investigate whether the performance of classifying partially observed GnRH models would be impaired if simpler reduced complexity GnRH model structures $M_2$ and $M_3$ of section~\ref{sec:GnRH} were used to infer the input posterior distributions representing observation sets generated from the full model $M_1$. 

For each time series data set, we evaluated the LiMS performance when using the $M_1$- (as a reference), $M_2$- and $M_3$-generated  posterior distributions representing the observation sequences. The results, summarised in Table~\ref{GnRH_MR_Results}, show that the performance is closely comparable for all inferential model structures $M_1$--$M_3$,  for all observation sets. Recall that two classification GnRH classes differ only in their frequency-response characteristics that are completely determined by the $K_{d_{\mbox{\tiny TF}_1}}$- and $K_{d_{\mbox{\tiny TF}_2}}$ values. All three model structures $M_1$--$M_3$ include compartment C3 which modulates the observable model output $[GSU]$  via Eq.~\ref{GnRH_output}. Moreover, the dynamics of $[GSU]$ is controlled by $K_{d_{\mbox{\tiny TF}_1}}$ and $K_{d_{\mbox{\tiny TF}_2}}$. Our results confirm one of the key points of this study: {\em For classification of PODS via Learning in the Model Space framework, it is nor necessary for the inferential model structure to be a perfect model of the underlying dynamical system generating the data, as long as the reduced complexity inferential model structure captures the essential characteristics needed for the given classification task.}

\begin{table}[!th]
\centering
\begin{tabular}{|c | c |  c | c | c |}
\hline
Data Sets                & ($\sigma$, $ISI$) & {\it M1}   &  {\it M2}     & {\it M3} \\ \hline
\multirow{3}{*}{Group 1} & (0.001, 75)        & 0.91 $\pm$ 0.02 &  0.88 $\pm$ 0.03   & 0.90 $\pm$ 0.01  \\ 
                         & (0.01,  75)        & 0.84 $\pm$ 0.01 &  0.84 $\pm$ 0.01   & 0.83 $\pm$ 0.01  \\
                         & (0.1,   75)        & 0.52 $\pm$ 0.01 &  0.54 $\pm$ 0.01   & 0.54 $\pm$ 0.01  \\ \hline
\multirow{3}{*}{Group 2} & (0.03,  90)        & 0.82 $\pm$ 0.00 &  0.81 $\pm$ 0.01   & 0.81 $\pm$ 0.01  \\ 
                         & (0.03, 30)        & 0.74 $\pm$ 0.01 &  0.73 $\pm$ 0.01   & 0.72 $\pm$ 0.01  \\
                         & (0.03, 45)        & 0.71 $\pm$ 0.01 &  0.70 $\pm$ 0.02   & 0.69 $\pm$ 0.01  \\ \hline
\multirow{3}{*}{Group 3} & (0.03,  90)        & 0.83 $\pm$ 0.02 &  0.82 $\pm$ 0.02   & 0.82 $\pm$ 0.01  \\ 
                         & (0.03, 30)        & 0.69 $\pm$ 0.01 &  0.71 $\pm$ 0.01   & 0.71 $\pm$ 0.01  \\
                         & (0.03, 30)        & 0.68 $\pm$ 0.02 &  0.69 $\pm$ 0.01   & 0.68 $\pm$ 0.02  \\ \hline                         
\end{tabular} 
\caption{The LiMS's task performance for classifying partially observed GnRH model when using three different inferential GnRH models (i.e. {\it M1},{\it M2}, and {\it M3} described in Section~\ref{sec:GnRH}) to infer the input posterior distributions from the $[GSU]$ time series. For these inferential models, their corresponding mean performance ($\pm$ standard deviation) obtained from nine different time series data sets are summarised in Column 3 -- 5 (respectively). The observation settings of these data sets are given in Column 1 -- 2 where $\sigma$ denotes the observation noise level  and  $ISI$ the sampling frequency.
\label{GnRH_MR_Results}}.
\end{table}

\subsection{Double-well Model}

For partially observed stochastical double-well systems (SDWs), the task is to classify posterior distributions over the models accessible through parameter vectors ($a$, $d$, $\kappa$). Recall that $a$ is the asymmetry parameter, $d$ is the well location parameter, and $\kappa$ represents the dynamical noise level. Also recall that the two classes of SDWs involved in our experiments are defined through two class-conditional Gaussian distributions in the parameter space: ($\bar d_1$ + $\epsilon_d$, $\bar \kappa_1$ + $\epsilon_{\kappa}$, $\bar a_1$) for {\it Class 1} and ($\bar d_0$ + $\epsilon_d$,  $\bar \kappa_0$ + $\epsilon_{\kappa}$, $\bar a_0$) for {\it Class 0}, where  ($\bar d_1$, $\bar \kappa_1$, $\bar a_1$) and ($\bar d_0$,  $\bar \kappa_0$, $\bar a_0$) denote the class-conditional prototypical model parameter; $\epsilon_d$ and $\epsilon_{\kappa}$ are Gaussian-distributed zero-mean random variables with standard deviations 0.1/3 and 0.05/3, respectively.

\begin{table}[!ht]
\begin{tabular}{| c |  c c |}
\hline
 $(\bar d, \bar \kappa, \bar a)$ & {\it Class 1} &  {\it Class 0} \\ \hline
{\it Task 1} & (1.0, 1.0, -0.1) & (1.3, 1.5, 0.1) \\ \hline     
{\it Task 2} & (1.0, 1.5, 0) & (1.3, 1.5, 0) \\ \hline     
{\it Task 3} & (1.0, 1.5, 0) & (1.2, 1.5, 0) \\ \hline              
\end{tabular} 
\begin{tabular}{| c |  c c | }
\hline
($\sigma$, $ISI$)  & {\it Group 1} & {\it Group 2}  \\ \hline        
{\it Set 1}        & (0.3, 0.5)      & (0.3, 0.5) \\ \hline           
{\it Set 2}        & (0.4, 0.5)      & (0.3, 1.0)  \\ \hline           
{\it Set 3}        & (0.6, 0.5)      & (0.3, 1.25)  \\ \hline           
\end{tabular}
\caption{Left: The specification of two classes of partially observed stochastical double-well systems in three classification tasks by their respective prototypical model parameters. Right: The specification of two groups of observation sets generated for each of three tasks by their respective observation noise level $\sigma$ and inter-sample interval $ISI$.
\label{sdw_task_specification}} 
\end{table}

To compare our LiMS  classifier with KME and PPK classifiers, we define a hierarchy of three tasks of increasing complexity, denoted by {\it Task 1}--{\it Task 3} (see Table~\ref{sdw_task_specification}). Furthermore, to investigate the relation between the level of model uncertainty and classifier performance, for each of the three tasks, we generate two groups of observation sets (denoted by {\it Group 1} and {\it Group 2}). Each group consists of three observation sets with varying degrees of model uncertainty. As in the GnRH experiment, the model uncertainty level induced by each observation set is determined by the corresponding observation noise level $\sigma$ and the inter-sample interval $ISI$. The increase of uncertainty level in  {\it Group 1} and {\it Group 2} is modulated by increasing $\sigma$ and $ISI$, respectively (see Table~\ref{sdw_task_specification} ). For both groups, the time series in each observation set were sampled at regularly spaced observation times (with inter-sample interval $ISI$)  within the time interval [0, 50].

As we adopt a finite-grid approximation approach to compute the model posteriors, the parameter space $\varTheta$ is discretised as follows: $d \in \{ 0.1, 0.2, ..., 1.9, 2.0\}$,  $\kappa \in \{ 0.1, 0.2, ..., 1.9, 2.0\}$, and $a \in \{ -0.2, -0.1, 0, 0.1, 0.2\}$.

\begin{figure}[!t]
\centering
\includegraphics[width=14cm,height=11cm]{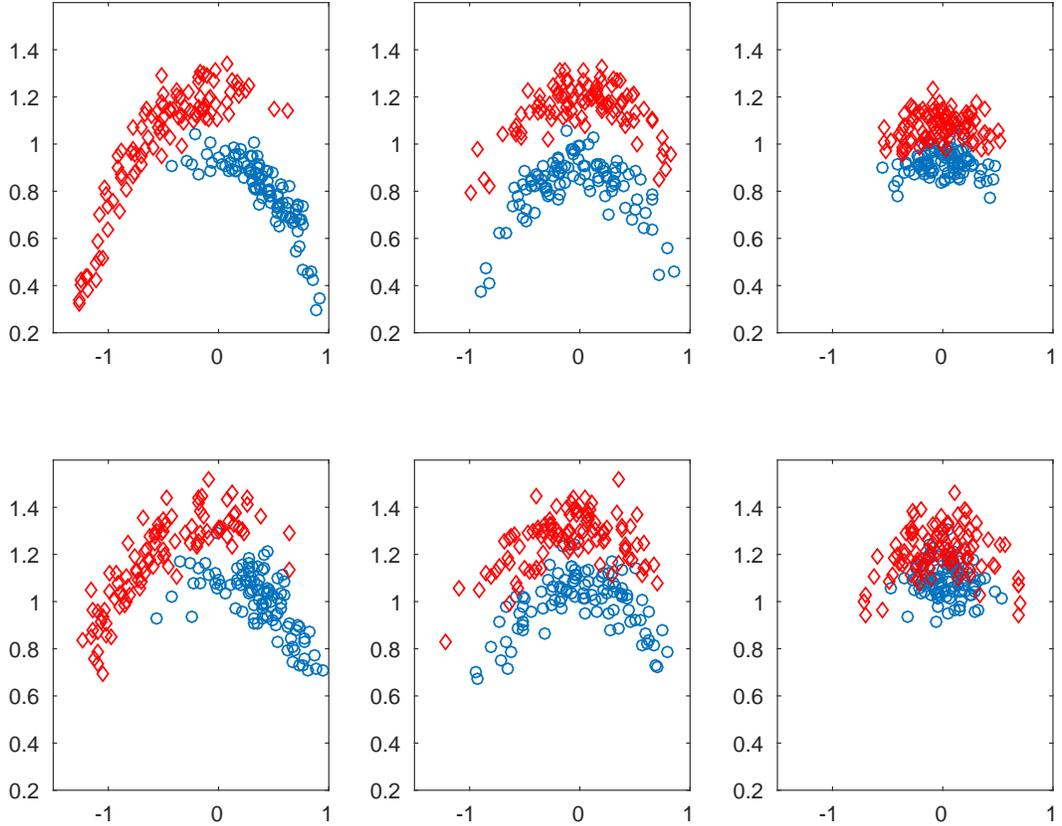}
\caption{Scatter plot of the mean and standard deviation pairs  $(\mu_y, \gamma_y)$ computed for the time series  $\{ y_t \}$ observed (1) from different tasks (From left to right: {\it Task 1}, {\it Task 2}, and {\it Task 3}) and (2) with different ($\sigma$, $ISI$) settings (From top to bottom: (0.2, 0.5), and (0.6, 0.5)). The data points in the scatter plots from {\it Class 1} and {\it Class 0} are displayed in red and blue, respectively.
\label{sdw_task_comparison}}
\end{figure}

One may argue that, given the nature of the classification tasks outlined above, the mean $\mu_y$ and standard deviation $\gamma_y$ of the observed time series $\{ y_t \}$  can provide useful features for building a  classifier solely operating in the signal space. Such feature vectors $(\mu_y, \gamma_y)$ can also provide an insight regarding the task complexity. Figure~\ref{sdw_task_comparison} shows six scatter plots  of $(\mu_y, \gamma_y)$ for {\it Task 1, 2} and {3} (left, middle and right column, respectively) and for ($\sigma$, $ISI$) = (0.3, 0.5) and  ($\sigma$, $ISI$) = (0.6, 0.5) (upper and lower row, respectively). The class labels are indicated by colours (red for {\it Class 1} and blue for {\it Class 0}). For {\it Task 1}, the asymmetry parameter $a$ is class-dependent and Figure~\ref{sdw_task_comparison} shows that in this case, the two classes can be separated simply by using the time series' means $\mu_y$. For example, a positive value of $a$ would cause the means $\mu_y$ of time series from the corresponding class to be biased towards a positive value and vice versa. In {\it Task 2}, $a=0$ for both classes and the means $\mu_y$ can no longer separate the two classes. However, classification is still possible in the joint space $(\mu_y, \gamma_y)$. By gradually reducing the difference between the two classes in terms of the dynamical noise level $\kappa$, the classes can  be brought closer together in the $(\mu_y, \gamma_y)$ space in a controlled manner.  To tease out possible advantages of the learning in the model space framework, in all SDW experiments we also employ a signal-space baseline KLR classifier (bKLR) solely operating on $(\mu_y, \gamma_y)$ .

\begin{figure}[!t]
\centering
\begin{minipage}[b]{0.45\linewidth}
\includegraphics[width=5cm,height=4.5cm]{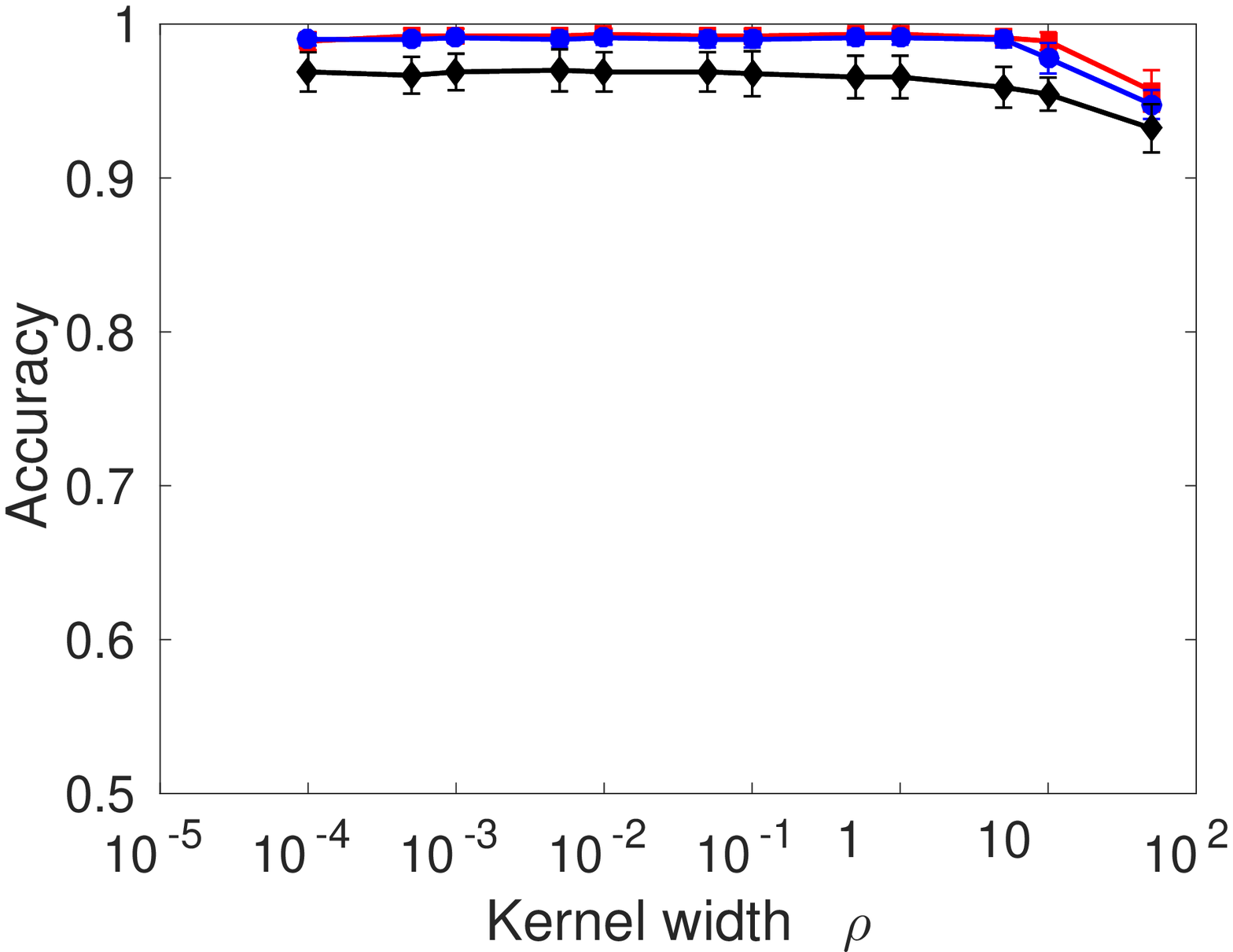}
\end{minipage}
\vspace{0.25cm}
\begin{minipage}[b]{0.45\linewidth}
\includegraphics[width=5cm,height=4.5cm]{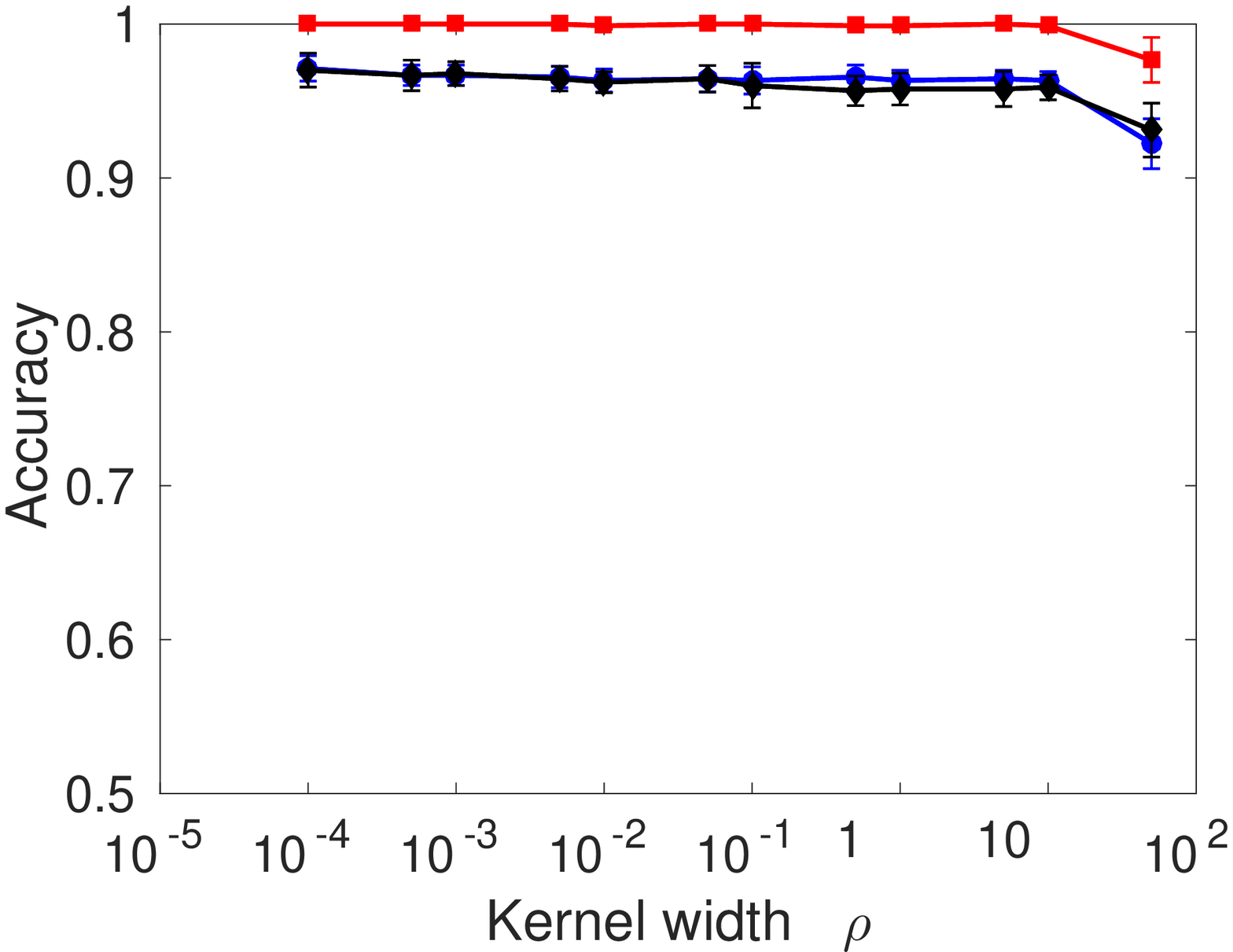}
\end{minipage}
\begin{minipage}[b]{0.45\linewidth}
\includegraphics[width=5cm,height=4.5cm]{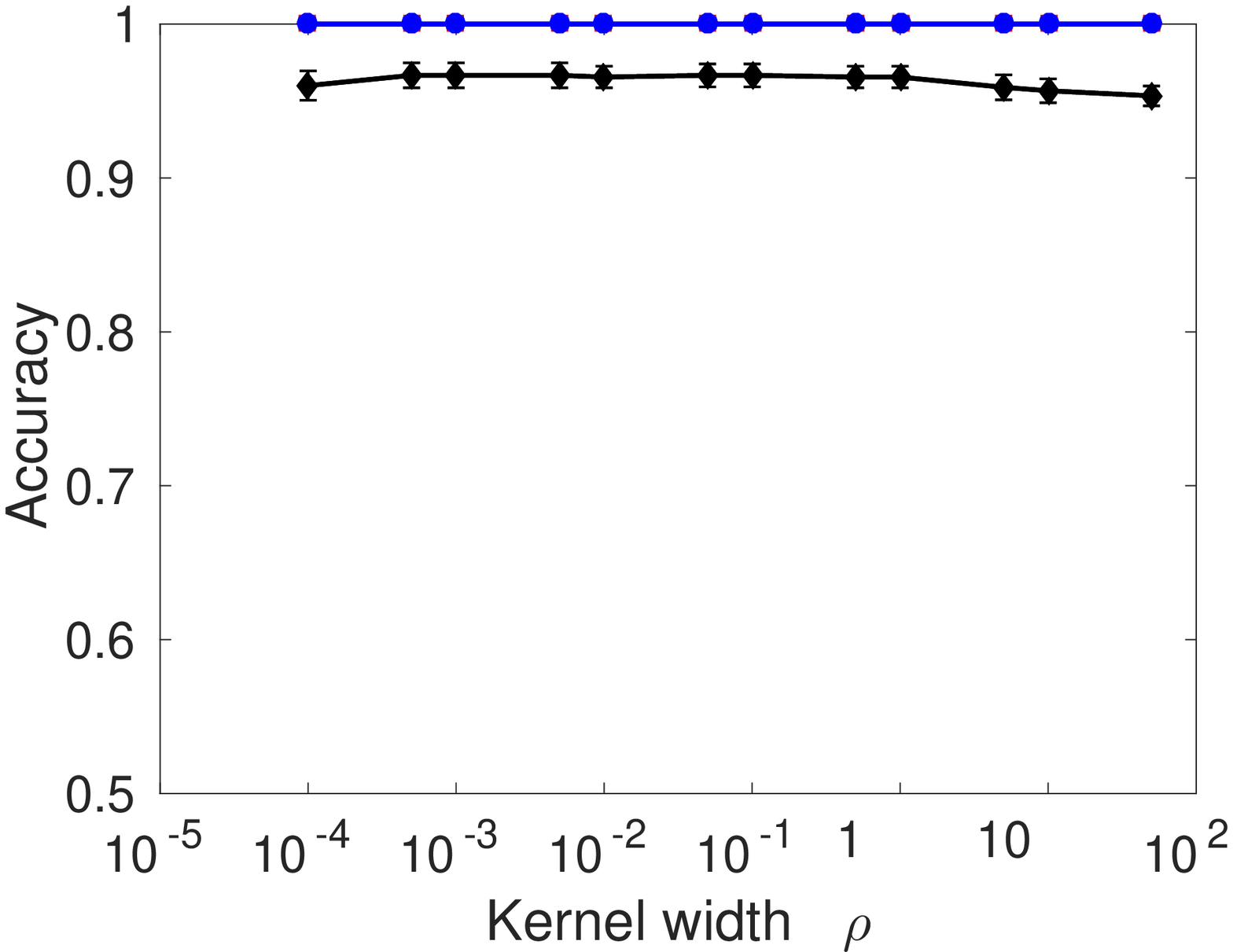}
\end{minipage}
\vspace{0.25cm}
\begin{minipage}[b]{0.45\linewidth}
\includegraphics[width=5cm,height=4.5cm]{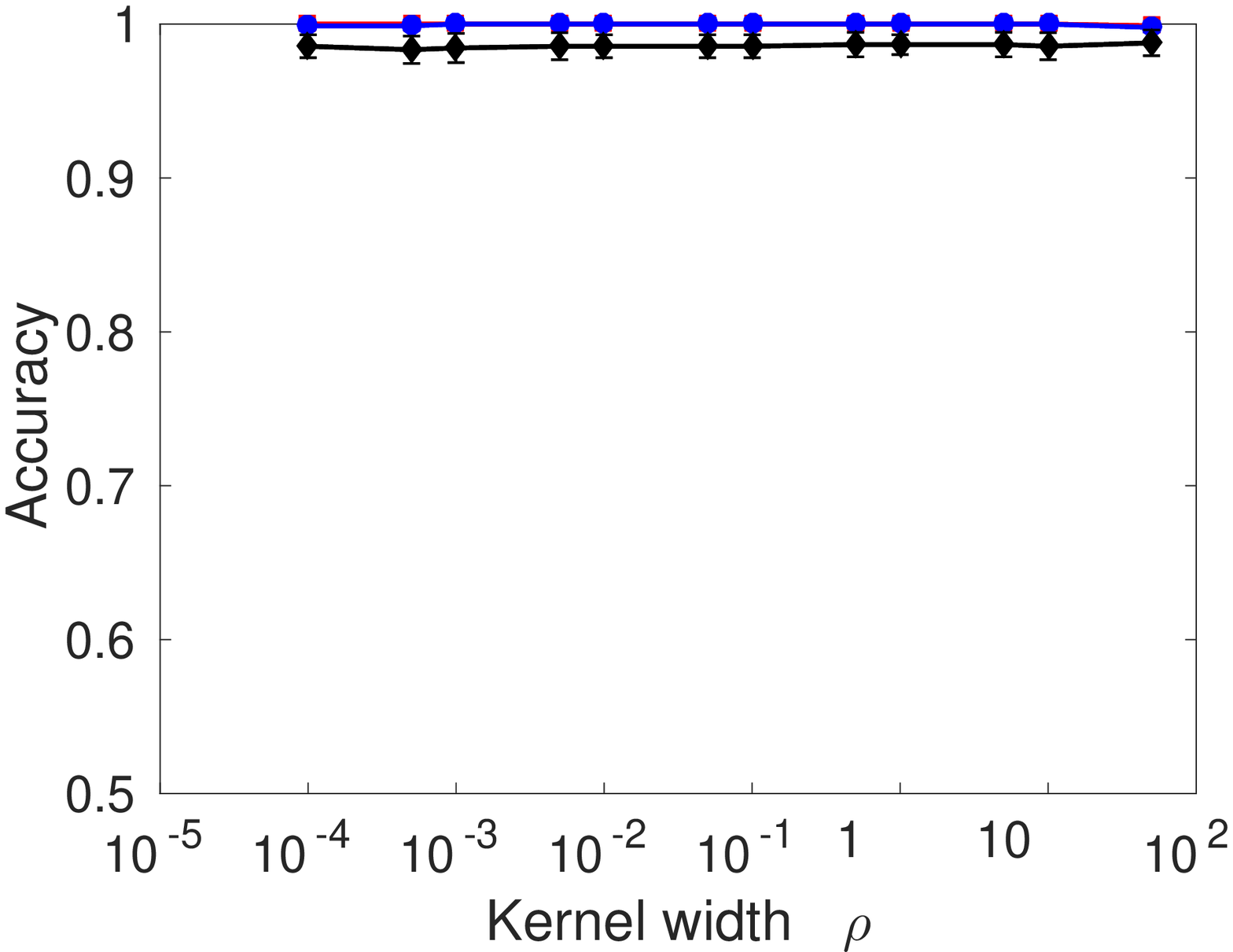}
\end{minipage}
\begin{minipage}[b]{0.45\linewidth}
\includegraphics[width=5cm,height=4.5cm]{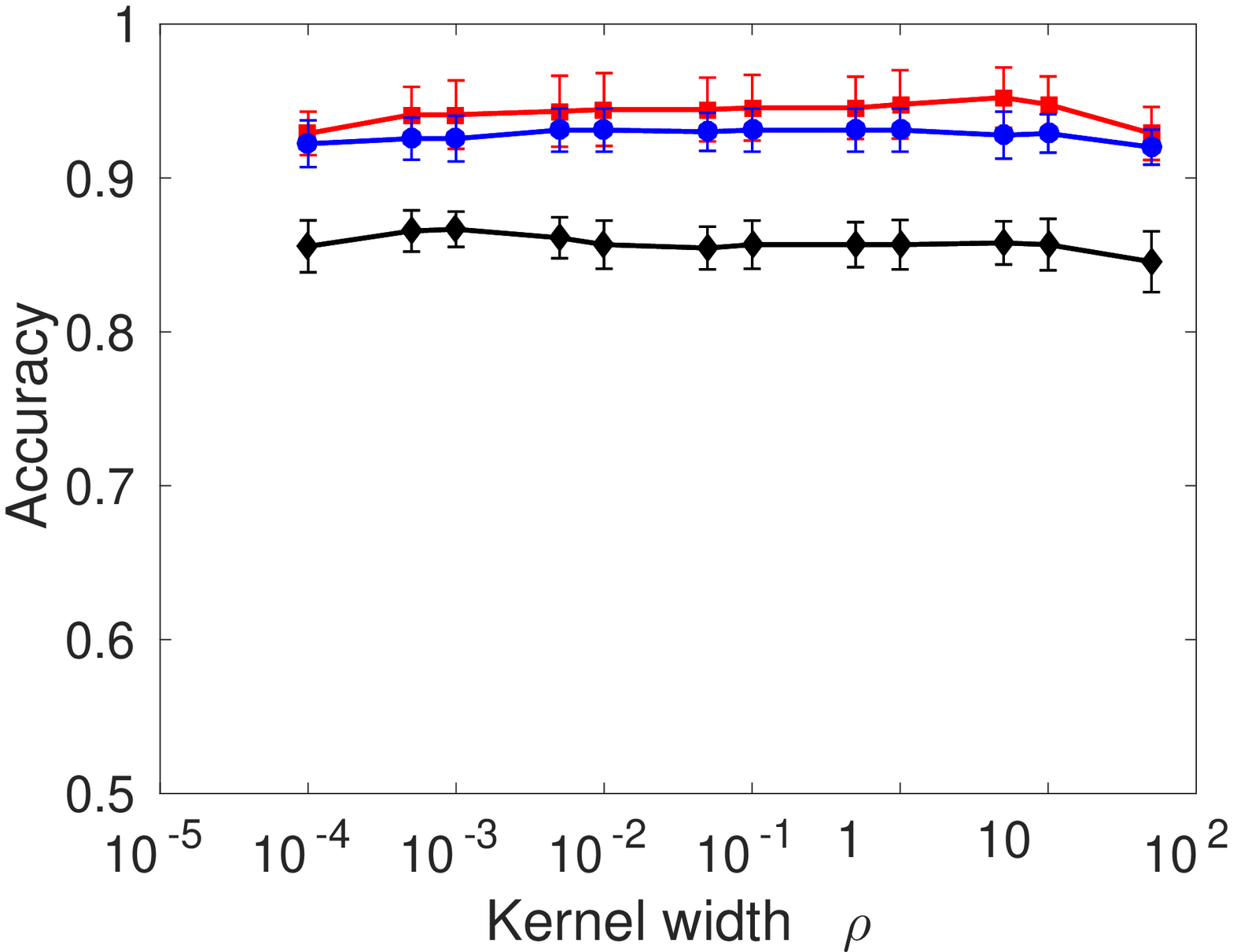}
\end{minipage}
\begin{minipage}[b]{0.45\linewidth}
\includegraphics[width=5cm,height=4.5cm]{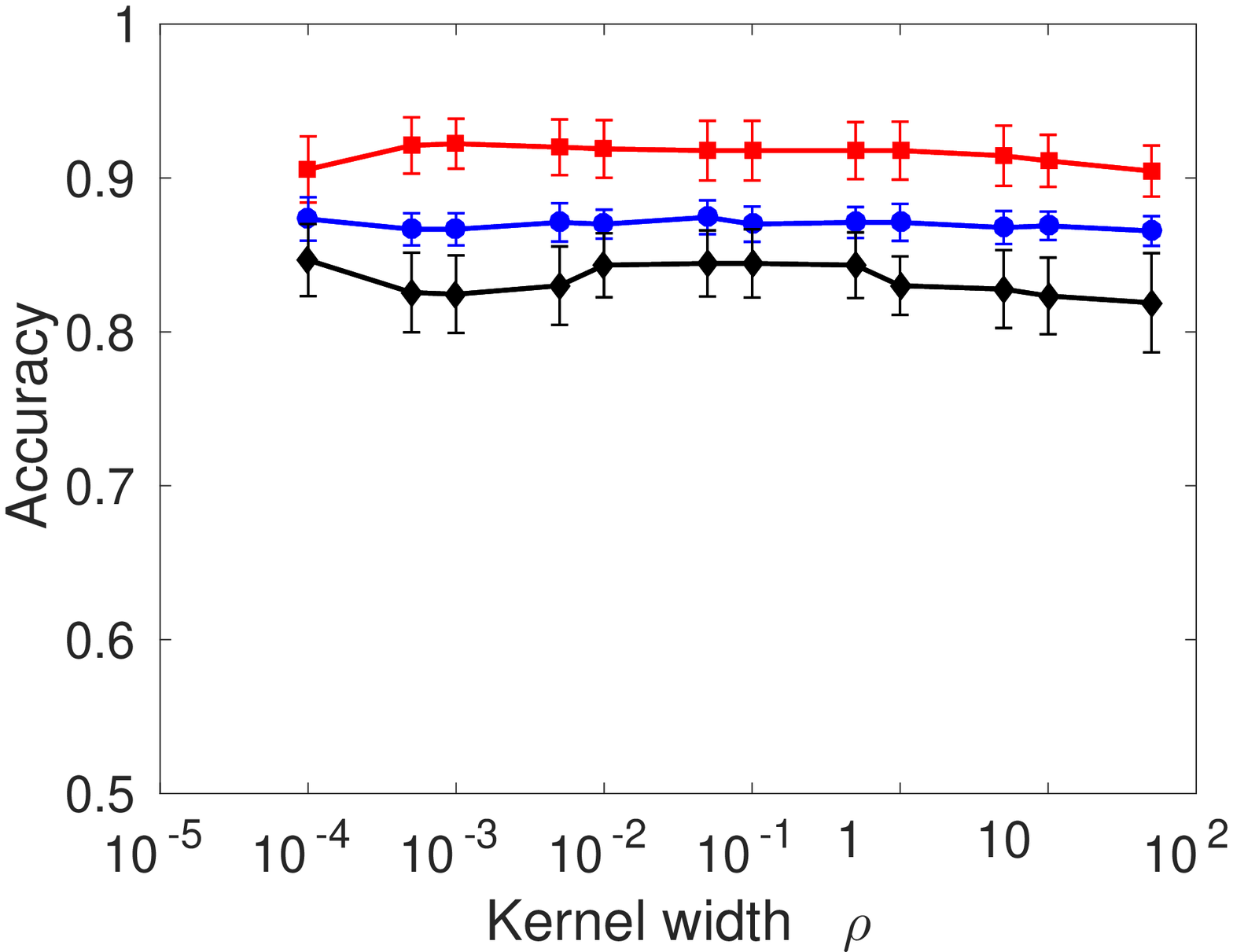}
\end{minipage}
\caption{Classification performance as function of log kernel width (i.e. $\log_{10} \rho$) using LiMS classifier to classify partially stochastic double-well systems
for different tasks (From top to bottom: {\it Task 1} -- {\it Task 3}) and for different observation settings (
Left: ($\sigma$, $ISI$) = (0.3, 0.5), (0.4, 0.5), and (0.6, 0.5) with red, blue, and black (respectively)
and 
Right: ($\sigma$, $ISI$) = (0.3, 0.5), (0.3, 1.0), (0.3, 1.25)  with red, blue, and black (respectively)
).
\label{PGDW-LiMS}}
\end{figure}

\begin{flushleft}
%\subsubsection
{\bf Experiment 1}
\end{flushleft} 
\vspace{0.1cm}
\noindent

\begin{figure}[!t]
\centering
\begin{minipage}[b]{0.45\linewidth}
\includegraphics[width=5cm,height=4.5cm]{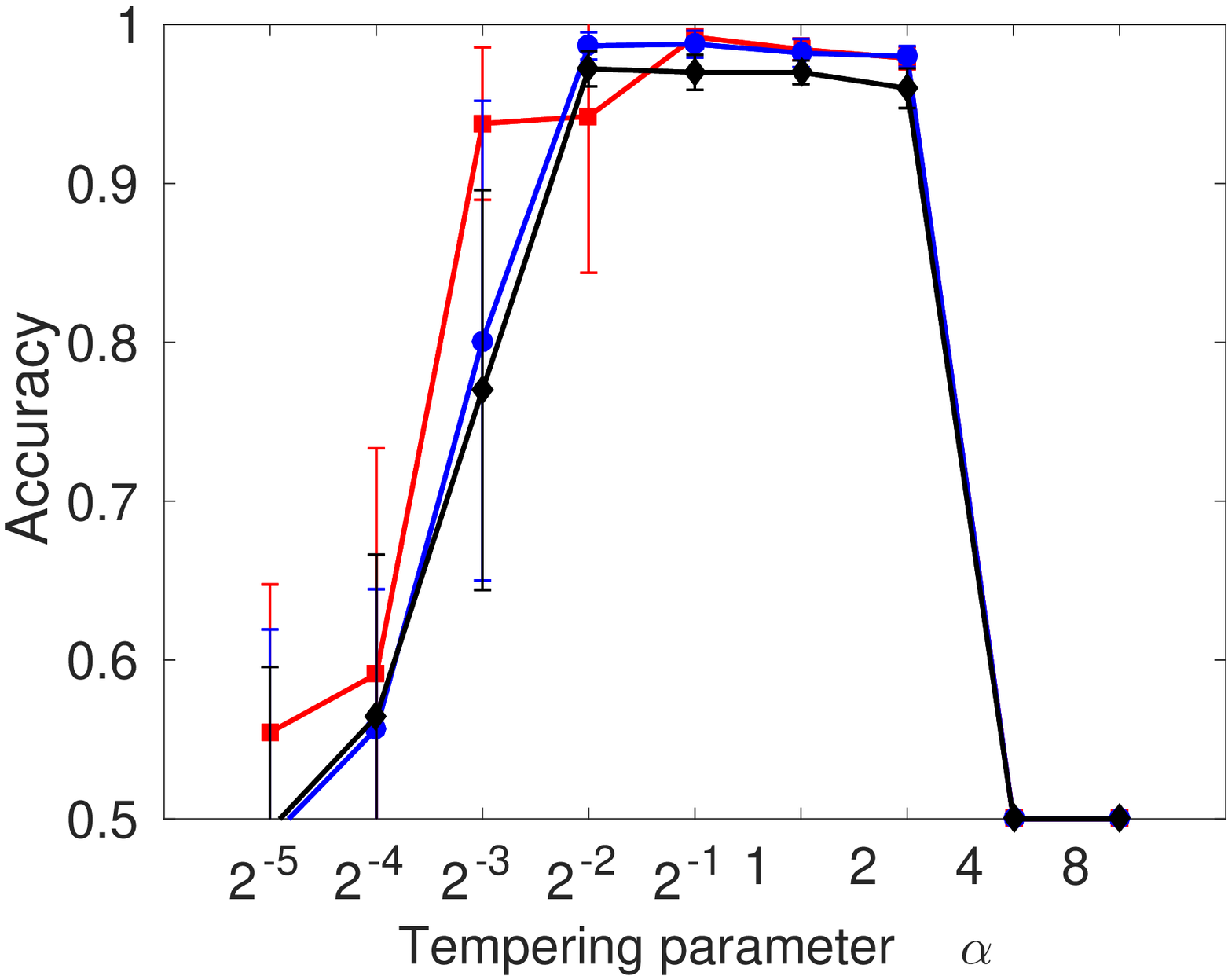}
\end{minipage}
\vspace{0.25cm}
\begin{minipage}[b]{0.45\linewidth}
\includegraphics[width=5cm,height=4.5cm]{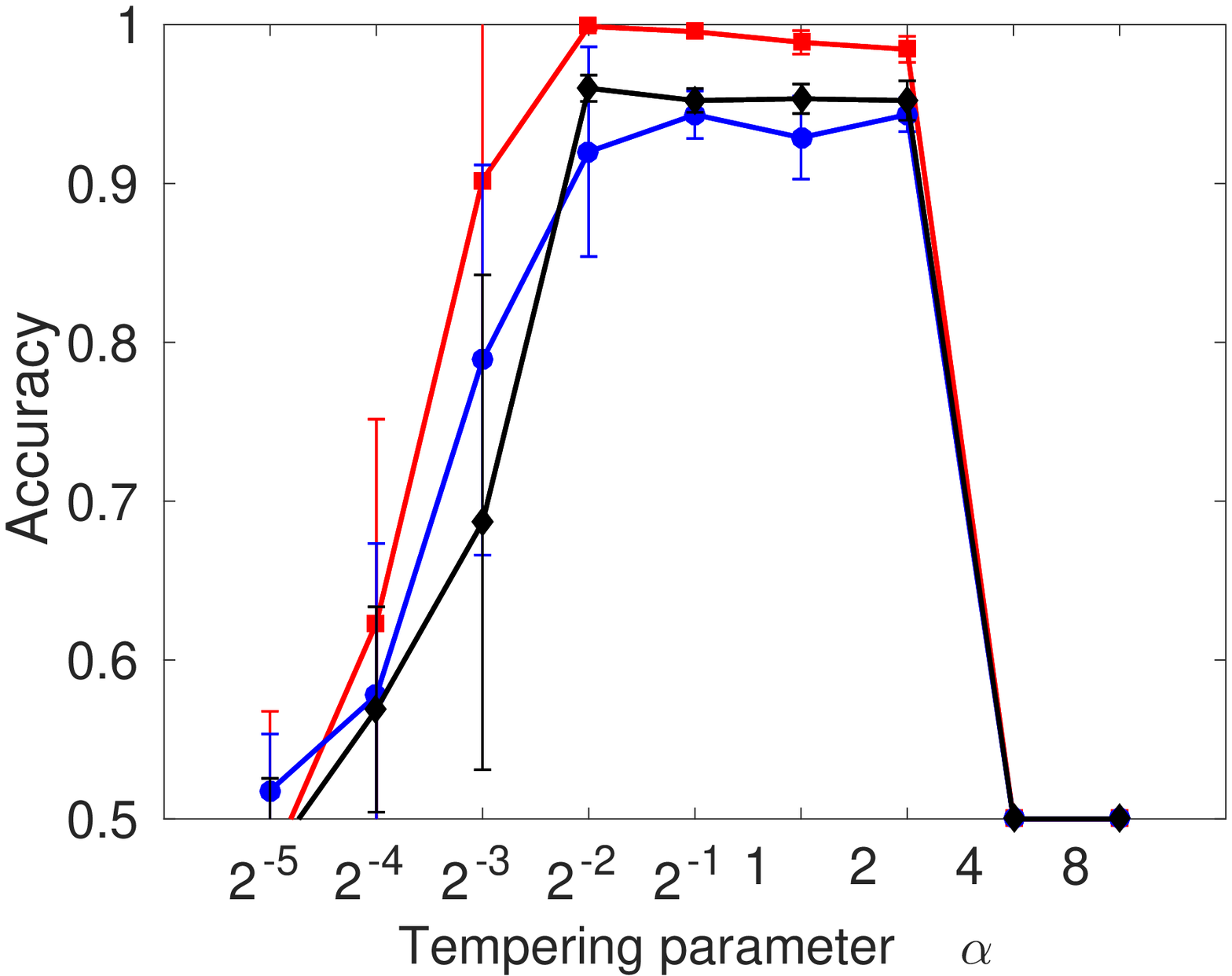}
\end{minipage}
\begin{minipage}[b]{0.45\linewidth}
\includegraphics[width=5cm,height=4.5cm]{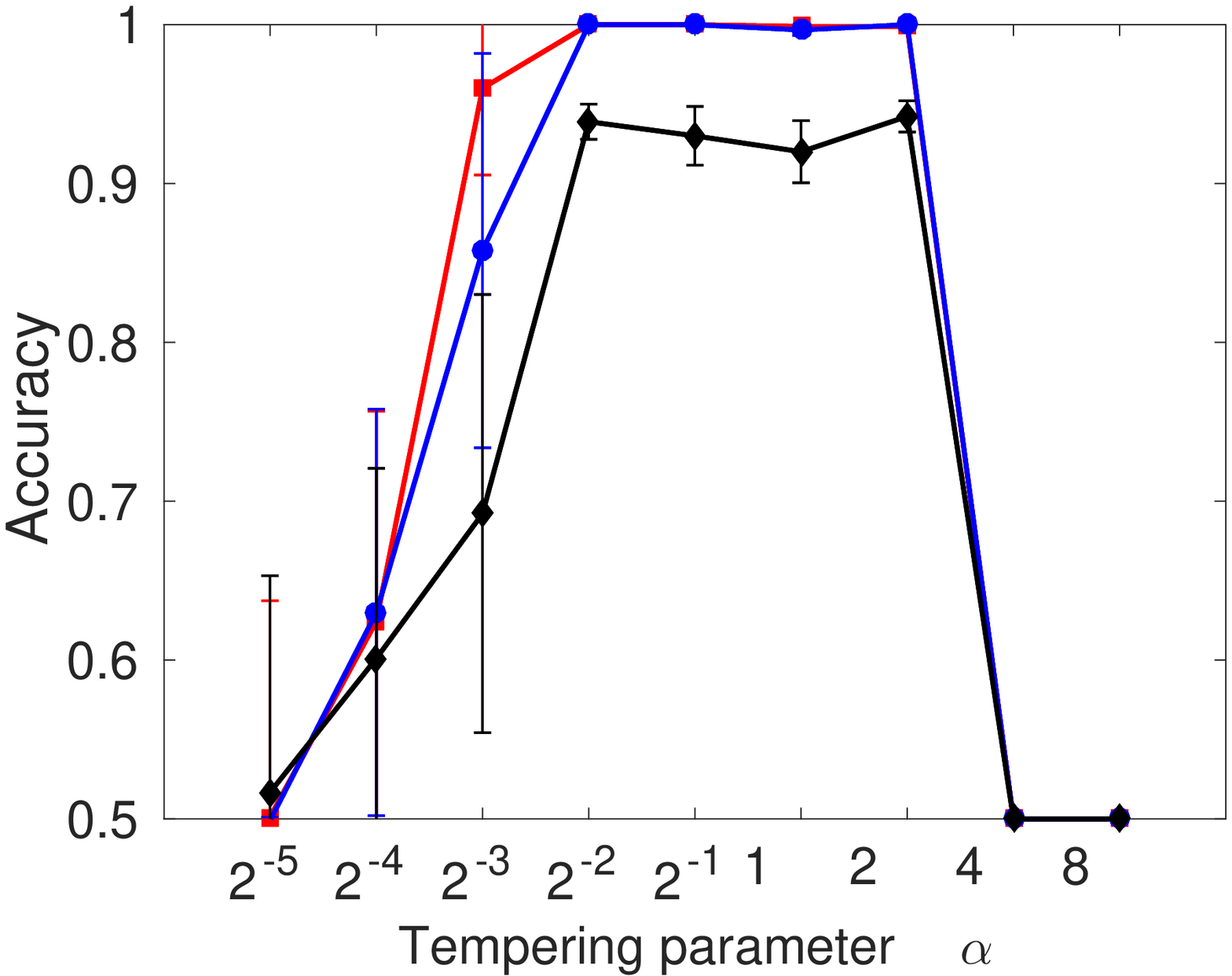}
\end{minipage}
\vspace{0.25cm}
\begin{minipage}[b]{0.45\linewidth}
\includegraphics[width=5cm,height=4.5cm]{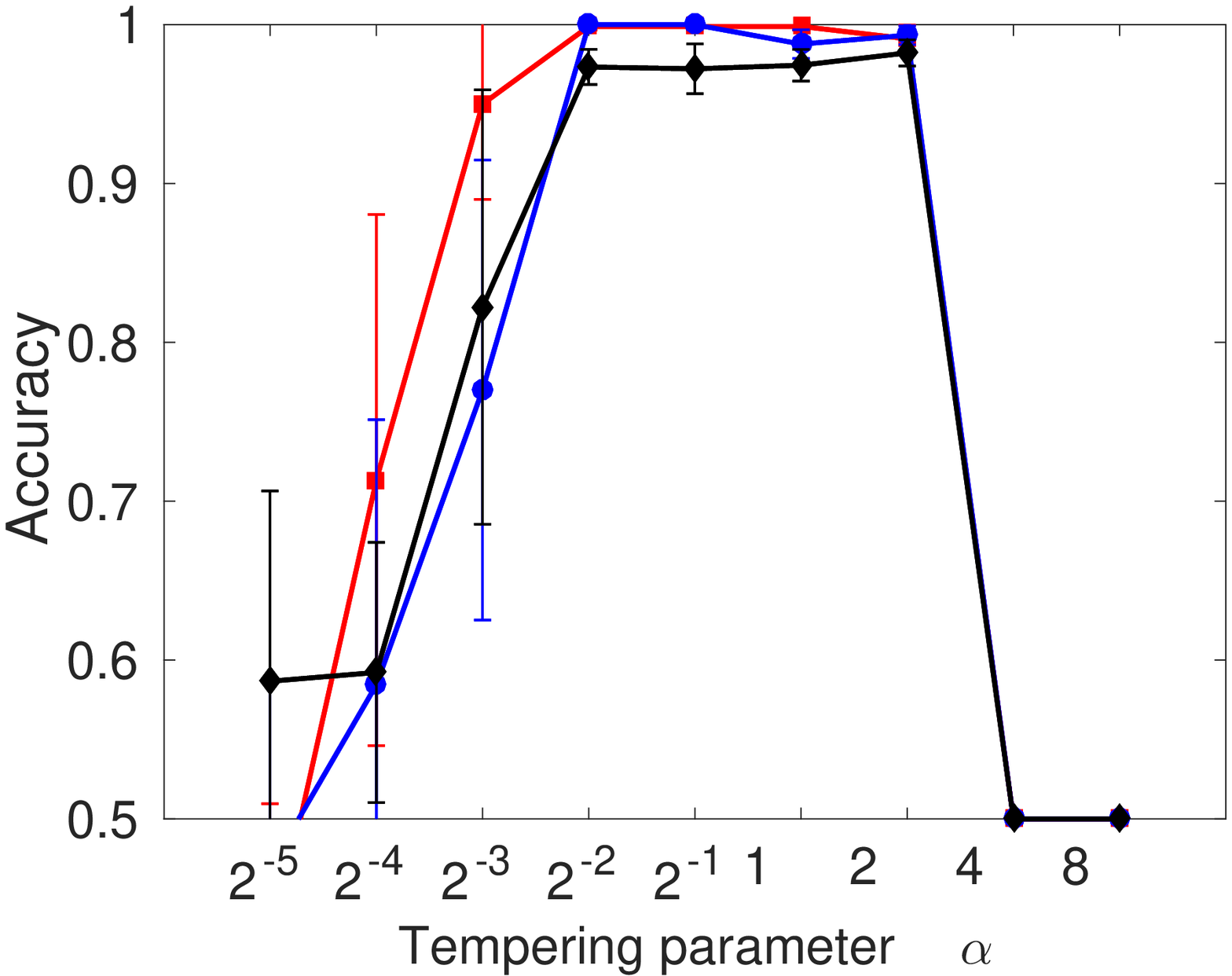}
\end{minipage}
%\quad
\begin{minipage}[b]{0.45\linewidth}
\includegraphics[width=5cm,height=4.5cm]{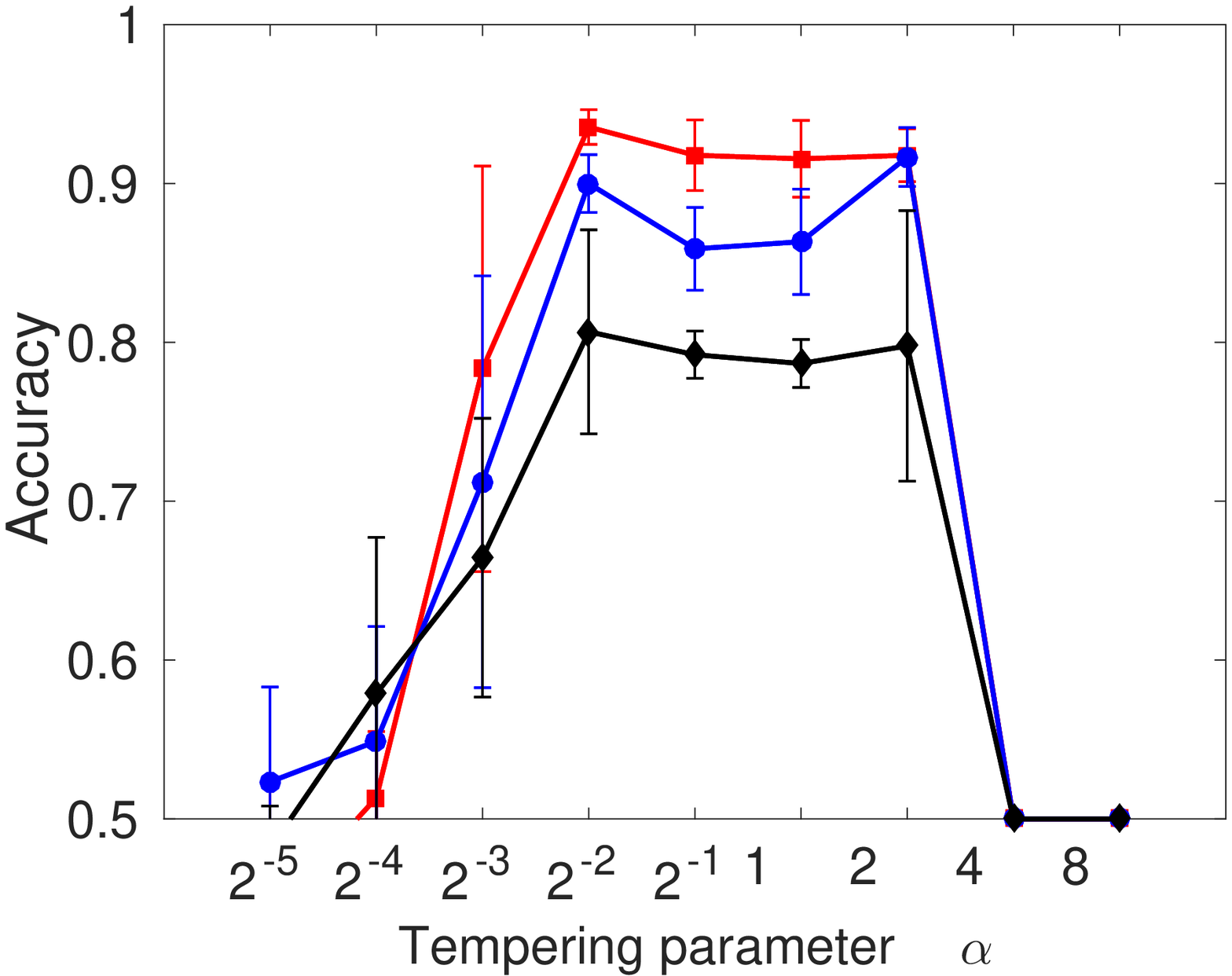}
\end{minipage}
\begin{minipage}[b]{0.45\linewidth}
\includegraphics[width=5cm,height=4.5cm]{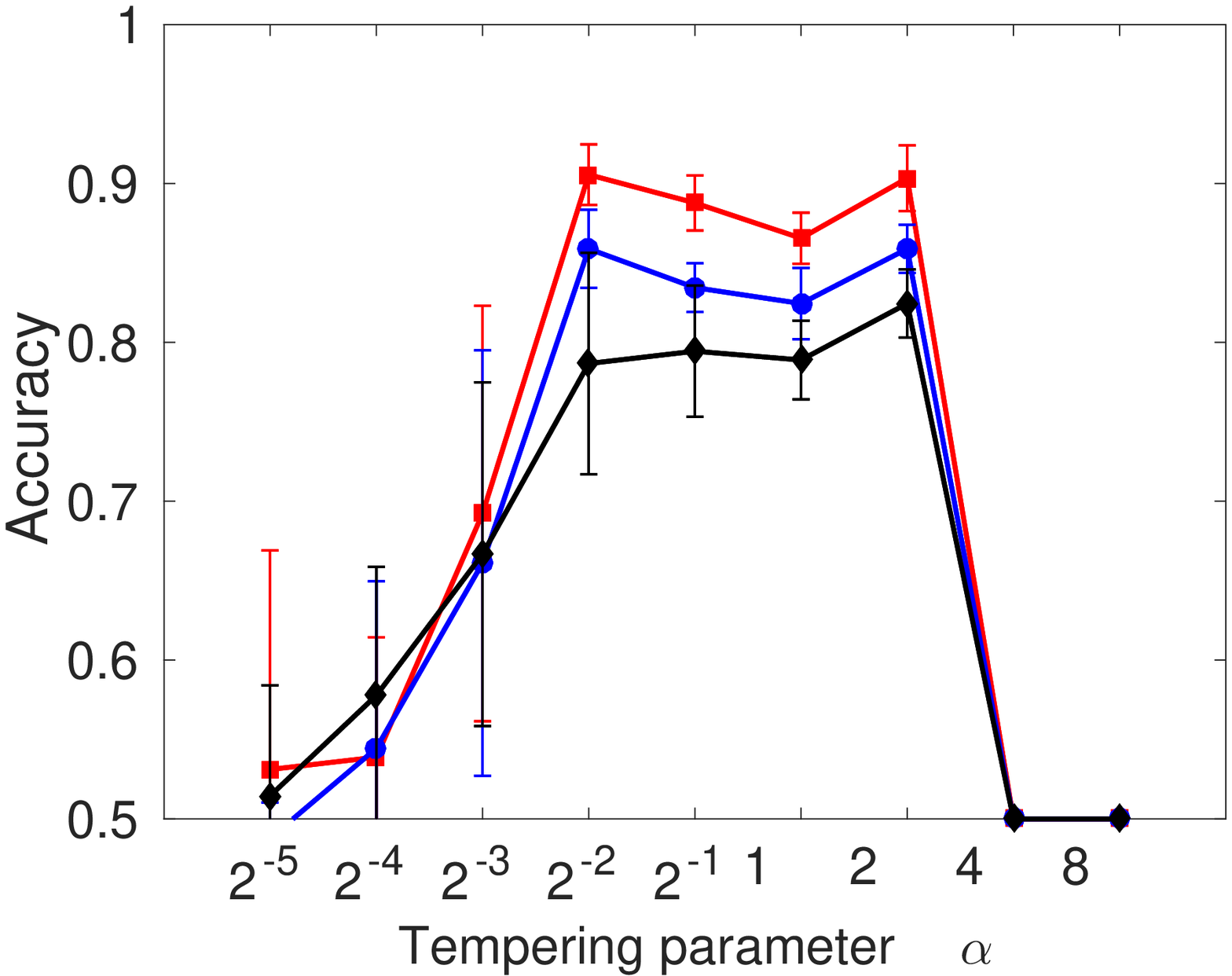}
\end{minipage}
\caption{The same as in Figure~\ref{PGDW-LiMS} but for classification performance as function of  log tempering parameter (i.e. $\log_{2} \alpha$) using Probability Product Kernel (PPK) method.
\label{PGDW-PPK}}
\end{figure}

\begin{figure}[!t]
\centering
\begin{minipage}[b]{0.45\linewidth}
\includegraphics[width=5cm,height=4.5cm]{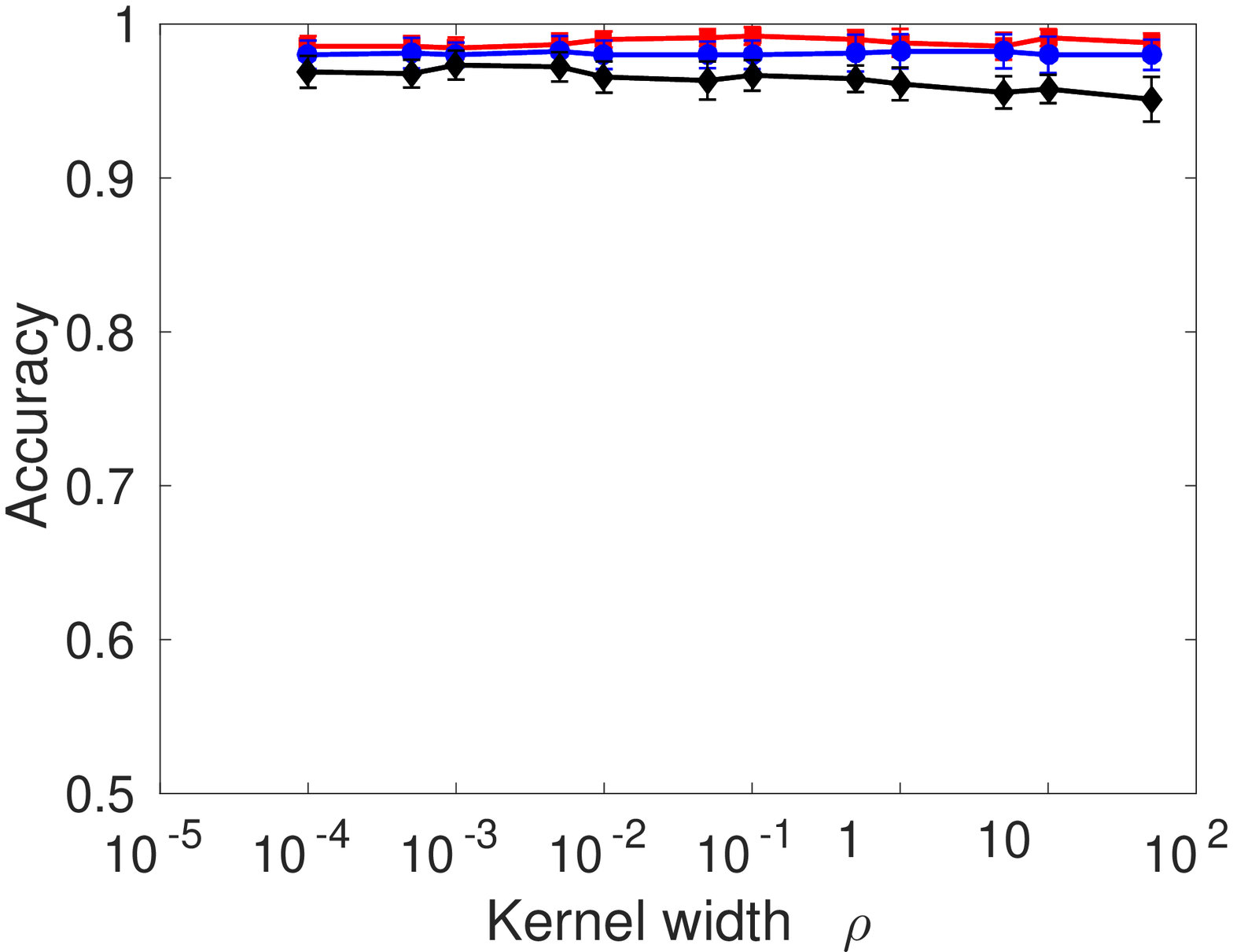}
\end{minipage}
\vspace{0.25cm}
\begin{minipage}[b]{0.45\linewidth}
\includegraphics[width=5cm,height=4.5cm]{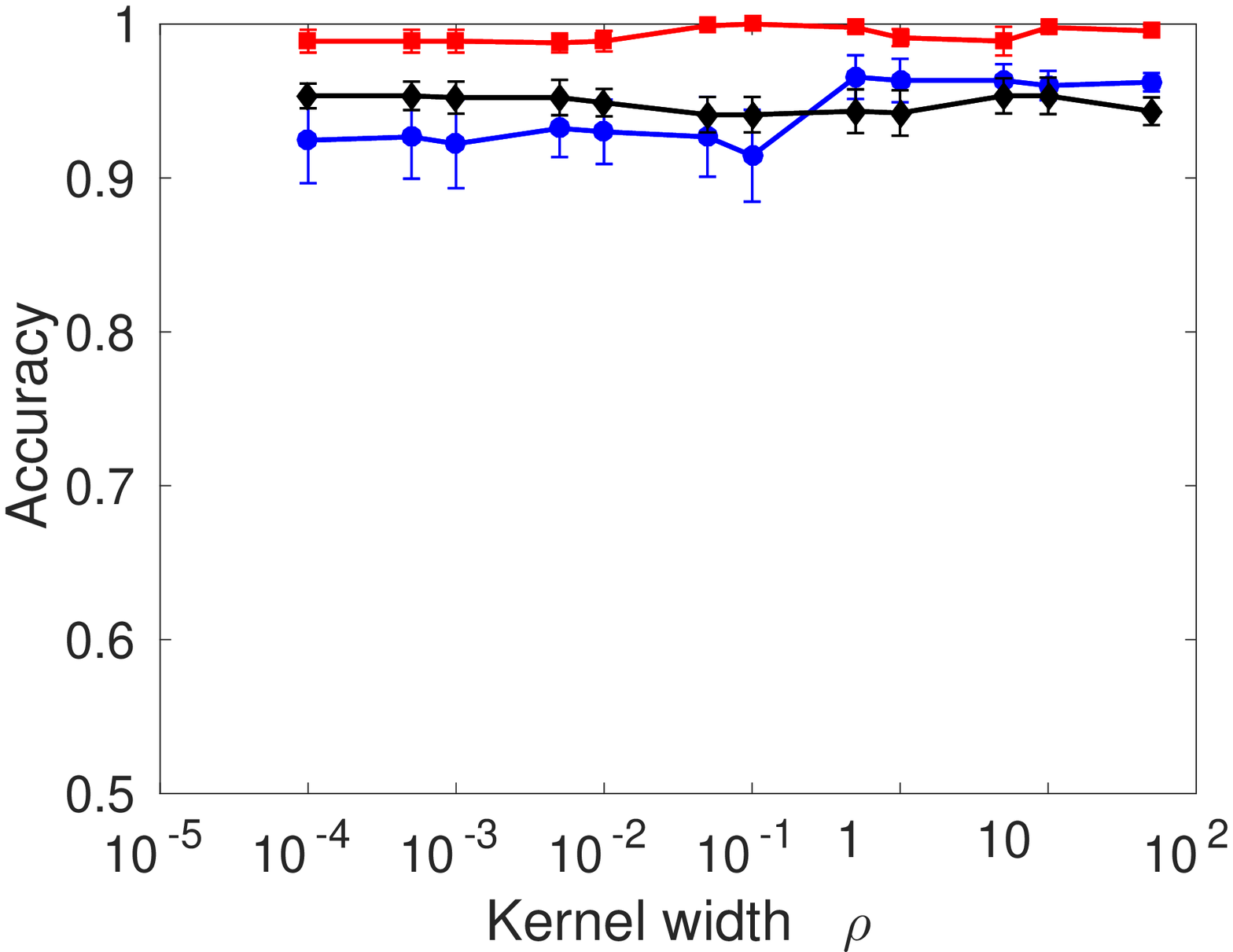}
\end{minipage}
\begin{minipage}[b]{0.45\linewidth}
\includegraphics[width=5cm,height=4.5cm]{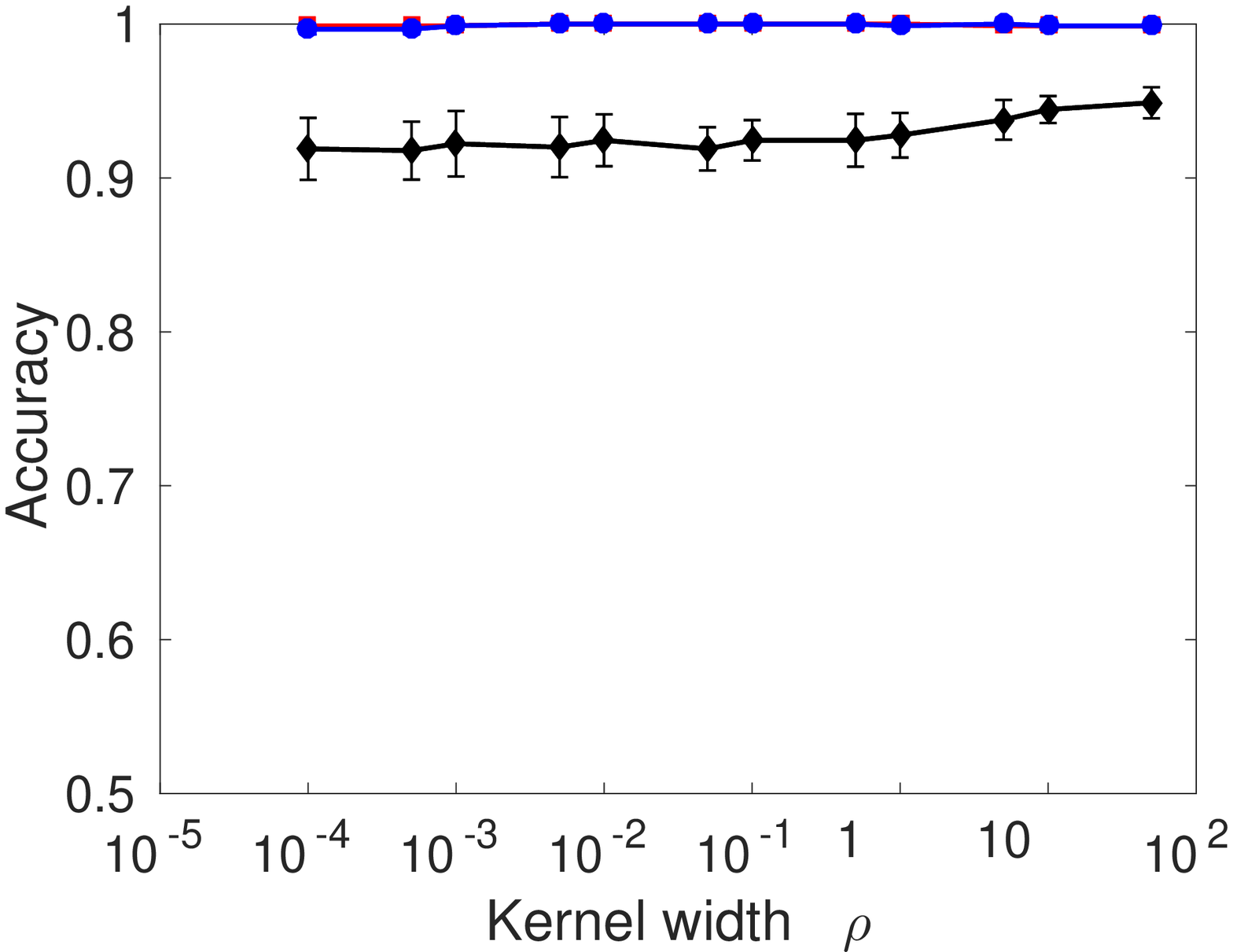}
\end{minipage}
\vspace{0.25cm}
\begin{minipage}[b]{0.45\linewidth}
\includegraphics[width=5cm,height=4.5cm]{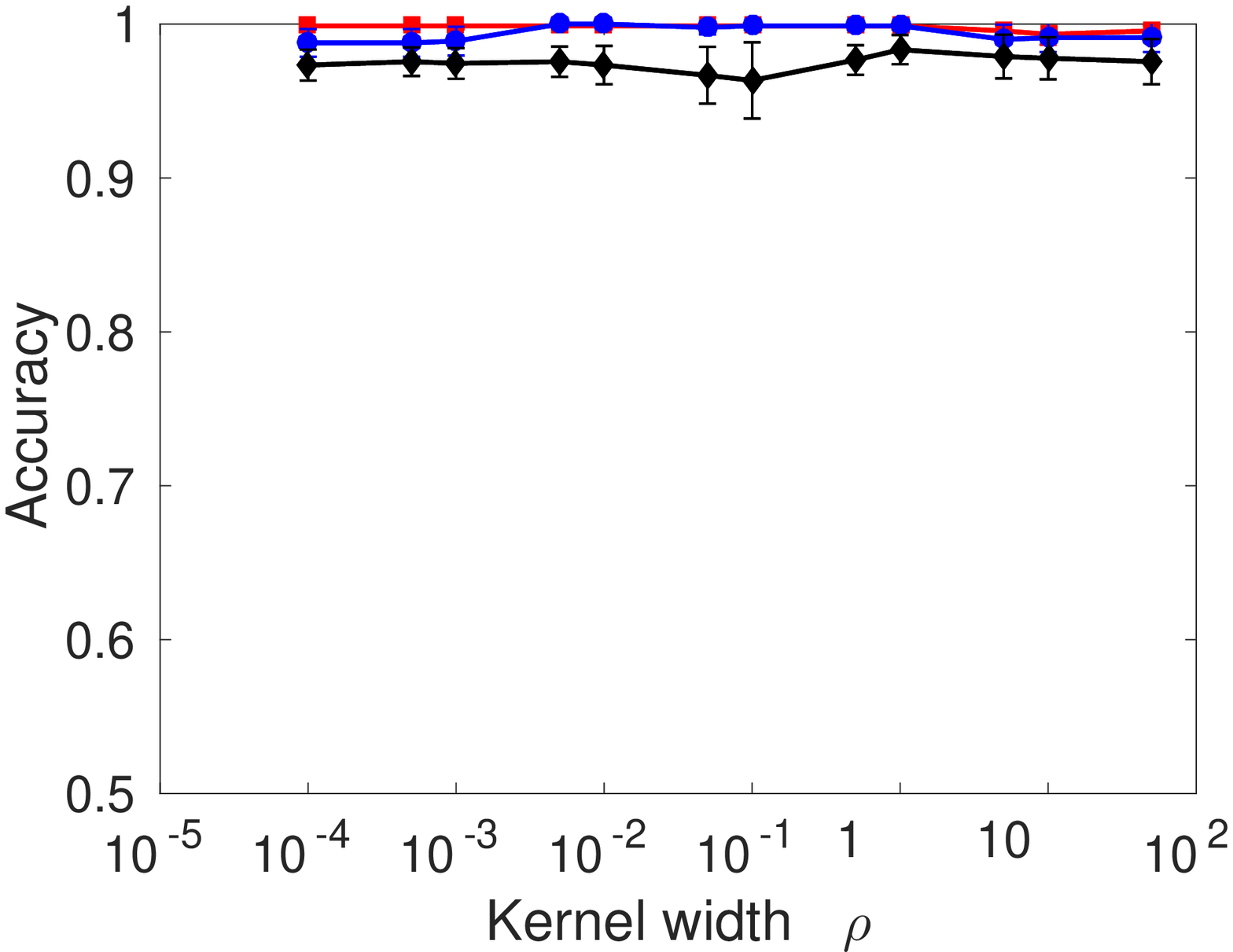}
\end{minipage}
%\quad
\begin{minipage}[b]{0.45\linewidth}
\includegraphics[width=5cm,height=4.5cm]{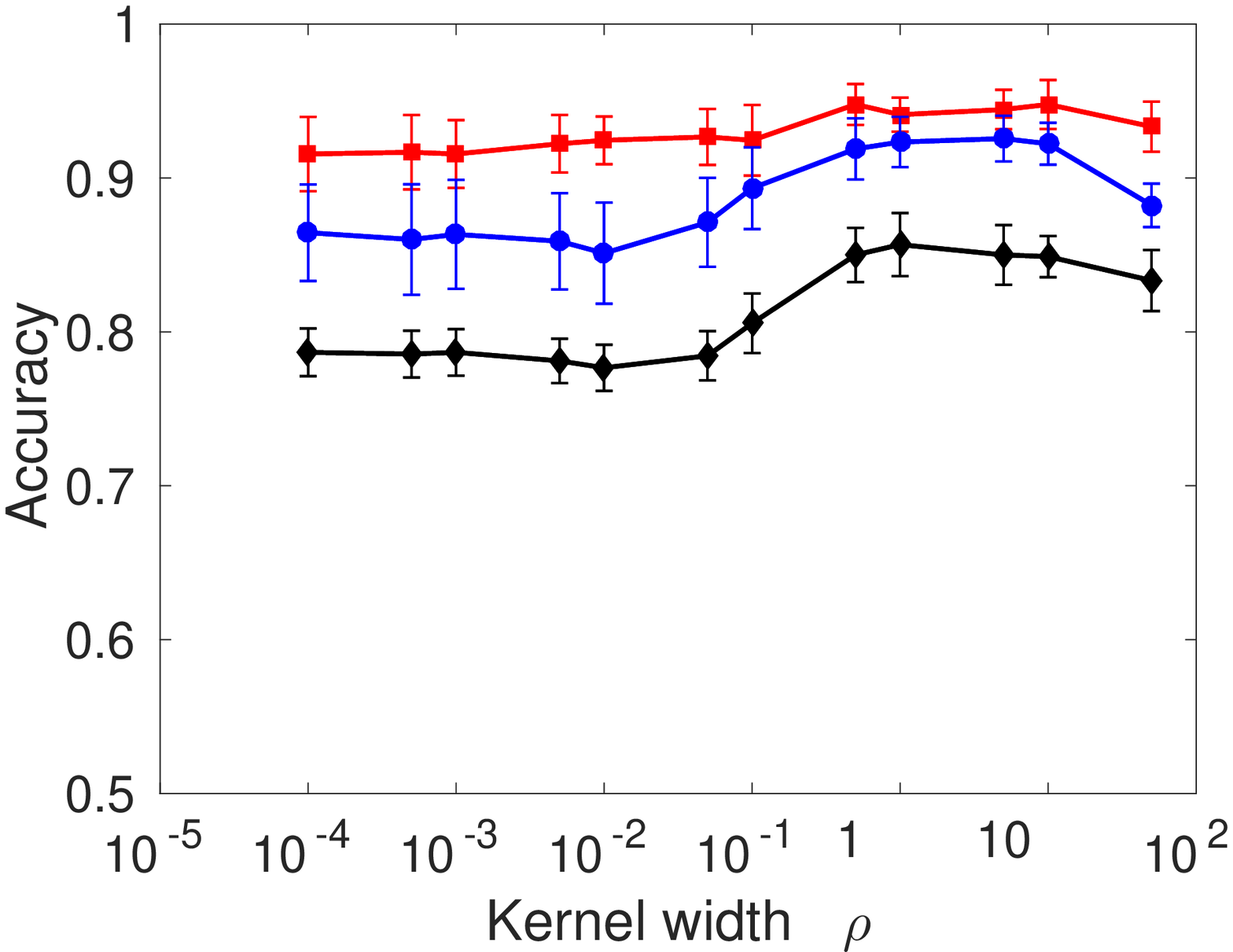}
\end{minipage}
\begin{minipage}[b]{0.45\linewidth}
\includegraphics[width=5cm,height=4.5cm]{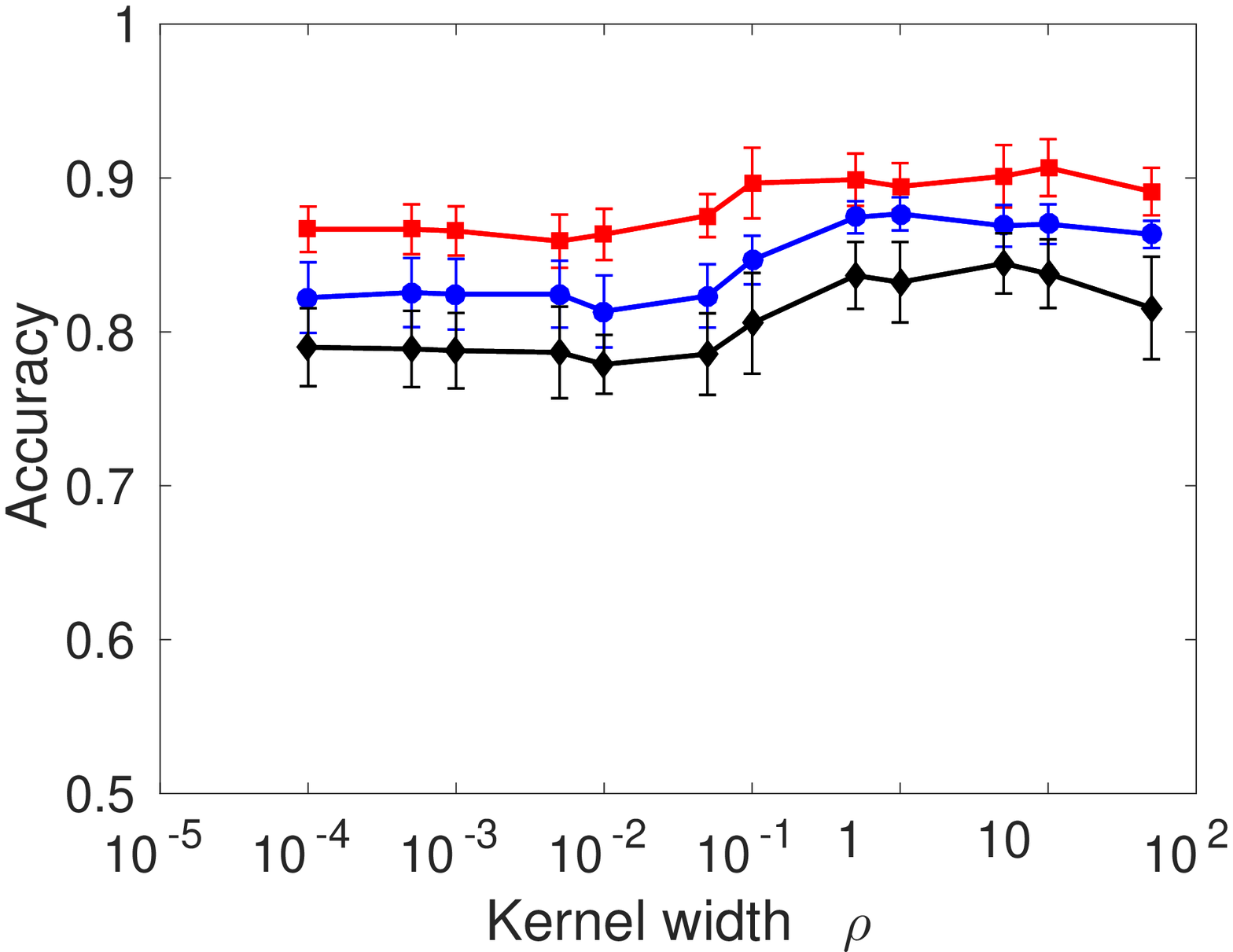}
\end{minipage}
\caption{The same as in Figure~\ref{PGDW-LiMS} but for KME classifiers. \label{PGDW-KME}}
\end{figure}

\begin{figure}[!ht]
\centering
\begin{minipage}[b]{0.45\linewidth}
\includegraphics[width=5cm,height=4.5cm]{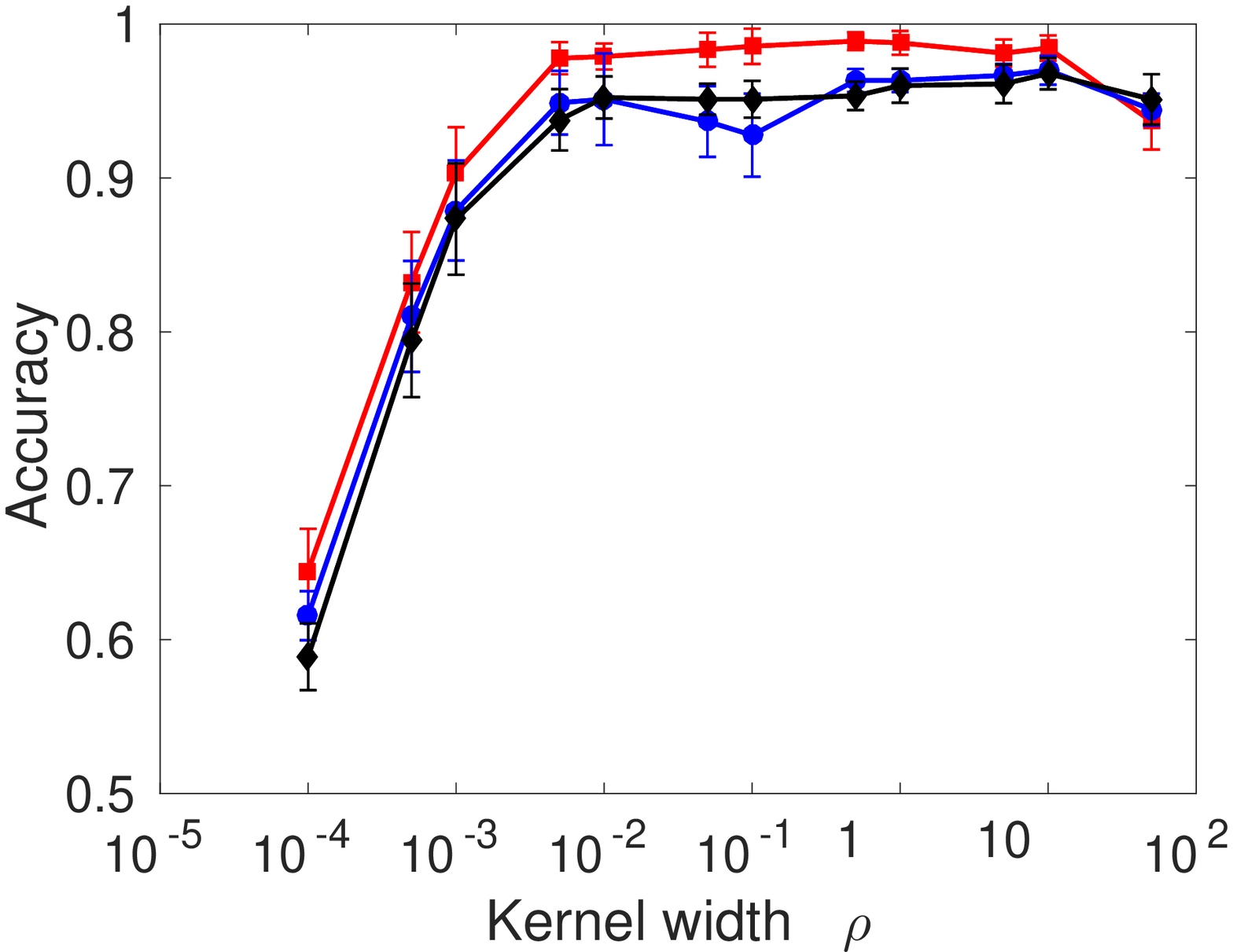}
\end{minipage}
\vspace{0.25cm}
\begin{minipage}[b]{0.45\linewidth}
\includegraphics[width=5cm,height=4.5cm]{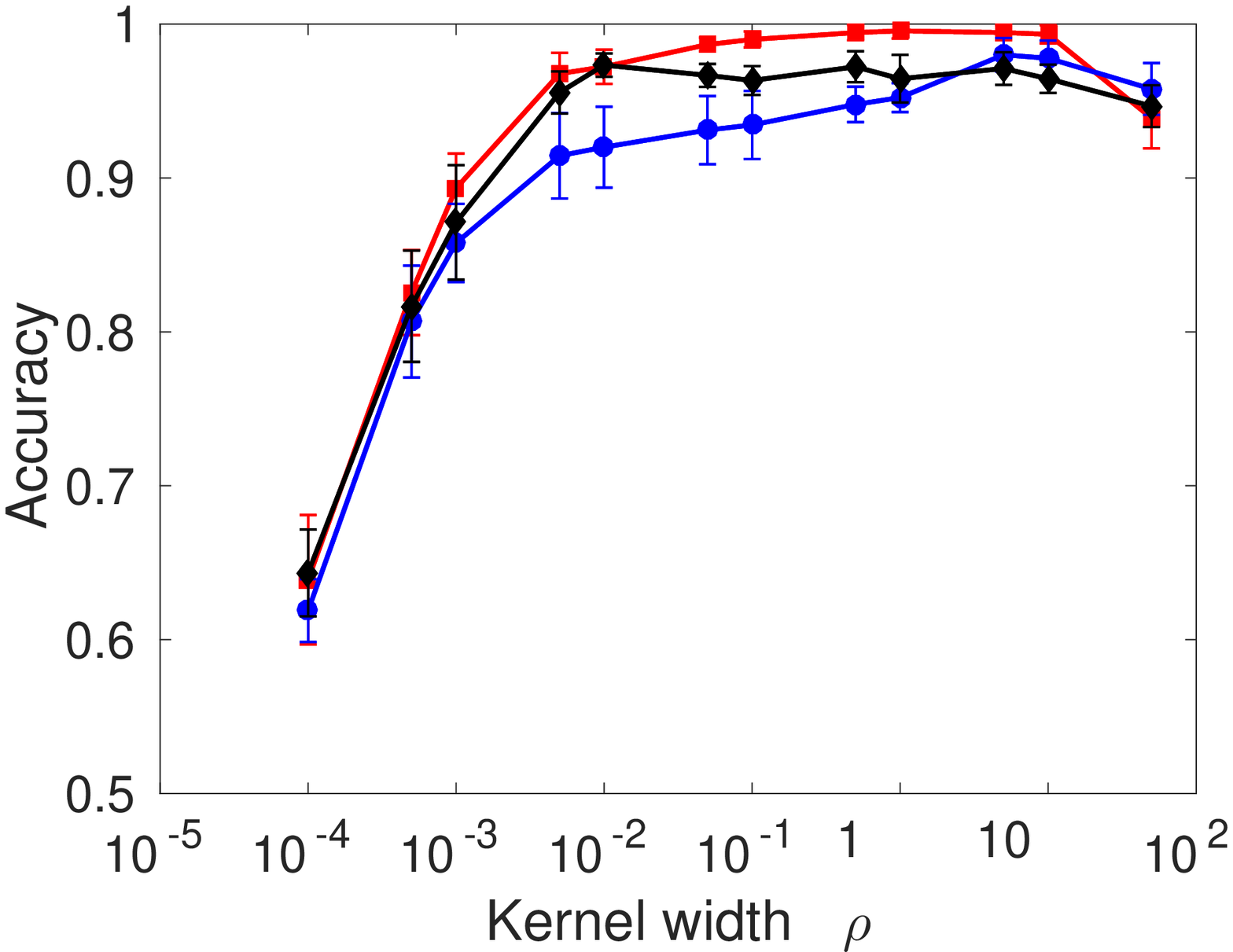}
\end{minipage}
\begin{minipage}[b]{0.45\linewidth}
\includegraphics[width=5cm,height=4.5cm]{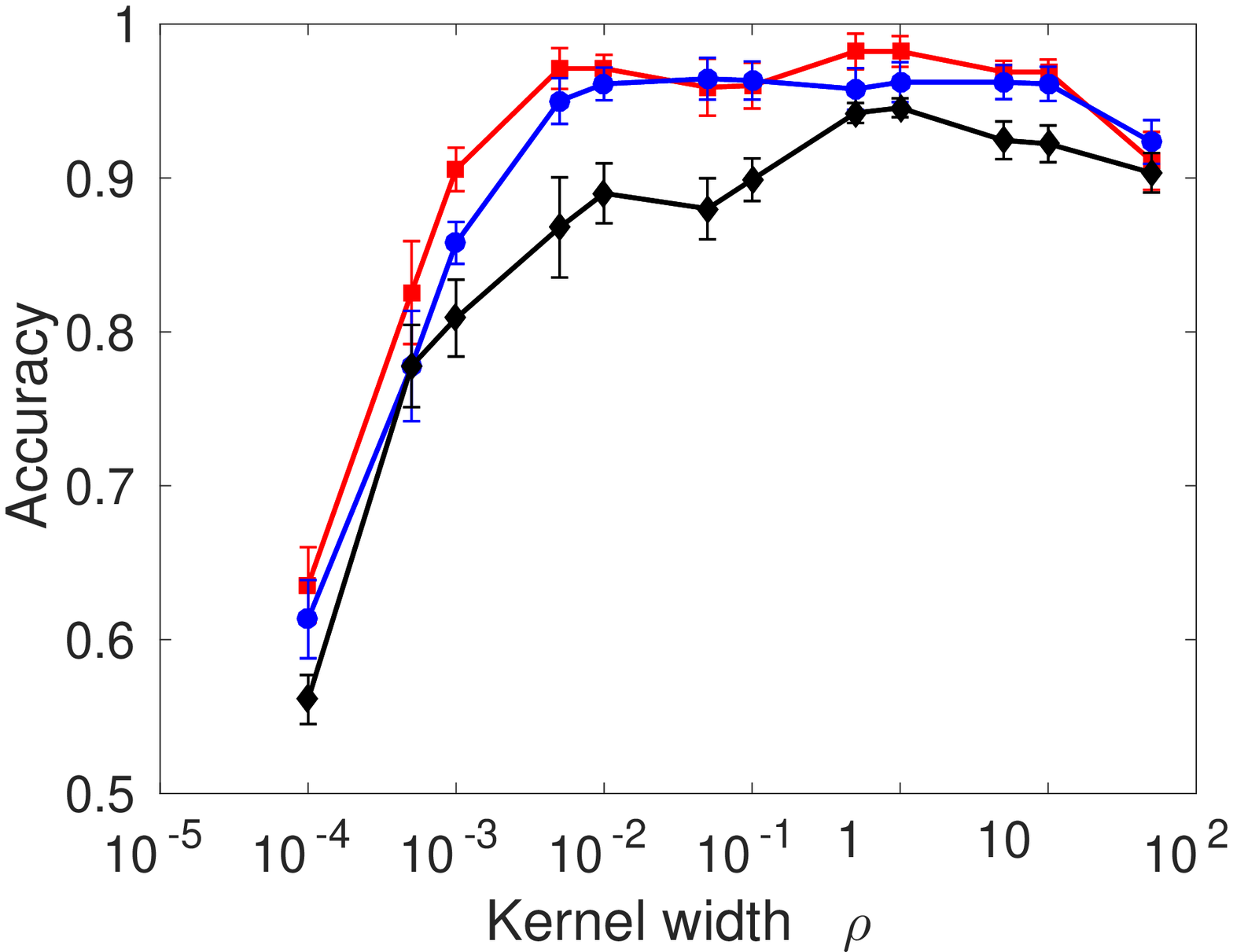}
\end{minipage}
\vspace{0.25cm}
\begin{minipage}[b]{0.45\linewidth}
\includegraphics[width=5cm,height=4.5cm]{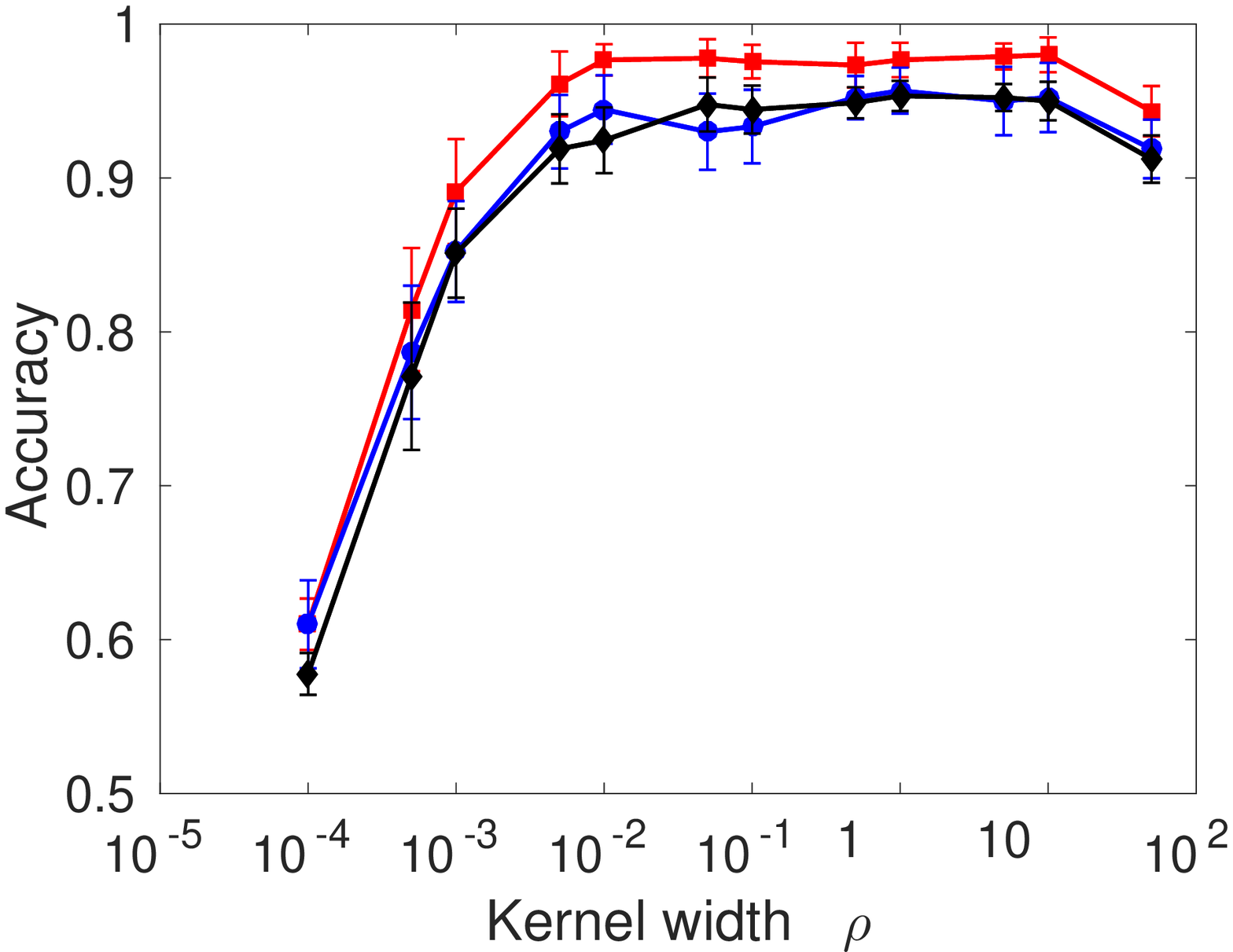}
\end{minipage}
%\quad
\begin{minipage}[b]{0.45\linewidth}
\includegraphics[width=5cm,height=4.5cm]{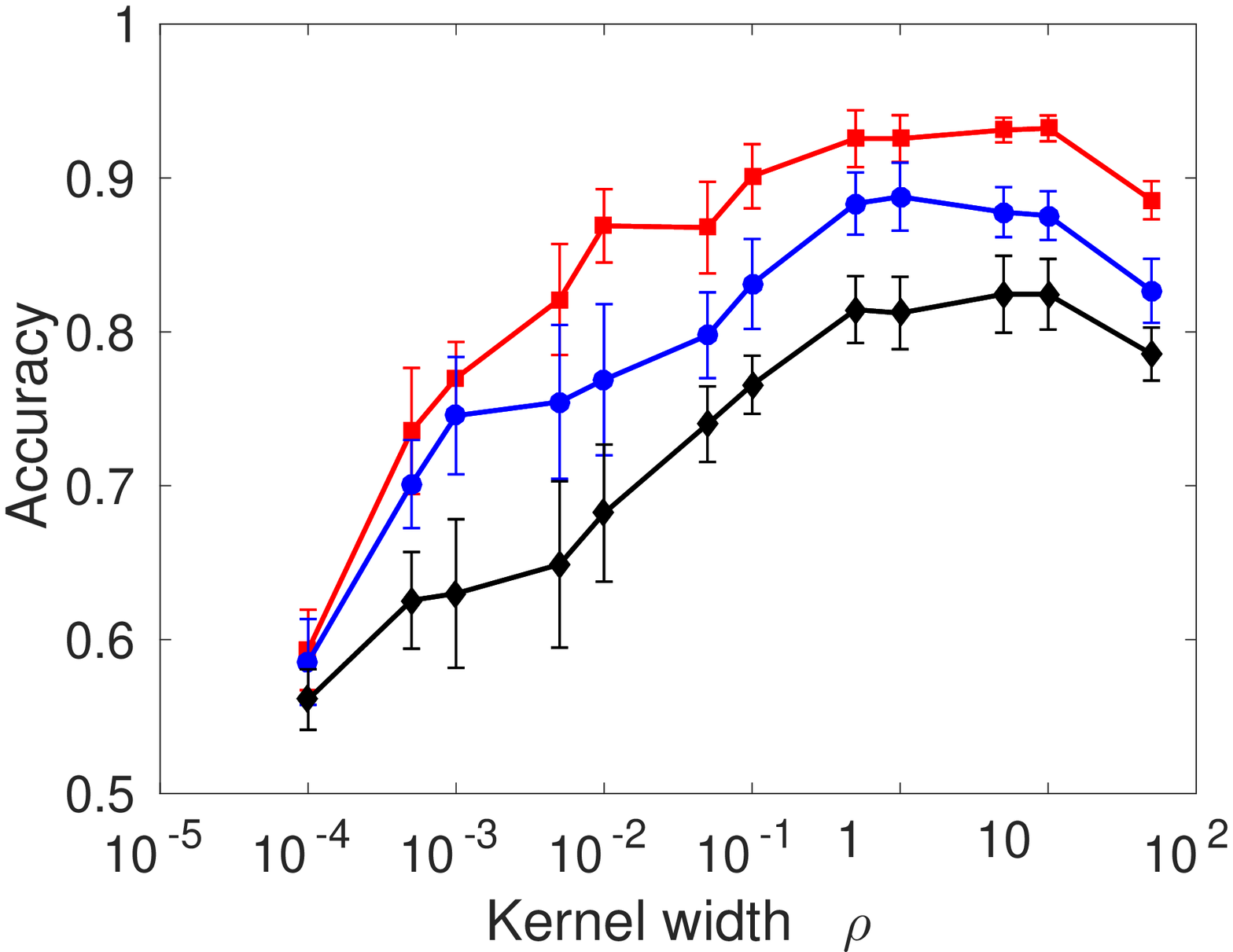}
\end{minipage}
\begin{minipage}[b]{0.45\linewidth}
\includegraphics[width=5cm,height=4.5cm]{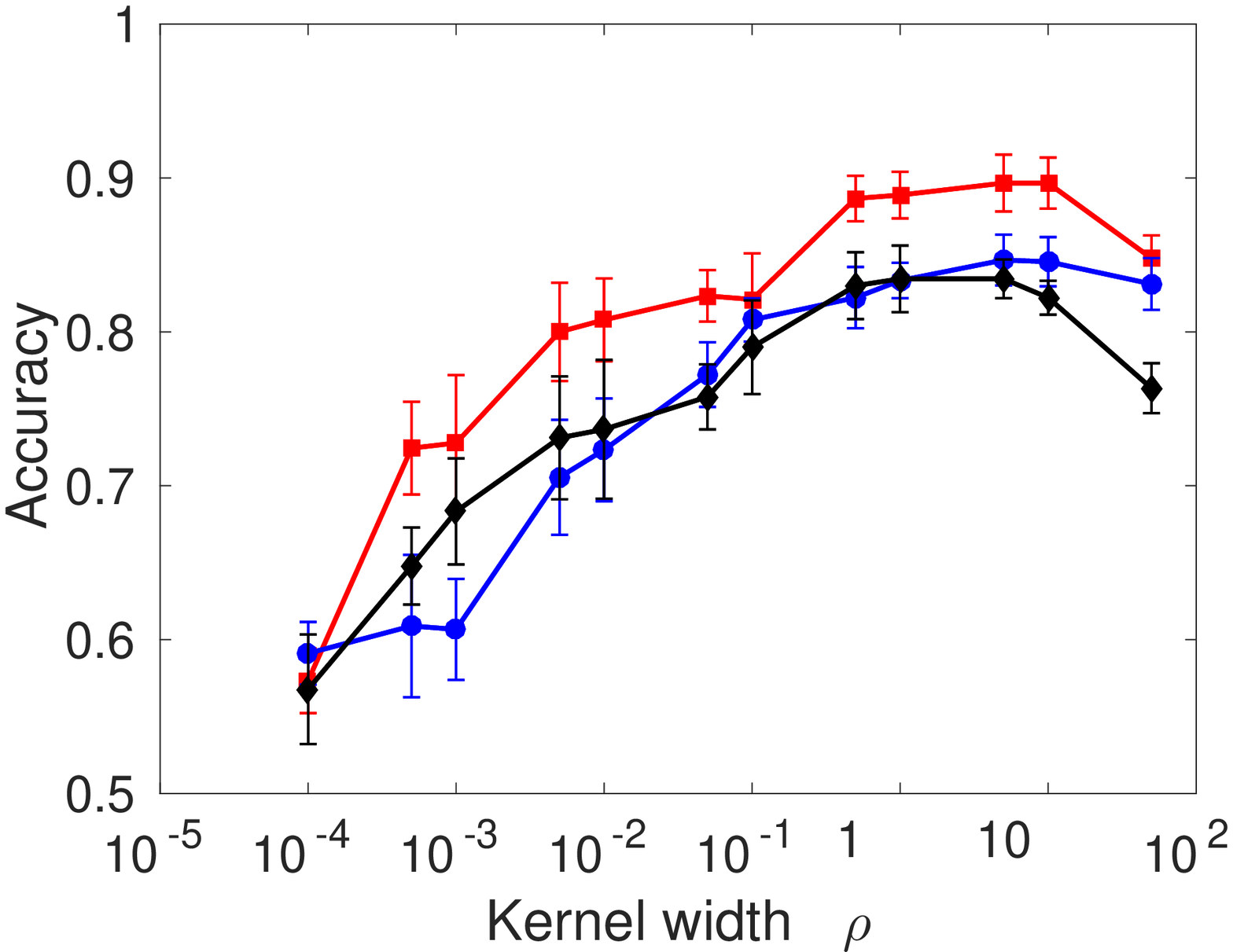}
\end{minipage}
\caption{The same as in Figure~\ref{PGDW-LiMS} but for bKLR classifiers. \label{PGDW-KLR}}
\end{figure}

Figure~\ref{PGDW-LiMS} shows the LiMS  performance as a function of kernel width $\rho$ for {\it Task 1} -- {\it Task 3} and for different combinations of $\sigma$ and $ISI$ values. In particular, in plots on the left the $ISI$ is fixed to 0.5 and $\sigma = 0.3, 0.4, 0.6$; in plots on the right the $\sigma$ is fixed to 0.3 and $ISI = 0.5,1,1.25$. Overall, the classification performance remains robust over a fairly large interval of intermediate $\rho$-values ranging from 0.01 to 1.0. Naturally, there is a drop in classification performance at very large kernel width $\rho$ = 10. Further, as expected, the performance decreased monotonically with increasing $\sigma$ or $ISI$ for all intermediate kernel widths. These findings match observations made in the GnRH experiments. Figure~\ref{PGDW-PPK} shows that the PPK classifier maintains its maximum performance over an interval of intermediate tempering parameter values ranging from $\alpha$ = $2^{-3}$ to $\alpha$ = 2. Recall that for GnRH models, the PPK classifier attained the best performance for $\alpha \le 1.0$. Values  of $\alpha > 1$ effectively make the  input posterior distributions over the models more peaked prior to classification. Unlike in the GnRH experiments, in general the KME classifier retains its best performance for larger kernel widths $\rho \ge 0.5$ (see Figure~\ref{PGDW-KME}). Figure~\ref{PGDW-KLR} shows that the performance of the baseline bKLR classifier increased steadily with the kernel width, achieving its best performance over a range of large kernel widths. For subsequent analysis,  we chose (using the validation data) $\rho$ = 0.05, $\alpha$ = 2, $\rho$ = 1 and $\rho$ = 1 as the overall kernel parameters for the LiMS, PPK KME and bKLR classifiers, respectively.

To compare the four classifiers in a statistical manner, we tested  six different one-sided hypothesis:
({\bf H1}) LiMS outperforms KME;
({\bf H2}) LiMS outperforms PPK;
({\bf H3}) KME outperforms PPK;
({\bf H4}) LiMS outperforms bKLR;
({\bf H5}) KME outperforms bKLR and
({\bf H6}) PPK outperforms bKLR.
In particular, the hypothesis {\bf H4}--{\bf H6} address the question what kind of learning in the model space classifiers can outperform the baseline classifier operating in the signal space. The results are summarised in Table~\ref{signrank_dw_1}. All $p$-values smaller than 0.15 are highlighted in bold font. Table~\ref{signrank_dw_1} shows that for {\it Task 2}, all three posterior-based classifiers clearly outperform the bKLR classifier. For {\it Task 3}, LiMS and KME still outperform bKLR.  The results for the simplest {\it Task 1} indicate that LiMS would have the upper hand against bKLR but the overall trend is not clear. Indeed, in {\it Task 1} the two classes can be conveniently separated in the signal space. This analysis shows the overall superiority of LiMS (but not KME) over PPK. Interestingly enough, we observed the same general trend of decreasing classifier performance with increasing model uncertainty in the input model posteriors.

\begin{table}[!ht]
\centering
\begin{tabular}{| c | c |  c | c | c | c | c | c |}
\hline
Task & Entropy & {\bf H1} & {\bf H2} &{\bf H3} &{\bf H4} &{\bf H5} &{\bf H6}  
\\ \hline
\multirow{5}{*} {\it Task 1} & 4.564 & {\bf 0.00} & {\bf 0.00} & {\bf 0.02}  & {\bf 0.07} & 0.73 & 0.99 \\
 & 4.634 & {\bf 0.01} & {\bf 0.10} & 0.89 & 0.48 & 1.00 & 0.93 \\ 
 & 4.654 & 0.25 & {\bf 0.04} & 0.30 & {\bf 0.00} & {\bf 0.01} & {\bf 0.01} \\ 
 & 4.656 & 0.67 & {\bf 0.01} &{\bf  0.01} & {\bf 0.03} & {\bf 0.09} & 0.89 \\ 
 & 4.756 & {\bf 0.13} & {\bf 0.06} & 0.62 & 0.22 & 0.39 & 0.47 \\ \hline
\multirow{5}{*} {\it Task 2} & 4.561 & 0.50 & {\bf 0.03} & {\bf 0.09} &  {\bf 0.00} &  {\bf 0.00} &  {\bf 0.00} \\ 
 & 4.682 & 0.50 & {\bf 0.06} & {\bf 0.13} & {\bf 0.00} & {\bf 0.00} & {\bf 0.00} \\ 
 & 4.693 & 0.50 & {\bf 0.13} & 0.50 & {\bf 0.00} & {\bf 0.01} & {\bf 0.01} \\ 
 & 4.707 & 0.50 & 1.00 & 1.00 & {\bf 0.00} & {\bf 0.00} & {\bf 0.00} \\ 
 & 4.835 & {\bf 0.00} & {\bf 0.00} & 0.94 & {\bf 0.00} & 1.00 & 0.73 \\ \hline
\multirow{5}{*} {\it Task 3} & 4.659 & {\bf 0.01} & {\bf 0.00} & {\bf 0.14} &  {\bf 0.01} &  {\bf 0.07} & 0.36 \\ 
 & 4.762 & 0.82 & {\bf 0.10} &  {\bf 0.03} & {\bf 0.00} & {\bf 0.00} & {\bf 0.02} \\ 
 & 4.775 & 0.22 & {\bf 0.05} & 0.31 & 0.40 & 0.73 & 0.78 \\ 
 & 4.837 & {\bf 0.13} & {\bf 0.06} & 0.16 & {\bf 0.00} & {\bf 0.02} & {\bf 0.04} \\ 
 & 5.026 & 0.73 & {\bf 0.13} & {\bf 0.08} & {\bf 0.00} & {\bf 0.01} & 0.37 \\ \hline 
\end{tabular} 
\caption{Sign-rank tests for comparing the classification performance between LiMS, KME, PPK, and bKLR classifiers in the three tasks of classifying partially observed double-well systems, using the following one-sided hypothesis:
({\bf H1}) LiMS outperforms KME;
({\bf H2}) LiMS outperforms PPK;
({\bf H3}) KME outperforms PPK;
({\bf H4}) LiMS outperforms bKLR;
({\bf H5}) KME outperforms bKLR and
({\bf H6}) PPK outperforms bKLR.
The $p$-values from these tests are given in Column 3--8 and all $p$-values smaller than 0.15 are highlighted in bold font. The level of model uncertainty is measured by (average) posterior entropy (Column 2).
\label{signrank_dw_1}}
\end{table}  

\begin{flushleft}
{\bf Experiment 2}
\end{flushleft} 
\vspace{0.1cm}
\noindent
 
Finally, we study to what degree can the use of a simpler model to obtain representative model posteriors  hamper the classier performance, provided the observations are generated by a much more complex model, yet the simpler model already embodies characteristics needed to perform the given classification task (see Section \ref{sec:issues}). In particular, we form an extended {\it Task 1}, {\it Task 1e},  in which time series in the observation sets were generated by complex stochastic multi-well systems with multimodal structure of the equilibrium distribution that can approximated (for the purposes of classification) by SDW systems (see Figure \ref{fig:pgdw}). The performance of LiMS classifier in {\it Task 1e}, reported in column 2 in Table~\ref{PGDW_reduction}, was compared with {\it Task 1} (column 3 of the same table). The $p$ values for the one-sided hypothesis stating that a better classification performance can be obtained in {\it Task 1e} than in  {\it Task 1} are given in column 4. Overall, the performance in {\it Task 1e} is as good as in {\it Task 1}. This confirms analogous findings in the GnRH experiment, where the use of simplified models, well aligned with the classification task, did not hamper the classification performance, even though the observation sequences were generated by much more complex models (see Section \ref{sec:GnRH_simple}, Experiment 2).

\begin{table}[!ht]
\centering
\begin{tabular}{| c |  c | c | c |}
\hline
($\sigma$, $ISI$) & {\it Task 1e}   &  {\it Task 1}   & $p$-value \\ \hline
(0.3,  0.5)        & 0.992 $\pm$ 0.005 &  0.996 $\pm$ 0.004   & 1.00  \\
(0.4, 0.5)        & 0.986 $\pm$ 0.006 &   0.987 $\pm$ 0.003   & 0.82  \\                         
{ (0.6, 0.5)}        & { 0.978 $\pm$ 0.009} &  { 0.974 $\pm$ 0.008}  & 0.09  \\ 
 (0.3, 1.0)        & 0.984 $\pm$ 0.004 &  0.996 $\pm$ 0.005   & 1.00  \\
(0.3, 1.25)        & 0.970 $\pm$ 0.006 &  0.991 $\pm$ 0.005   & 1.00  \\ \hline                        
\end{tabular} 
\caption{Comparison of classification performance between two different classes of data-generating SDW systems: multi-well systems vs. double well systems. Note that double well systems are the inferential model used in both cases.
\label{PGDW_reduction}}.
\end{table}

\section{Discussion and Conclusion}

\label{Sect:Conclusion}

In this paper, we have presented a general learning in the model space (LiMS) framework for classifying partially observed dynamical systems. The key ingredient of this framework is the use of posterior distributions over models to represent the individual observation sets, taking into account in a principled manner the uncertainty due to both the generative (observational and/or dynamic noise) and observation (sampling in time) processes. This is in contrast to the existing learning in the model space classification approaches that use model point estimates to represent data items. Another key ingredient of our approach is a new distributional classifier for classifying  posterior distributions over dynamical systems. 

We evaluated this classifier on two testbeds, namely a biological pathway model and a stochastic double-well system. Empirically the classifier clearly outperforms the classifier based on probability product kernel (PPK) - a state-of-the-art kernel method for classifying distributions. Moreover, its performance is comparable with a  recent distributional classification method based on kernel mean embedding. We derived a deep connection linking those three seemingly diverse approaches to distributional classification and provided a plausible explanation concerning superiority of the proposed classifier over the PPK classifier.

The experiments show a clear relation between model uncertainty and classification performance. As expected,  the performance drops with increasing model uncertainty. Principled treatment of model uncertainty in the learning in the model space approach is crucial in situations characterized by non-negligible observational noise and/or limited observation times. To illustrate this point further we also trained a baseline classifier that, given the observed time series, completely ignores the model uncertainty and instead of posterior distribution only employs the MAP point estimate of the model parameter. As all the other classifiers, the baseline classifier (referred to as MAP) is also implemented in the KLR framework.

We compared the three posterior based classifiers with the MAP classifier using both testbeds. The comparison follows the philosophy of comparing baseline classifier (bKLR) with the distributional classifiers in the SDW experiment (Columns 6--8 in Table~\ref{signrank_dw_1}). In particular, in the GnRH experiment, we tested three hypotheses (distributional classifier outperforms MAP) at nine uncertainty levels (see Column 1 in Table~\ref{signrank}).  Both LiMS and KME classifiers outperform (in the mean) the MAP classifier in all, except for one, uncertainty levels. For LiMS and KME, this superiority is statistically significant ($p$ $<$ 0.05) in all cases except for the lowest and the two lowest uncertainty levels, respectively. This is to be expected, as at low uncertainty levels the posterior over the models can be reasonably approximated by the MAP model estimate. In contrast, PPK classifier outperforms the MAP classifier only at 4 uncertainty levels, with statistical significance obtained only at the three highest uncertainty levels. In the SDW experiment, the tests were performed at 15 uncertainty levels (see Column 1--2 in Table~\ref{signrank_dw_1}). The LiMS, KME and PPK classifiers outperform the MAP classifier at all (15), 11 and 7 uncertainty levels, with statistical significance obtained at 9, 4 and 4 uncertainty levels, respectively. 

Crucially, we showed that the classifier performance would not be impaired when the model class used for inferring posterior distributions is much more simple than the observation-generating model class, provided the reduced complexity inferential model class captures the essential characteristics needed for the given classification task. This finding is potentially very significant for real-world applications.  Although mechanistic models encode expert domain knowledge and are of huge importance in forward modelling (e.g. assessing response to drug at certain dosage), such models may be too complex for the inferential (inverse-task) purposes. Fortunately,  much reduced model alternatives can be used in the learning in the model space framework if, as explained above, they already encode features important for the classification task. A semi-automated task-driven model simplification for learning in the model space framework is a matter of our future research. 

\section*{Acknowledgements}
This work was supported by the EPSRC grant ``Personalised Medicine Through Learning in the Model Space'' (grant number EP/L000296/1). KT-A gratefully acknowledges the financial support of the EPSRC via grant EP/N014391/1.

\end{document}